\documentclass[conference]{IEEEtran}

\usepackage{graphicx}
\usepackage{amsmath}
\usepackage{amssymb}
\usepackage{subfigure}
\usepackage[ruled]{algorithm2e}

\usepackage[compress]{cite}
\usepackage{multirow}
\usepackage{setspace}

\usepackage[pagebackref=true,breaklinks=true,letterpaper=true,colorlinks,bookmarks=false]{hyperref}

\def\mrm#1{{\mathrm{{#1}}}}

\newcommand{\argmin}{\operatornamewithlimits{argmin}}

\def\b#1{{\mathbf #1}}

\def\mrm#1{{\mathrm{{#1}}}}

\def\tr{\mrm{tr}}

\DeclareGraphicsExtensions{.jpg,.png}

\begin{document}

\title{Face frontalization for Alignment and Recognition}

\author{Christos Sagonas$^*$,  Yannis Panagakis$^*$, Stefanos Zafeiriou$^*$, and Maja Pantic$^{*\dagger}$\\
\begin{tabular}{cc}
$^*$Department of Computing, &  $^\dagger$EEMCS,\\ 
Imperial College London, & University of Twente,\\ 
180 Queens Gate, & Drienerlolaan 5,\\
London SW7 2AZ, U.K. & 7522 NB Enschede, The Netherlands
\end{tabular}\\
{\tt\small \{\href{mailto:c.sagonas@imperial.ac.uk}{c.sagonas}, \href{mailto:i.panagakis@imperial.ac.uk}{i.panagakis}, \href{mailto:s.zafeiriou@imperial.ac.uk}{s.zafeiriou}, \href{mailto:m.pantic@imperial.ac.uk}{m.pantic}\}@imperial.ac.uk}
}

\maketitle

\begin{abstract}
   Recently, it was shown that excellent results can be achieved in both face landmark localization and pose-invariant face recognition. These breakthroughs are attributed to the efforts of the community to manually annotate facial images in many different poses and to collect 3D faces data. In this paper, we propose a novel method for joint face landmark localization and frontal face reconstruction (pose correction) using a small set of \textbf{frontal images only}. By observing that the frontal facial image is the one with the minimum rank from all different poses we formulate an appropriate model which is able to jointly recover the facial landmarks as well as the frontalized version of the face. To this end, a suitable optimization problem, involving the minimization of the nuclear norm and the matrix $\ell_1$ norm, is solved. The proposed method is assessed in frontal face reconstruction (pose correction), face landmark localization, and pose-invariant face recognition and verification by conducting experiments on $6$ facial images databases. The experimental results demonstrate the effectiveness of the proposed method.

\end{abstract}

\section{Introduction} \label{sec:intro}

Face analysis is one of the most popular computer vision problems. Important topics in face analysis include generic face alignment \cite{sdm,drmf} and automatic face recognition \cite{Feret_akshay_3d_iccv_2011,Feret_yi_3d_cvpr_2013}. These problems have been considered as separate, both creating a wealth of scientific research in Computer Vision. In particular state-of-the-art face alignment and landmark localization methods \cite{drmf,sdm} model the problem discriminatively by capitalizing on the availability of annotated, with regards to facial landmarks, data \cite{sagonas2013semi,sagonas300}. Unfortunately, the annotation of facial landmark is laborious, expensive, and time consuming process. This is more evident in cases where the face is not in frontal pose and some facial features and the boundary are neither visible nor well-defined.\footnote{From experience we know that annotation of facial image with poses take in many cases twice the time compared with frontal poses.}

\begin{figure}
\centering
\includegraphics[scale=0.15]{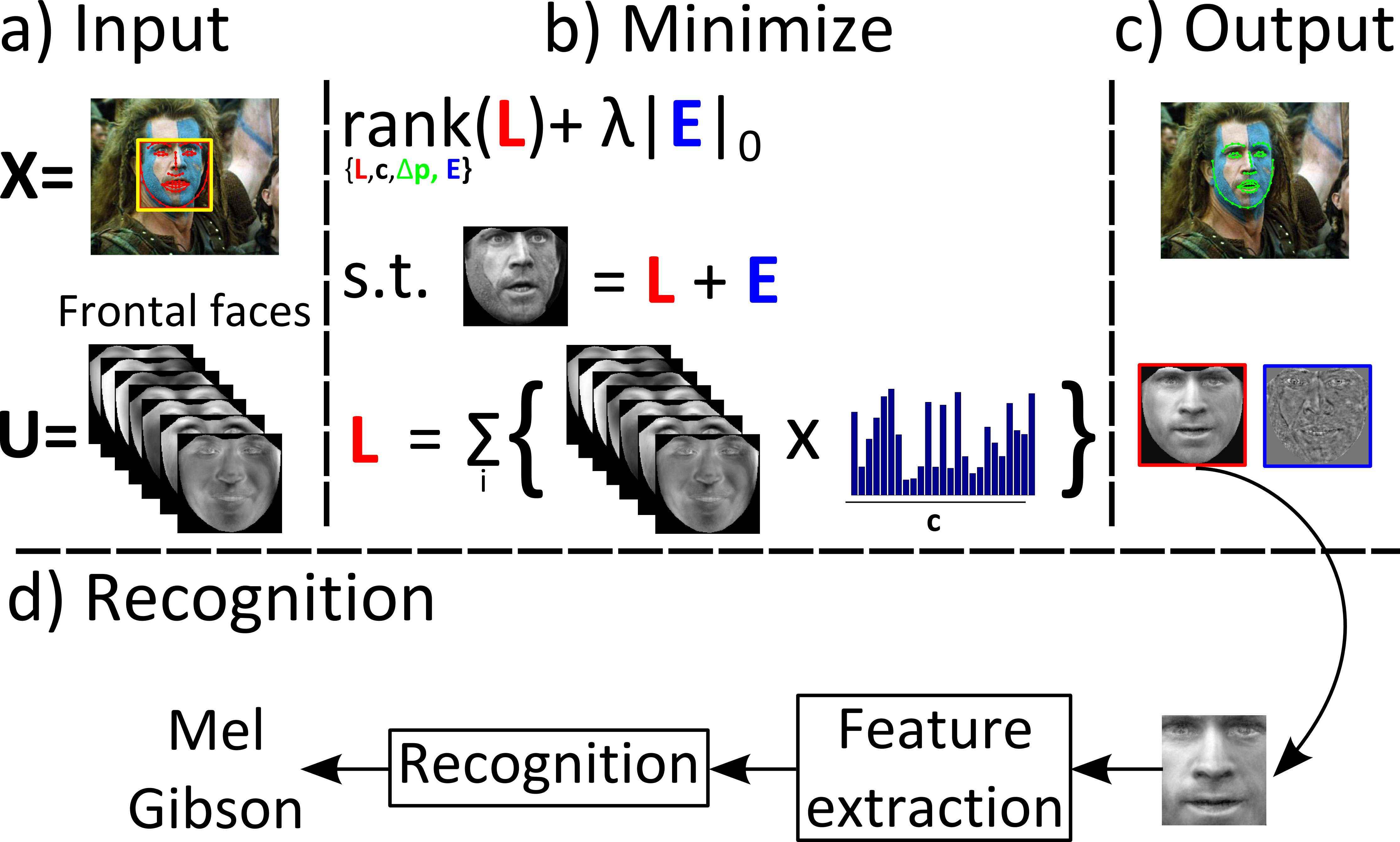}
\caption{Flowchart of the proposed method: a) Given an input image, the results from a detector, and a statistical model $\b{U}$, built on frontal images only, b) a constrained low-rank minimization problem is solved. c) Face alignment and frontal view reconstruction are performed  simultaneously. Finally, d) face recognition is succeeded using the frontalized image.}
\label{fig:flow_chart}
\end{figure}

Existing methods for face alignment can be roughly divided into two main categories: (a) \textit{Holistic} methods which use the whole texture of face as representation, and (b) \textit{part-based} methods which represent the face by using a set of local image patches extracted around of the predefined landmark points. The most-well known methods from the first category are the Active Appearance Models (AAMs) \cite{aams_cootes,aams_baker,aams_tzimiro} and the 3D Deformable Models (3DMs) \cite{3d_dm}. In the second category, methods such as the Active Shape Models (ASMs) \cite{asm_1} and the Constrained Local Models (CLMs) \cite{clm_1,clm_sharag} are included. Many of the above mentioned face alignment methods have achieved state-of-the-art results (e.g., \cite{drmf}, \cite{sdm}) in facial landmark localization under in-the-wild conditions but they are trained on many annotated samples from various poses. 

On the other hand, in the majority of face recognition systems the first and arguably, most important, step of face alignment is taken for granted using off the self methods \cite{sdm,viola2001rapid}. Even in the recent state-of-the-art face recognition methods, where millions of image are used to train feature extractors and classifiers, the pivotal step that increases their performance is that of face alignment  \cite{deepface,Feret_yi_3d_cvpr_2013}.
In such cases, the alignment step is very elaborate requiring to both locate landmarks and use 3D face models for pose corrections. In general, 3D model-based methods have high recognition accuracy due to the incorporation of the 3D model. However, such  methods cannot by widely applied since they require: (a) a method for accurate landmark localization in various poses, (b) to fit learned 3D model of faces, which is expensive to built, and (c) to develop robust image warping algorithms to reconstruct the frontal image \cite{deepface}. A recent approach that does not require a 3D model but only a small set of landmarks is presented in \cite{Feret_chellapa_2d_tip_2013}. This method aims to reconstruct the virtual view of an non-frontal image by employing  Markov Random Field. The main drawback of the aforementioned method is that for each non-frontal image an exhaustively batch-based alignment algorithm trained on frontal patches is applied. Clearly, such a procedure is time consuming.

\begin{figure}
\centering
\includegraphics[scale=0.35]{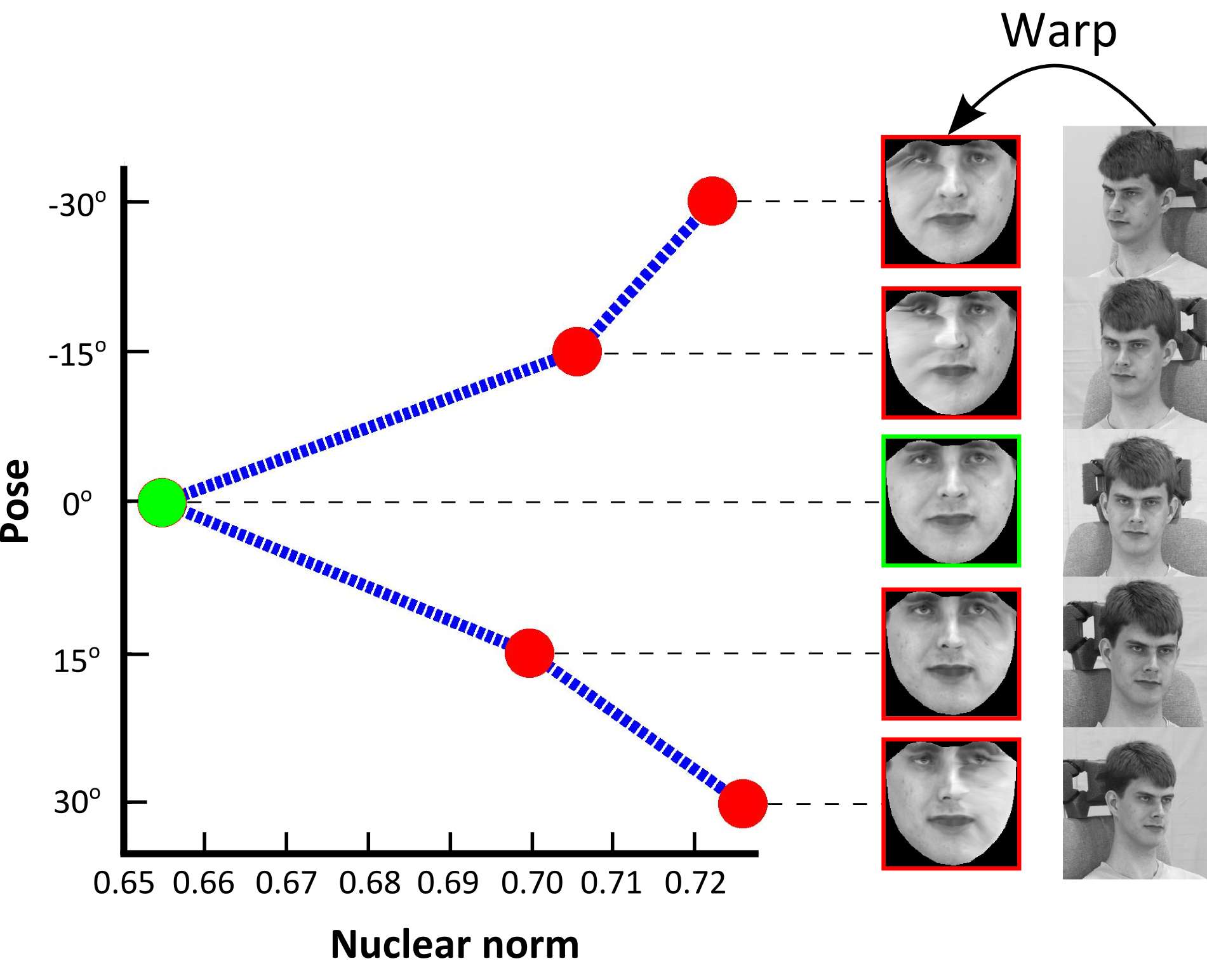}
\caption{The average value of nuclear norm computed based on neutral images of twenty subjects from MultiPIE database under poses $-30^0:30^0$. The initial images and the warped ones of a subject are also depicted.}\label{fig:low_rank}
\end{figure}

In this paper, motivated by the observation that the rank of a frontal facial image, due to the approximately structure of human face, is much smaller than the rank of facial images in other poses, we propose a unified framework for joint face frontalization (pose correction), landmark localization and (single sample) pose invariant face recognition. We show this can be achieved by using a model built from \textbf{frontal images only}. To validate the above observation `Neutral' images of twenty objects from MultiPIE database under poses $-30^0:30^0$ were warped into a reference frame and the nuclear norm (convex surrogate of the rank) of each shape-free texture was computed. In Fig. \ref{fig:low_rank} the average value of the nuclear norm for the different poses is depicted.  Clearly, the frontal pose has the smallest nuclear norm value compared the corresponding value of the rest of the poses. The flowchart of the proposed method (coined as FAR - \textit{Face frontalization for Alignment and Recognition}) is depicted in Fig \ref{fig:flow_chart}.

\begin{figure}[ht]
\begin{center}
\begin{tabular}{cc}
\begin{subfigure}{\includegraphics[width=0.2\textwidth]{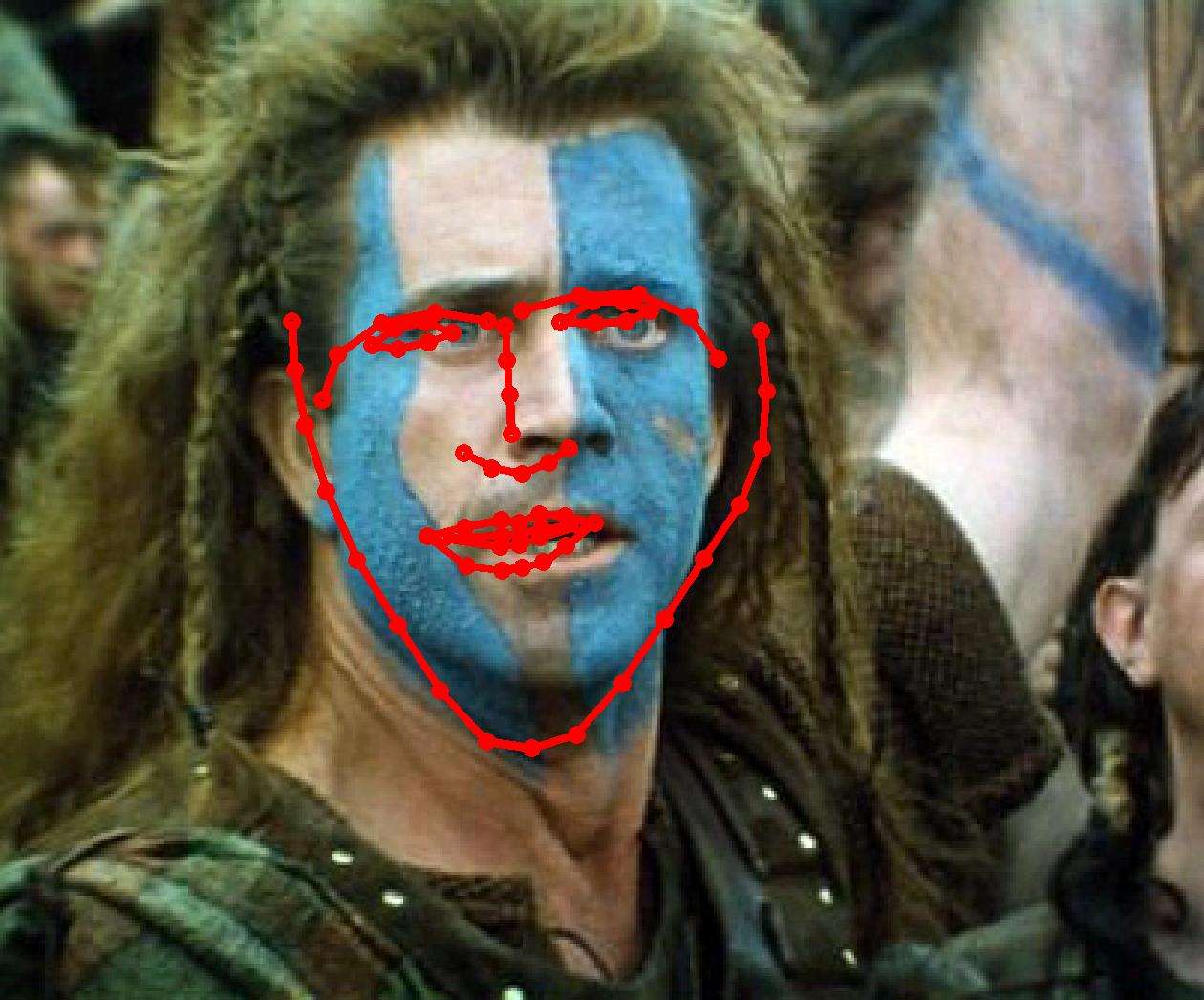}}\end{subfigure} &
\begin{subfigure}{\includegraphics[width=0.2\textwidth]{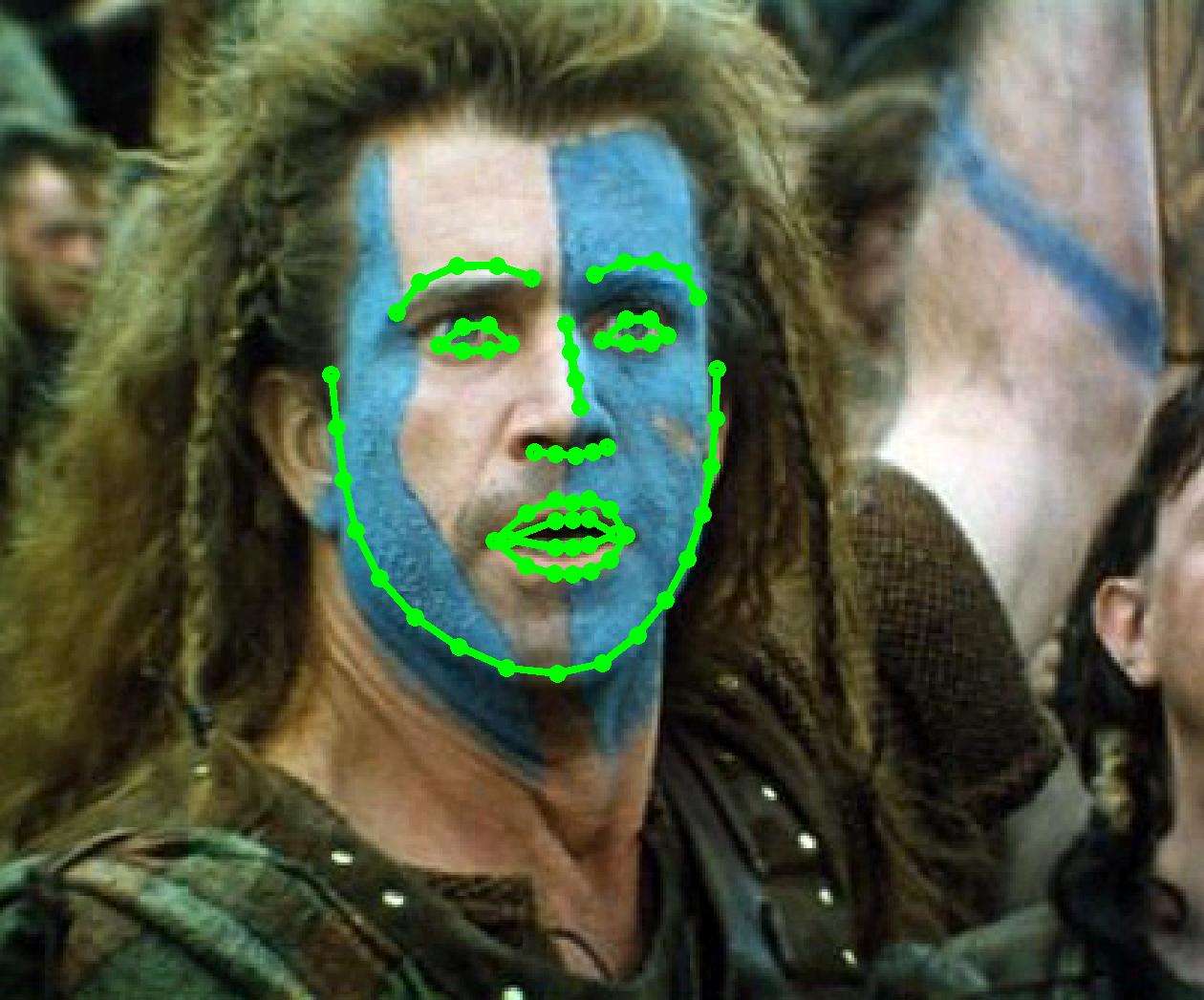}}\end{subfigure}\\
(a) & (b)
\end{tabular}
\end{center}
\caption{Results produced by the (a) non-rigid TILT and (b) FAR.}
\label{fig:tilt_vs_far}
\end{figure}

The most closely related work to the proposed method is the Transform Invariant Low-rank Textures (TILT) method \cite{tilt}. In TILT, texture rectification is obtained by applying a global affine transformation onto a low-rank term, modelling the texture. By blindly imposing low-rank constraints without regularization, for non-rigid alignment opposite effects may occur. We applied  the TILT with a non-rigid facial shape model and its performance was very poor as it can be observed in Fig. \ref{fig:tilt_vs_far}. Recently, it was demonstrated \cite{arasl,raps}, that non-rigid deformable models cannot be straightforward combined with optimization problems \cite{peng2012rasl} that involve low-rank terms without a proper regularization.
To overcome this and ensure that unnatural faces will be not created, a model of frontal images is employed. In that sense, our method can be seen as a deformable TILT model regularized within a frontal face subspace.

Summarizing, the contributions of the paper are:
\begin{itemize}
\vspace{-0.2cm}
\item Technical contribution: We develop a joint landmark localization and face frontalization method by proposing a deformable and appropriately regularized TILT.
\item Applications in computer vision:
\begin{enumerate}
\item  To the best of our knowledge this is the first generic landmark localization method which achieves state-of-the-art results using a  model of \textbf{frontal images only}.
\item It is possible to improve or match the state-of-the-art in unconstrained face recognition using only frontal faces and simple features for classification unlike other complex feature extraction procedures e.g., \cite{simonyan2013fisher}.\footnote{We note that we refer to the restricted protocol of the LFW \cite{lfwDB} and not to the unrestricted which unfortunately we cannot compete since we do not have access to millions of annotated faces.}
\end{enumerate} 
\end{itemize}

\section{Notation and preliminaries} \label{sec:notation}

Throughout the paper, scalars are denoted by lower-case letters, vectors (matrices) are denoted by lower-case (upper-case) boldface letters i.e., $\b{x}$, ($\b{X}$). $\b{I}$ denotes the identity matrix of compatible dimensions. The $i$th column of $\b{X}$ is denoted by $\b{x}_{i}$. A vector $\b{x} \in \mathbb{R}^{m \cdot n}$ (matrix $\b{X} \in \mathbb{R}^{m \times n}$) is reshaped into a matrix (vector) via the reshape operator : $\mathcal{R}_{m \times n}(\b{x}) = \b{X} \in \mathbb{R}^{m \times n}$, \big($\mbox{vec}(\b{X}) = \b{x} \in \mathbb{R}^{m\cdot n}$\big).

The $\ell_1$ and the $\ell_2$  norms of $\b{x}$ are defined as
$\Vert \b{x} \Vert_1 = \sum_i \vert x_i\vert$ and $\Vert \b{x} \Vert_2 = \sqrt{\sum_i  x_i^2}$, respectively. The matrix $\ell_1$ norm is defined as $\Vert \b{X} \Vert_1 = \sum_i\sum_j \vert x_{ij}\vert$, where  $\vert \cdot \vert$ denotes the absolute value operator.  The Frobenius norm is defined as $\Vert \b{X} \Vert_F = \sqrt{\sum_i\sum_j x_{ij}^2}$, and the nuclear norm of $\b{X}$ (i.e., the sum of singular values of a matrix) is denoted by  $\Vert \b{X} \Vert_*$. $\b{X}^T$ is the transpose of $\b{X}$. If $\b{X}$ is a square matrix, $\b{X}^{-1}$ is its inverse, provided that the inverse matrix exists.

The warp function $\b{x}(\mathcal{W}(\b{p}))$ \big($\b{X}(\mathcal{W}(\b{p}))$\big) denotes the warping of 2D coordinates arranged as vector (matrix) by a warp parameter vector $\b{p} \in \mathbb{R}^v$, where $v$ is the number of the warping parameters, back to reference coordinate system. To simplify the notation $\b{x}(\b{p})$ \big($\b{X}(\b{p})$\big) will be used throughout the paper instead of $\b{x}(\mathcal{W}(\b{p}))$ \big($\b{X}(\mathcal{W}(\b{p}))$\big).

\section{Proposed method} \label{sec:proposed_method}

\subsection{Problem formulation}

In this paper, the goal is to recover the frontal view (i.e., $\b{L} \in \mathbb{R}^{m \times n}$)  of a warped facial image (i.e., $\b{X}(\b{p}) \in \mathbb{R}^{m \times n}$) which is possibly corrupted by sparse error of large magnitude. Such sparse errors indicate that  only a small fraction of the image pixels may be corrupted by non-Gaussian noise and occlusions. In particular, based on the observation that the frontal view of a face lies onto a low-rank subspace (please refer to Fig. \ref{fig:low_rank}), it can be expressed as a linear combination of a small number of orthonormal learned basis (i.e. $\b{U} =[\b{u}_1, \b{u}_2, \ldots, \b{u}_k] \in \mathbb{R}^{f \times k}$, $\b{U}^T\b{U} = \b{I}$) that span a generic clean frontal face subspace, that is $\b{L} = \sum_{i=1}^k \mathcal{R}_{m \times n}(\b{u}_i) c_i$. Therefore, a warped corrupted image is written as:
\begin{equation}\label{eq:init_min_prob}
\b{X}(\b{p}) = \b{L}  + \b{E} =  \sum_{i=1}^k \mathcal{R}_{m \times n}(\b{u}_i) c_i + \b{E},
\end{equation}
where $\b{E}$ is the sparse error matrix.

To match nicely the specifications of the frontal image and the sparse error one can find the low-rank frontal image, the linear combination coefficients, the increments of warp parameters,  and the error matrix by solving:
\begin{equation}\label{eq:relaxed_min_prob}
\begin{aligned}
& \underset{\b{L},\b{E},\b{c}, \Delta\b{p}}{\argmin} 
& & \Vert \b{L} \Vert_* + \lambda \Vert \b{E} \Vert_1  \\
& \text{s.t.} & &  \b{X}(\b{p}) = \b{L} + \b{E}, &  \b{L} = \sum_{i=1}^k \mathcal{R}_{m \times n}(\b{u}_i) c_i,
\end{aligned}
\end{equation}
where the nuclear norm $\Vert \b{L} \Vert_*$ and the $\ell_1$ norm $\Vert \b{E} \Vert_1$ are utilized to promote low-rank on $\b{L}$ and sparsity in $\b{E}$, while $\lambda$ is a positive parameter balancing the norms. The nuclear and $\ell_1$ norms are the closest convex surrogates to the natural criteria of rank \cite{fazel2002matrix} and cardinality \cite{donoho2004most} which are NP-hard in general to optimize \cite{natarajan1995sparse,Vandenberghe:96}. However, (\ref{eq:relaxed_min_prob}) is difficult to be solved due to the non-linearity of the constraint $\b{X}(\b{p}) = \b{L} + \b{E}$.  
 To remedy this, a first order Taylor linear approximation is applied on the vectorized form of the constrained: $\b{x}(\b{p}+\Delta\b{p}) \approx \b{x}(\b{p}) + \b{J}(\b{p})\Delta\b{p}$.
 where $\mbox{vec}(\b{X}(\b{p})) = \mbox{vec}(\b{L} + \b{E}) = \b{U}\b{c} + \b{e}= \b{x}(\b{p})$ and $\b{J}(\b{p})=\nabla \b{x}(\b{p}) \frac{\partial W}{\partial \b{p}}$ is the Jacobian matrix with the steepest descent images as its columns. Consequently, (\ref{eq:relaxed_min_prob}) is written as: 
\begin{equation}\label{eq:vec_relaxed_min_prob}
\begin{aligned}
& \underset{\b{L},\b{e},\b{c}, \b{p}}{\argmin} 
& & \Vert \b{L} \Vert_* + \lambda \Vert \b{E} \Vert_1  \\
& \text{s.t.} & &  h_1(\Delta \b{p}, \b{c}, \b{e})=0, & h_2(\b{L},\b{c}) = 0,
\end{aligned}
\end{equation}
where
$h_1(\Delta \b{p}, \b{c}, \b{e}) = \b{x}(\b{p}) + \b{J}(\b{p})\Delta\b{p} - \b{U}\b{c} -\b{e}$ and $h_2(\b{L},\b{c}) = \b{L} - \sum_{i=1}^k \mathcal{R}_{m \times n}(\b{u}_i) c_i$.

\subsection{Optimization}
To solve (\ref{eq:vec_relaxed_min_prob}), the \textit{augmented} Lagrangian \cite{bertsekas1982constrained} is introduced:
\begin{align}\label{eq:aug_lag_fun}
&\mathcal{L}(\b{L},\b{c},\Delta \b{p},\b{e},\b{a},\b{B}) = \Vert \b{L} \Vert_* + \lambda\Vert \b{e} \Vert_1 + \b{a}^T(h_1(\Delta \b{p}, \b{c}, \b{e}))\nonumber \\
&+ tr\left(\b{B}^T(h_2(\b{L},\b{c}))\right) + \frac{\mu}{2}\big(\Vert h_1(\Delta \b{p}, \b{c}, \b{e}) \Vert_2^2 + \Vert h_2(\b{L},\b{c}) \Vert_F^2 \big),
\end{align}
where $\b{a}$ and $\b{B}$ the Lagrange multipliers for the equality constraints in (\ref{eq:vec_relaxed_min_prob}) and $\mu > 0$ is a penalty parameter. By employing the Alternating Directions Method (ADM) of multipliers \cite{bertsekas1982constrained}, (\ref{eq:vec_relaxed_min_prob}) is solved by minimizing (\ref{eq:aug_lag_fun}) with respect to each variable in an alternating fashion and finally the Lagrange multipliers are updated at each iteration as outlined in Algorithm~\ref{alg:main_alg} its derivation is provided next. 

Let $t$ be the iteration index. Given the  $\b{L}_{t}, \b{c}_{t}, \Delta \b{p}_{t}, \b{e}_{t},\b{a}_{t}$, and $\b{B}_{t}$, the updates are computed by solving the following sub-problems:
\begin{align}
\b{L}_{t+1} =&\argmin_{\b{L}_{t}} \mathcal{L}(\b{L}_{t},\b{c}_{t},\Delta \b{p}_{t},\b{e}_{t},\b{a}_{t},\b{B}_{t}) \label{eg:sub_L}\\
\b{c}_{i,t+1} =& \argmin_{\b{c}_{i,t}} \mathcal{L}(\b{L}_{t+1},\b{c}_{t},\Delta \b{p}_{t},\b{e}_{t},\b{a}_{t},\b{B}_{t}) \label{eg:sub_c}\\ 
\Delta\b{p}_{t+1} =& \argmin_{\Delta\b{p}_{t}} \mathcal{L}(\b{L}_{t+1},\b{c}_{t+1},\Delta \b{p}_{t},\b{e}_{t},\b{a}_{t},\b{B}_{t}) \label{eg:sub_dp}\\
\b{e}_{t+1} =& \argmin_{\b{e}_{t}} \mathcal{L}(\b{L}_{t+1},\b{c}_{t+1},\Delta \b{p}_{t+1},\b{e}_{t},\b{a}_{t},\b{B}_{t}) \label{eg:sub_e}
\end{align}

\subsubsection*{\textbf{Solving for $\b{L}$}}
Fixing the $\b{c}_{t}, \Delta \b{p}_{t}, \b{e}_{t},\b{a}_{t}$, and $\b{B}_{t}$, sub-problem (\ref{eg:sub_L}) is reduced to:
\begin{equation}
\label{eq:red_L_sub}
\argmin_{\b{L}_{t}} \Vert \b{L} \Vert_* + tr(\b{B}^T(h_2(\b{L},\b{c}))) + \frac{\mu}{2}\Vert h_2(\b{L},\b{c}) \Vert_F^2 .
\end{equation}
The nuclear norm regularized least squared problem (\ref{eq:red_L_sub}) has the following closed-form solution:
\begin{equation}
\b{L}_{t+1} = \mathcal{D}_{\frac{1}{\mu_{t}}}\left[ \sum_{i=1}^k R_{m\times n} (\b{u}_i)c_{i_{t}} - \b{B}_{t} / \mu_{t}  \right].
\label{eq:sub_L_solution}
\end{equation}	
The singular value thresholding (SVT) operator is defined for any matrix $\b{Q}$ with $\b{Q}=\b{U} \b{\Sigma} \b{V}^T$ as $\mathcal{D}_{\tau}[\b{Q}] = \b{U}\b{\mathcal{S}}_{\tau}\b{V}^T$ \cite{cai2010singular}, with $\b{\mathcal{S}}_{\tau}[\sigma ] = $sgn$(\sigma)\max (|\sigma |-\tau,0)$ being the (element-wise) shrinkage operator \cite{candes2011robust}.

\begin{algorithm}
	\label{alg:main_alg}
	\SetAlgoLined
	\KwData{Test image $\b{X}$, initial deformation parameters $\b{p}$, clean face subspace $	\b{U}$, and the parameter  $\lambda$}
	\KwResult{The low-rank clean image $\b{L}$, the sparse error $\b{e}$, the coefficient vector 	$\b{c}$, and the increments of the deformation parameters $\Delta \b{p}$.}
	\While{not converged}{
		Warp and normalize the image\;
		Compute the Jacobian matrix\;
		Initialize: $\b{L}_{0} = \b{0}, \b{e}_{0} = \b{0}$, $ \b{c}_{0} = \b{0}$, $\b{a}_{0} = 	\b{0}$, $\b{B}_{0} = \b{0}$, $\mu_{0} >0$ , $\rho >1$ \; 
		\While{not converged}{
			Update $\b{L}_{t+1}$ by (\ref{eq:sub_L_solution})\;
			Update $\b{c}_{i,t+1}$ by (\ref{eq:sub_c_solution})\;
			Update $\Delta \b{p}_{t+1}$ by (\ref{eq:sub_dp_solution})\;
			Update $\b{e}_{t+1}$ by (\ref{eq:sub_e_solution})\;
			Update the Lagrange multipliers by
			$\b{a}_{t+1} \text{ } = \b{a}_{t} + \mu_{t}(h_1(\Delta\b{p}_{t+1},\b{c}_{t+1},\b{e}		_{t+1}))$\; 
			$\b{B}_{t+1} = \b{B}_{t} + \mu_{t}((h_2(\b{L}_{t+1}, \b{c}_{t+1}))$\;
			Update $\mu_{t+1}$  by $\mu_{t+1} \leftarrow \min(\rho\cdot\mu_{t},10^{10})$\;
			Check convergence conditions (\ref{eq:cont_1}) and (\ref{eq:cont_2})\;
		
$t \leftarrow t + 1$\;
		}
		$\b{p} \leftarrow \b{p} + \Delta \b{p}$\;
	}
\caption{Solving (\ref{eq:aug_lag_fun}) by the ADM method}
\end{algorithm}

\subsubsection*{\textbf{Solving for $\b{c}$}}
Fixing the other variables, sub-problem (\ref{eg:sub_c}) is reduced to:
\begin{align}
\label{eq:sub_c_2}
\argmin_{\b{c}_{t}}\text{  } &\b{a}_{t}^T(h_1(\Delta \b{p}_{t}, \b{c}_{t}, \b{e}_{t}))+tr\left(\b{B}^T(h_2(\b{L}_{t+1},\b{c}_{t}))\right)\nonumber \\+\frac{\mu_{t}}{2}&\big(\Vert h_1(\Delta \b{p}_{t}, \b{c}_{t}, \b{e}_{t}) \Vert_2^2 + \Vert h_2(\b{L}_{t+1},\b{c}_{t}) \Vert_F^2 \big).
\end{align}
(\ref{eq:sub_c_2}) is a quadratic problem which for each $c_i, i \in \{1,\dots k\}$ admits a closed form solution given by:
\begin{equation}
\b{c}_{i,t+1} = \frac{\b{a}_{t}^T\b{u}_i+\tr(\b{B}_{t}^TR_{m \times n}(\b{u}_i))}{2\mu_t} + \frac{\hat{\b{x}}^T\b{u}_i + \tr(\b{L}_{t+1}^TR_{m \times n}(\b{u}_i))}{2},
\label{eq:sub_c_solution}
\end{equation}
where $\hat{\b{x}} = \b{x}(\b{p}) + \b{J}(\b{p})\Delta\b{p}_{t}-\b{e}_t$.

\subsubsection*{\textbf{Solving for $\Delta \b{p}$}}
Sub-problem (\ref{eg:sub_dp}) is written as:
\begin{align}
\label{eq:sub_dp2}
\argmin_{\Delta\b{p}_{t}} \text{  } \b{a}_{t}^T(h_1(\Delta \b{p}_{t}, \b{c}_{t+1}, \b{e}_{t}))+ \frac{\mu}{2} \Vert h_1(\Delta \b{p}_{t}, \b{c}_{t+1}, \b{e}_{t}) \Vert_2^2.
\end{align}
By exploiting the fact that $\b{U}$ is orthonormal, each part of (\ref{eq:sub_dp2}) is decomposed into the term projected in the $\b{U}\b{U}^T$ and the term projected into orthogonal complement $\b{I} - \b{U}\b{U}^T$. The update of the $\Delta \b{p}$ is obtained by minimizing the projected into orthogonal complement part, i.e.,
\begin{align}
\label{eq:sub_dp_3}
\argmin_{\Delta\b{p}_{t}} \text{  }
 & \b{a}_{t}^T(\b{I}-\b{U}\b{U}^T)(\b{x}(\b{p}_t) + \b{J}(\b{p})\Delta\b{p}_{t} - \b{e}_{t+1}) \nonumber \\
& + \frac{\mu}{2} \Vert \b{x}(\b{p}) + \b{J}(\b{p}_t)\Delta \b{p}_{t} - \b{e}_{t+1}\Vert_{2,\b{I}-\b{U}\b{U}^T}^2.
\end{align}
The solution of (\ref{eq:sub_dp_3}) is given by:
\begin{equation}
\label{eq:sub_dp_solution}
\Delta\b{p}_{t+1} = -\big(\tilde{\b{J}}(\b{p})^T\tilde{\b{J}}(\b{p})\big)^{-1}\tilde{\b{J}}(\b{p})^T \big(\b{x}(\b{p})-\b{e}_{t+1}\big),
\end{equation}
where $\tilde{\b{J}}(\b{p})$ is the projected Jacobian in $\b{I}-\b{U}\b{U}^T$. To calculate efficiently the term $\tilde{\b{J}}^T\tilde{\b{J}}$ the following formulation is used: $\tilde{\b{J}}^T\tilde{\b{J}} = \b{J}^T (\b{I}-\b{U}\b{U})^T \b{J}^T =(\b{U}^T \b{J})^T (\b{U}^T \b{J})$.

\subsubsection*{\textbf{Solving for} $\b{e}$}
Using $\b{L}_{t+1}, \b{c}_{t+1}, \Delta \b{p}_{t+1}$ (\ref{eg:sub_e}) is written as:
\begin{align}
\label{eq:red_E_sub}
\argmin_{\b{e}_{t}} \quad &\lambda\Vert \b{e} \Vert_1 + \b{a}_{t}^T(h_1(\Delta \b{p}_{t+1}, \b{c}_{t+1}, \b{e}_{t})) \nonumber \\ 
&+ \frac{\mu}{2} \Vert h_1(\Delta \b{p}_{t+1}, \b{c}_{t+1}, \b{e}_{t}) \Vert_2^2.
\end{align}
The closed-form solution of (\ref{eq:red_E_sub}) is given by applying element-wise the shrinkage operator onto: $\b{x}(\b{p}_t) + \b{J}(\b{p})\Delta\b{p}_{t+1} - \b{U}\b{c}_{t+1} + \b{a}_{t}/ \mu_{t}$, namely
\begin{equation}
\b{e}_{t+1} = \mathcal{S}_{\frac{\lambda}{\mu_{t}}}\left[ \b{x}(\b{p}_t) + \b{J}(\b{p})\Delta\b{p}_{t} - \b{U}\b{c}_{t} + \b{a}_{t}/ \mu_{t} \right].
\label{eq:sub_e_solution}
\end{equation}

\subsubsection*{\textbf{Convergence criteria}}
The inner loop of the Algorithm~\ref{alg:main_alg} terminates when
\begin{equation}
\label{eq:cont_1}
\max\left(\Vert\b{e}_{t} - \b{e}_{t-1}  \Vert_2/\Vert \b{x}(\b{p}) \Vert_2,\Vert\b{L}_{t} - \b{L}_{t-1} \Vert_F/\Vert \b{x}(\b{p}) \Vert_2\right)
\leq \epsilon_2,
\end{equation}
and

\begin{align}
\label{eq:cont_2}
\max(\Vert h_1&(\Delta \b{p}_{t+1}, \b{c}_{t+1}, \b{e}_{t+1}) \Vert_2/\Vert \b{x}(\b{p}) \Vert_2, \nonumber\\ &\Vert h_2(\b{L}_{t+1},\b{c}_{t+1}) \Vert_F/\Vert \b{x}(\b{p}) \Vert_2)
\leq \epsilon_3.
\end{align}

The Algorithm~\ref{alg:main_alg} terminates when the change of the $\Vert \b{L} \Vert_* + \lambda \Vert \b{E} \Vert_1$ between two successive iterations is smaller than the threshold $\epsilon_1$ or the maximum number of the outers' loop iterations is reached. The dominant cost of each iteration of Algorithm~\ref{alg:main_alg} is that of the SVD algorithm involved in the  computation of the SVT operator in update of $\b{L}$. Consequently, the computational complexity of Algorithm~\ref{alg:main_alg} is $\mathcal{O}(T(min(m, n)^3 + n^2k))$, where $T$ is the total number of iterations until convergence.

\begin{figure}[!h]
\begin{center}
\includegraphics[scale=0.2]{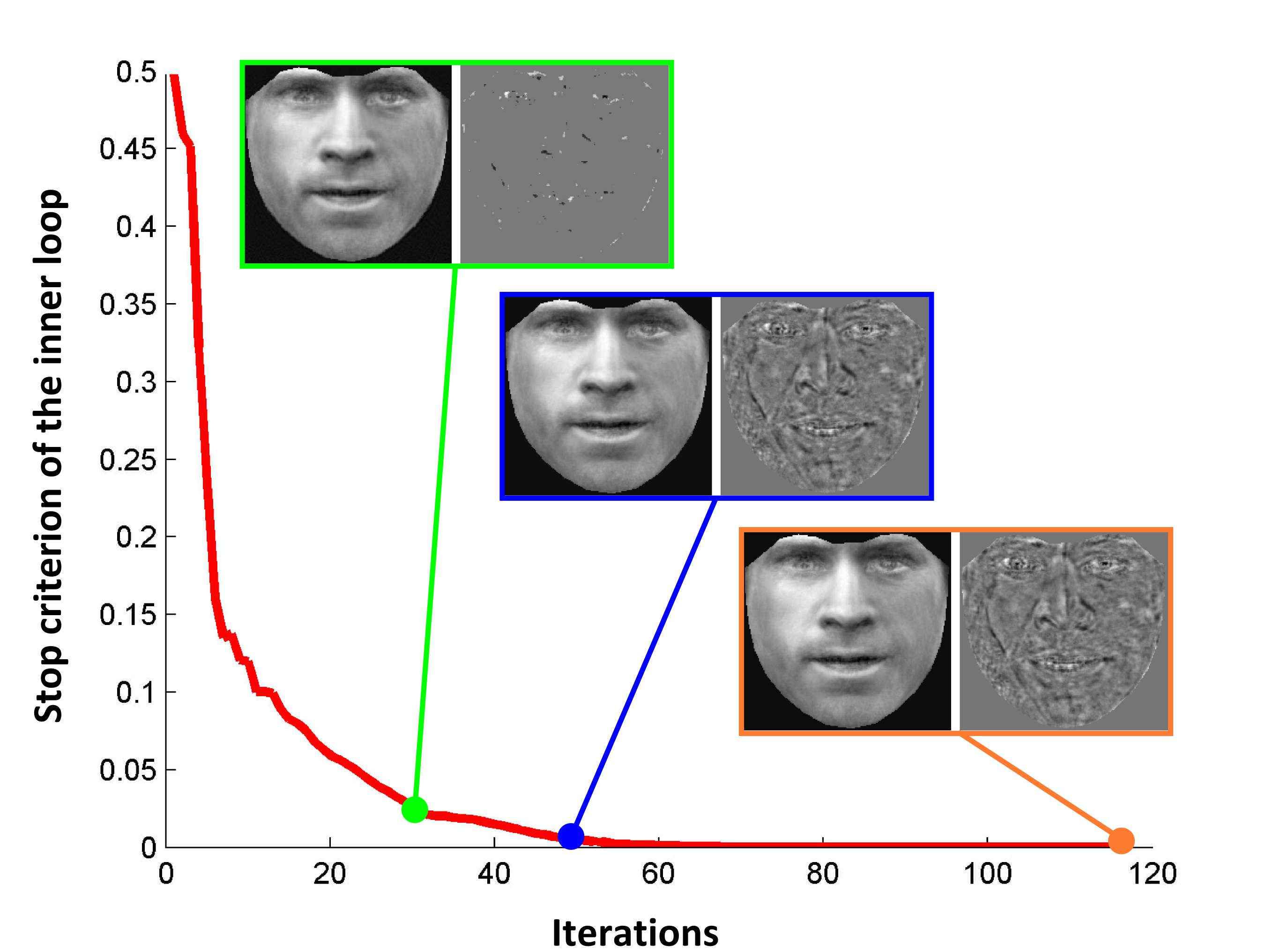}
\end{center}
\caption{The convergence curve of the Algorithm's \ref{alg:main_alg} inner loop.}
\label{fig:convergence}
\end{figure}

In Fig. \ref{fig:convergence}, the convergence of the inner loop of Algorithm 1 is depicted. The low-rank and error images produced after $30$, $50$ and $117$ iterations, respectively, are also shown.

\section{Experimental Results} \label{sec:experiments}

The performance of the FAR is assessed in: a) frontal face reconstruction, b) landmark localization, and c) pose invariant face recognition and verification, by conducting experiments in $6$ facial image databases, which are described briefly next.


\subsection{Data description}\label{sec:data_description}

\textbf{LFPW}: The Labeled Faces Parts in-the-wild (LFPW) \cite{lfpwDB} database contains images downloaded from the internet (i.e., gooogle.com, flickr.com etc), images exhibiting multiple variations such as pose, expression, illumination, and occlusions. Since only the URLs of images were provided, $811$ out of the $1,132$ training images and $224$ out of the $300$ test images were downloaded.

\textbf{HELEN}: The HELEN \cite{helenDB} database consists of $2,300$ images downloaded from Flickr web service, containing a broad range of appearance variation, including pose, lighting, expression, occlusion, and individual differences. The size of the face in each of the images was approximately $500 \times 500$ pixels.
 
\textbf{AFW}: The Annotated Faces in-the-wild (AFW) \cite{ramanan} database consists of $250$ images with $468$ faces. That is more than one faces are annotated in each image. The images exhibit similar variations with those in the LFPW and HELEN databases.

\textbf{FERET}: The Facial Recognition Technology (FERET) \cite{feretDB} database consists of $14,051$ images of $200$ different subjects. All images capture the same `Neutral' expression for $9$ different poses under different illuminations, where each subject also has an additional image with a random facial expression.

\textbf{MultiPIE}: The CMU Multi Pose Illumination and Expression (MultiPIE) \cite{multipieDB} database consists of approximately $750,000$ images from $337$ subjects, captured under  $6$  different expressions, $15$ poses, and $19$ illuminations. 

\textbf{LFW}: The Labeled Faces in the Wild (LFW) \cite{lfwDB} database contains 13,233 images of 5,749 people downloaded from the Web and is designed as a benchmark for the problem of unconstrained automatic face verification. All images are characterized by the existence of large pose, expression and occlusion variations.

\subsection{Experimental setup}

In all experiments, the orthonormal clean face subspace $\b{U}$ was constructed by employing only frontal view without occlusions face images. In total $500$ frontal images ($217$ from the training set of the LFPW and $283$ from the training set of the HELEN databases) were selected to build the bases $\b{U}$. The frontal images were warped in a common frame ($185 \times 193$ pixels) by using a piece-wise affine motion model and subsequently the PCA was applied on the warped shape-free textures. The first $k=450$ eigen-images were kept. Unless otherwise stated, throughout the experiments, the same $\b{U}$ was used and the  parameters of the Algorithm \ref{alg:main_alg} were fixed: $\lambda = 0.3$, $\rho = 1.1$, $\mu_0 = 10^{-6}$, $\epsilon_1=10^{-3}$, $\epsilon_2=10^{-5}$, and $\epsilon_3=10^{-7}$. 

\subsection{Frontal face reconstruction}\label{subsec:frontal_rec}

Next, the ability of the FAR to reconstruct frontal faces from non-frontal images of unseen subjects is investigated by using two unseen subjects taken from MultiPIE and from FERET databases and $5$ in-the-wild images.

\renewcommand\tabcolsep{0pt}
\begin{figure}[ht]
\begin{center}
\begin{tabular}{ccccccc}
\begin{subfigure}{\includegraphics[scale=0.12]{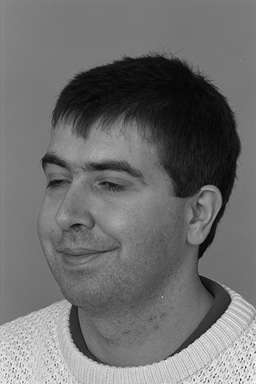}}\end{subfigure}&
\begin{subfigure}{\includegraphics[scale=0.12]{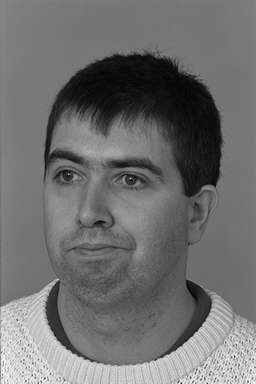}}\end{subfigure}&
\begin{subfigure}{\includegraphics[scale=0.12]{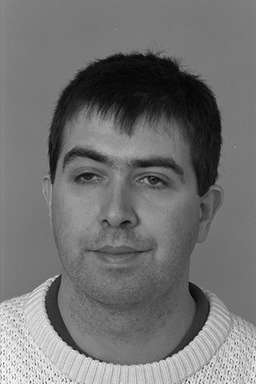}}\end{subfigure}&
\begin{subfigure}{\includegraphics[scale=0.12]{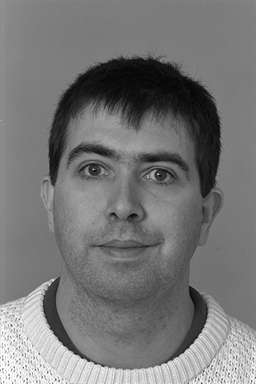}}\end{subfigure}&
\begin{subfigure}{\includegraphics[scale=0.12]{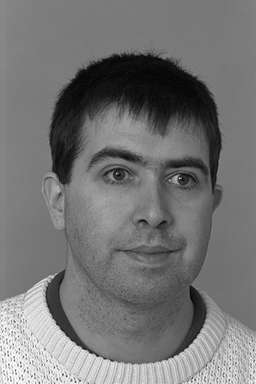}}\end{subfigure}&
\begin{subfigure}{\includegraphics[scale=0.12]{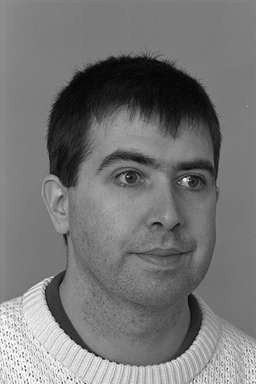}}\end{subfigure}&
\begin{subfigure}{\includegraphics[scale=0.12]{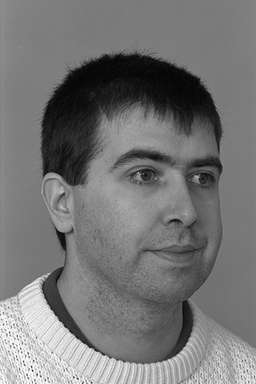}}\end{subfigure} \\
\vspace{-0.76cm} \\
\begin{subfigure}{\includegraphics[scale=0.05]{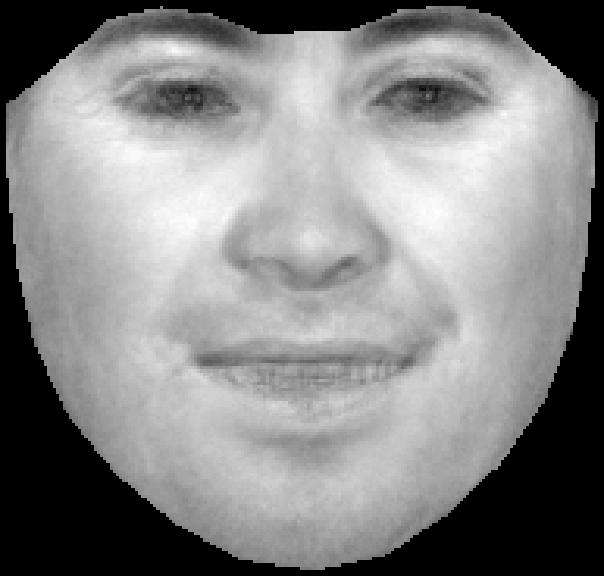}}\end{subfigure}&
\begin{subfigure}{\includegraphics[scale=0.05]{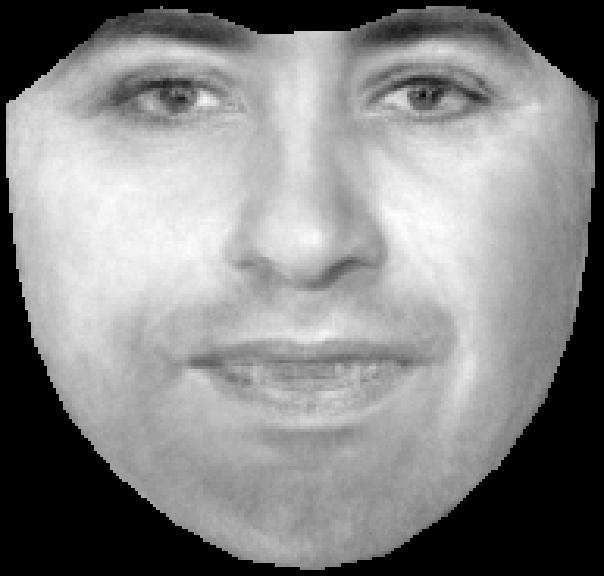}}\end{subfigure}&
\begin{subfigure}{\includegraphics[scale=0.05]{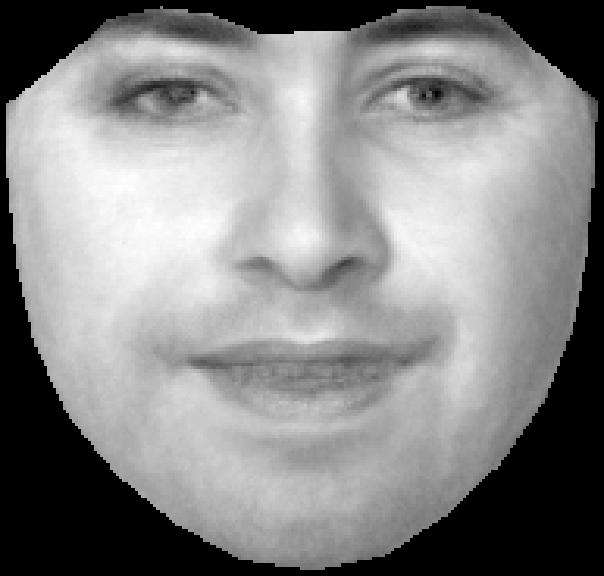}}\end{subfigure}&
\begin{subfigure}{\includegraphics[scale=0.05]{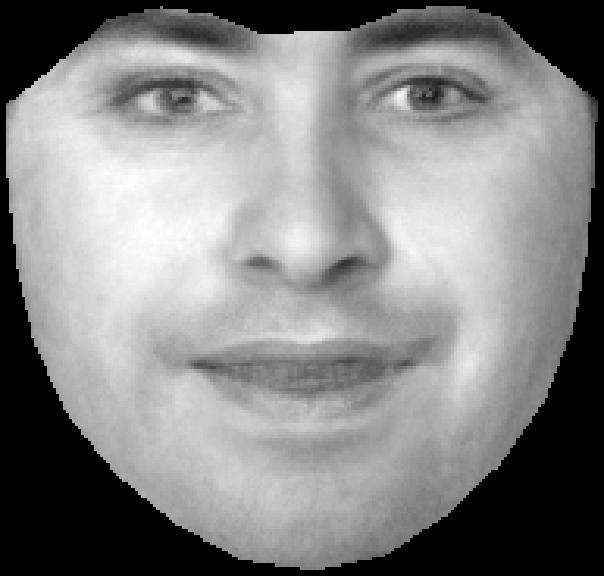}}\end{subfigure}&
\begin{subfigure}{\includegraphics[scale=0.05]{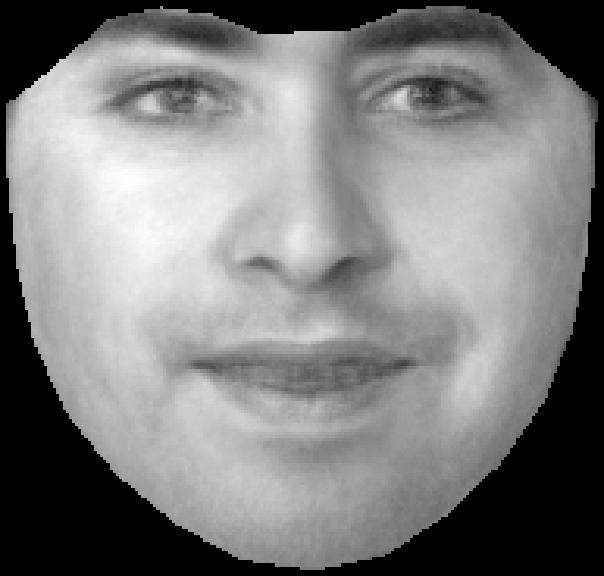}}\end{subfigure}&
\begin{subfigure}{\includegraphics[scale=0.05]{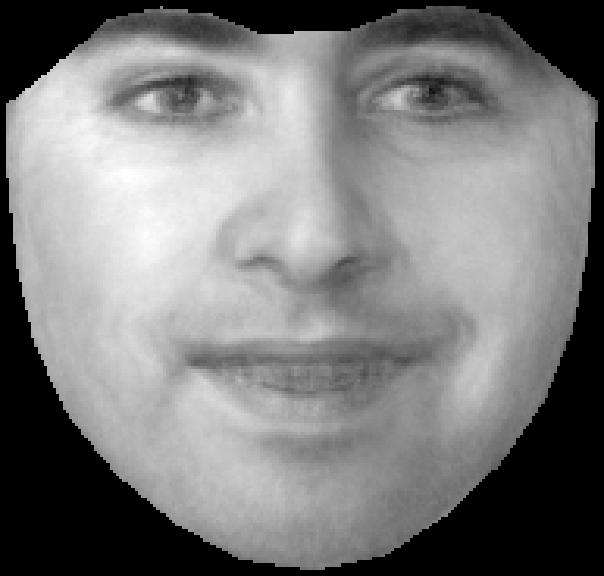}}\end{subfigure}&
\begin{subfigure}{\includegraphics[scale=0.05]{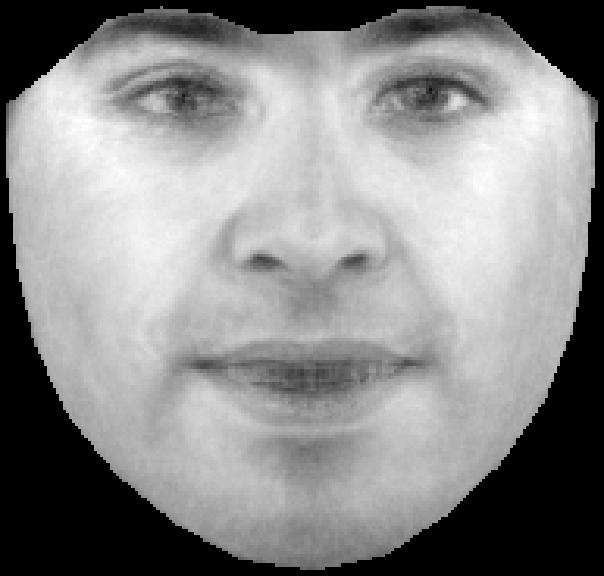}}\end{subfigure}
\end{tabular}
\newline
FERET
\vspace{-0.7cm}
\end{center}
\end{figure}

\begin{figure}[ht]
\begin{center}
\begin{tabular}{cccccc}
\begin{subfigure}{\includegraphics[scale=0.065]{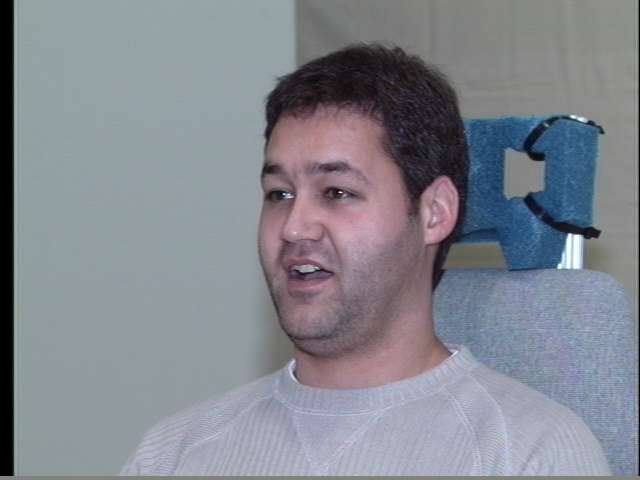}}\end{subfigure}&
\begin{subfigure}{\includegraphics[scale=0.065]{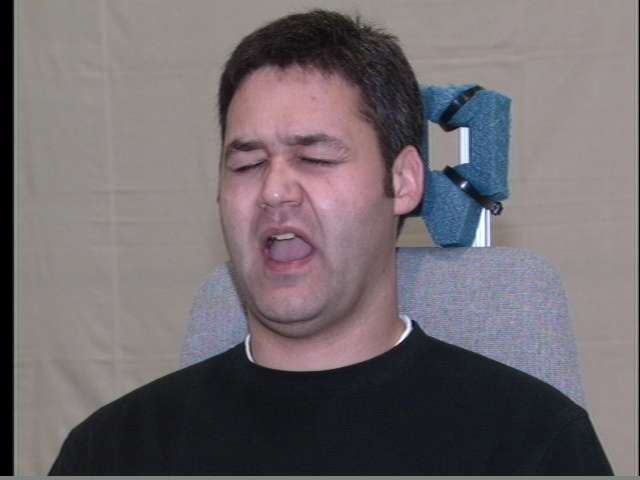}}\end{subfigure}&
\begin{subfigure}{\includegraphics[scale=0.065]{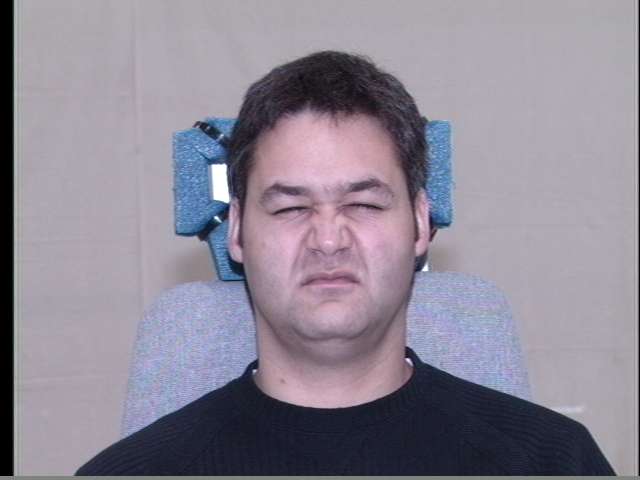}}\end{subfigure}&
\begin{subfigure}{\includegraphics[scale=0.065]{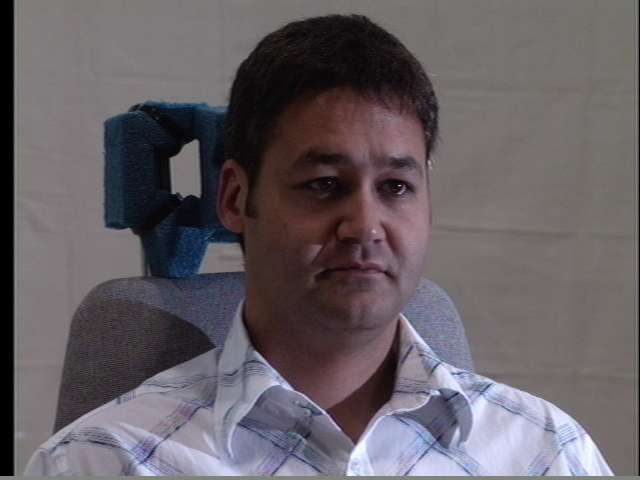}}\end{subfigure}&
\begin{subfigure}{\includegraphics[scale=0.065]{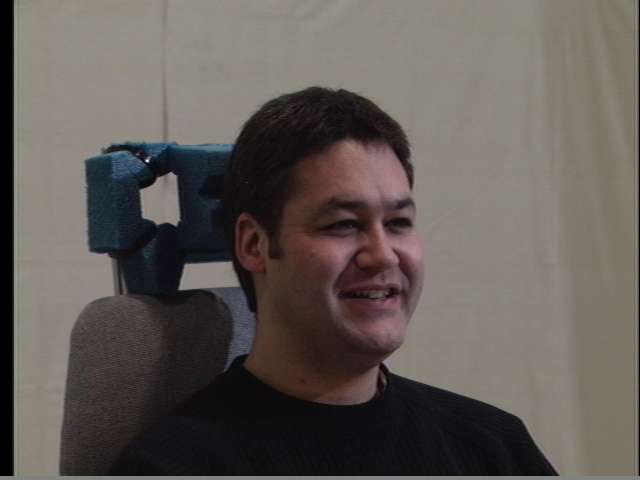}}\end{subfigure} \\
\vspace{-0.75cm} \\
\begin{subfigure}{\includegraphics[scale=0.069]{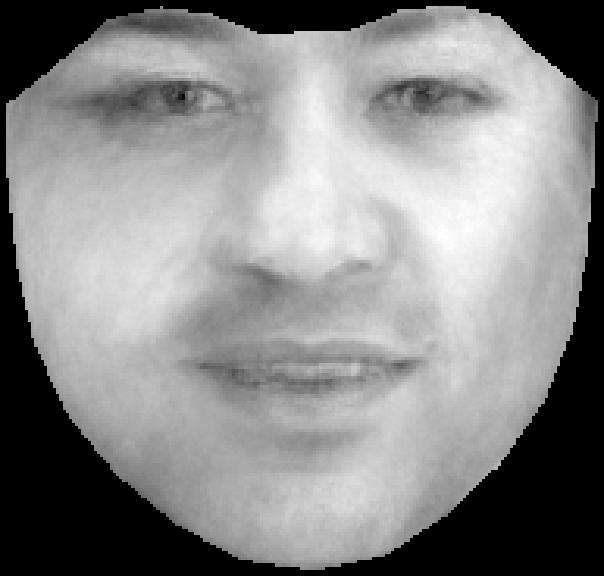}}\end{subfigure}&
\begin{subfigure}{\includegraphics[scale=0.069]{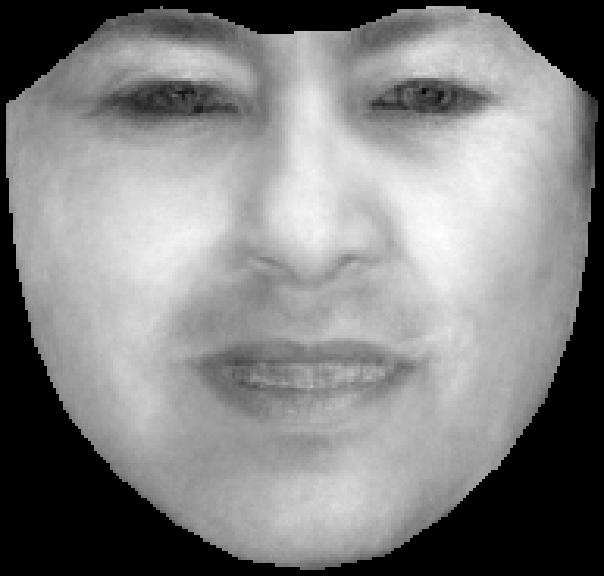}}\end{subfigure}&
\begin{subfigure}{\includegraphics[scale=0.069]{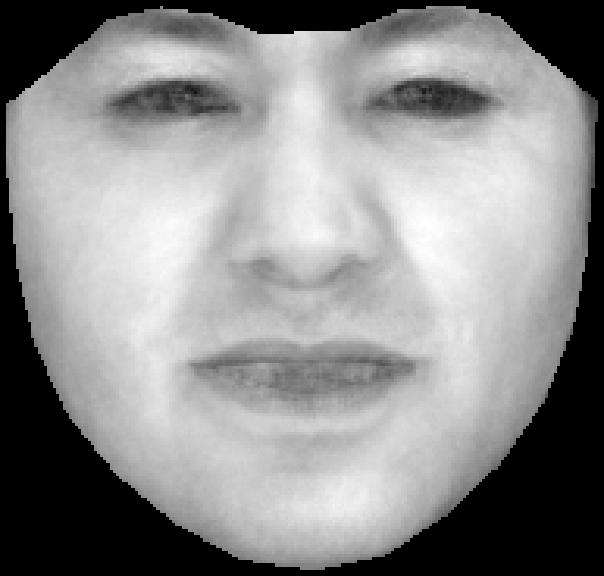}}\end{subfigure}&
\begin{subfigure}{\includegraphics[scale=0.069]{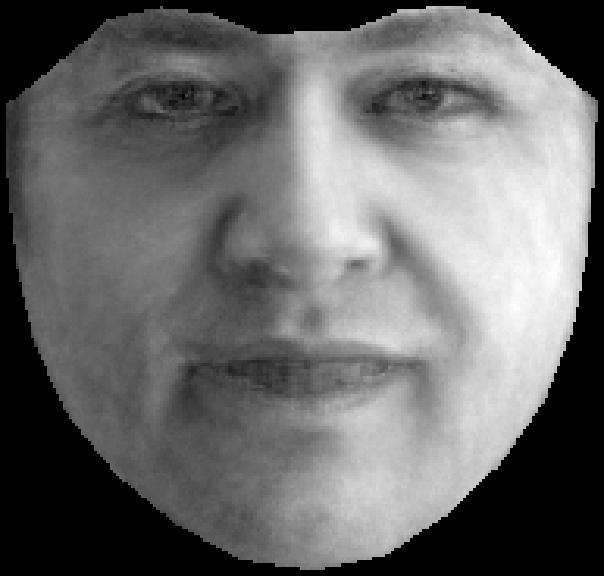}}\end{subfigure}&
\begin{subfigure}{\includegraphics[scale=0.069]{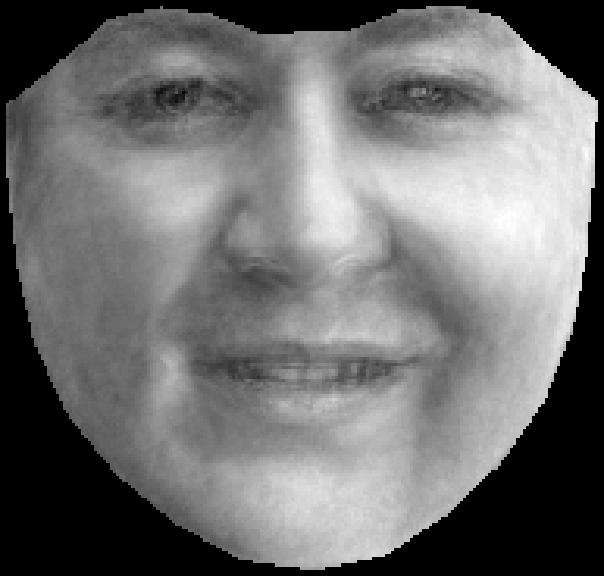}}\end{subfigure} \\
\end{tabular}
\newline
MultiPIE
\vspace{-0.7cm}
\end{center}
\end{figure}

\begin{figure}[ht]
\begin{center}
\begin{tabular}{ccccc}
\begin{subfigure}{\includegraphics[scale=0.105]{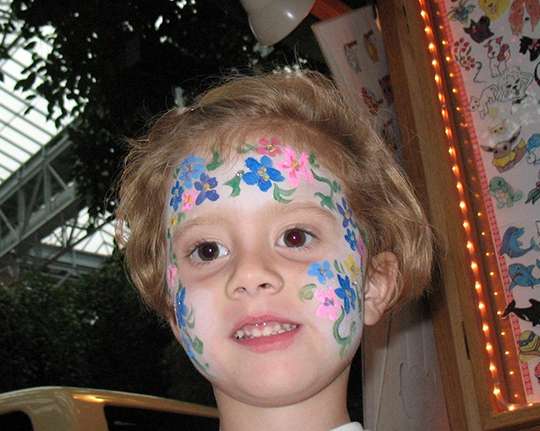}}\end{subfigure}&
\begin{subfigure}{\includegraphics[scale=0.215]{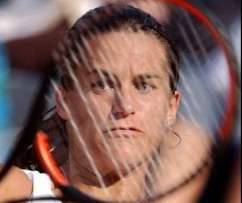}}\end{subfigure}&
\begin{subfigure}{\includegraphics[scale=0.083]{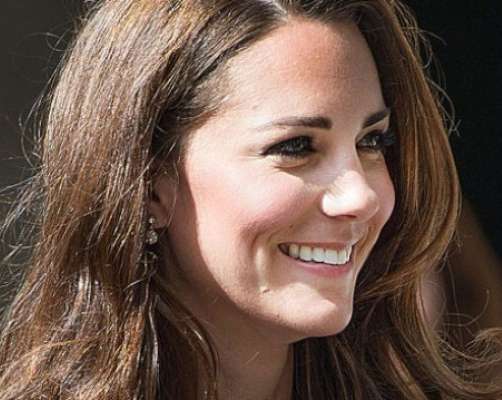}}\end{subfigure}&
\begin{subfigure}{\includegraphics[scale=0.085]{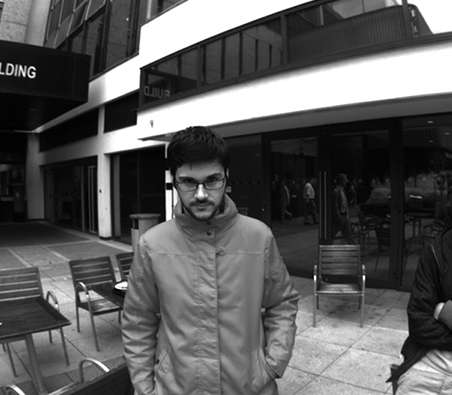}}\end{subfigure}&
\begin{subfigure}{\includegraphics[scale=0.19]{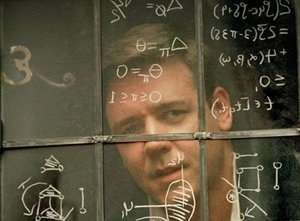}}\end{subfigure} \\
\vspace{-0.75cm} \\
\begin{subfigure}{\includegraphics[scale=0.067]{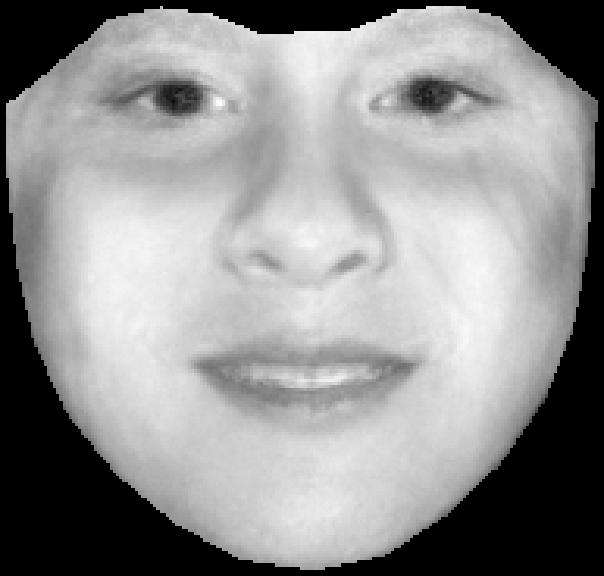}}\end{subfigure}&
\begin{subfigure}{\includegraphics[scale=0.067]{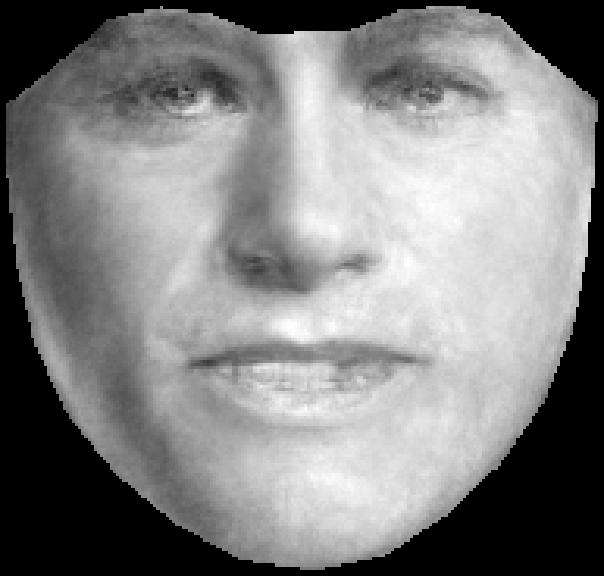}}\end{subfigure}&
\begin{subfigure}{\includegraphics[scale=0.067]{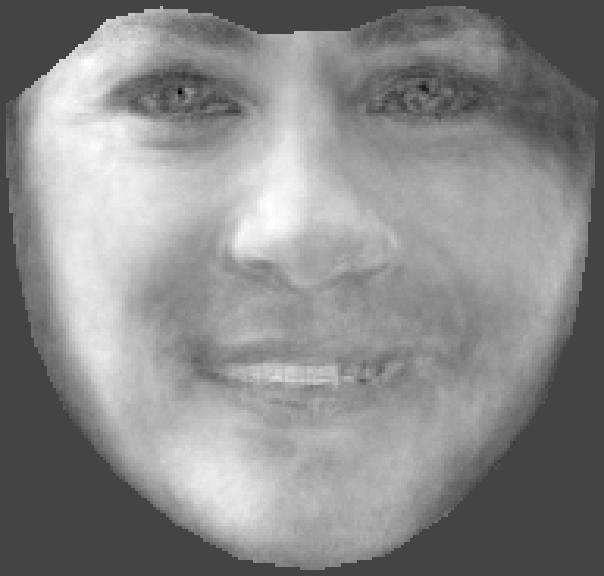}}\end{subfigure}&
\begin{subfigure}{\includegraphics[scale=0.067]{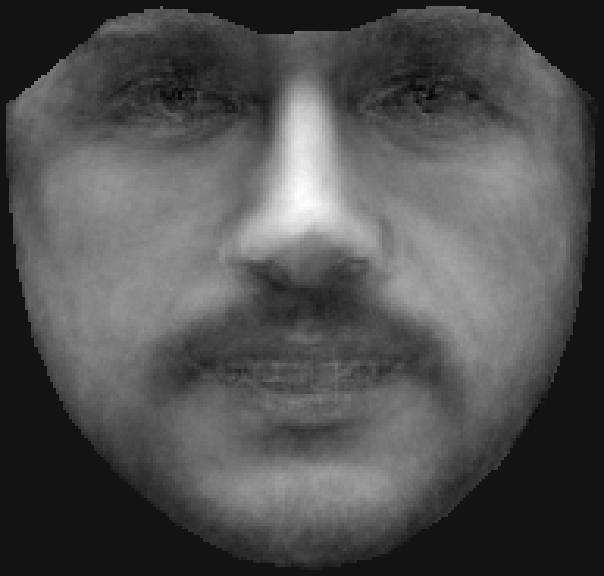}}\end{subfigure}&
\begin{subfigure}{\includegraphics[scale=0.067]{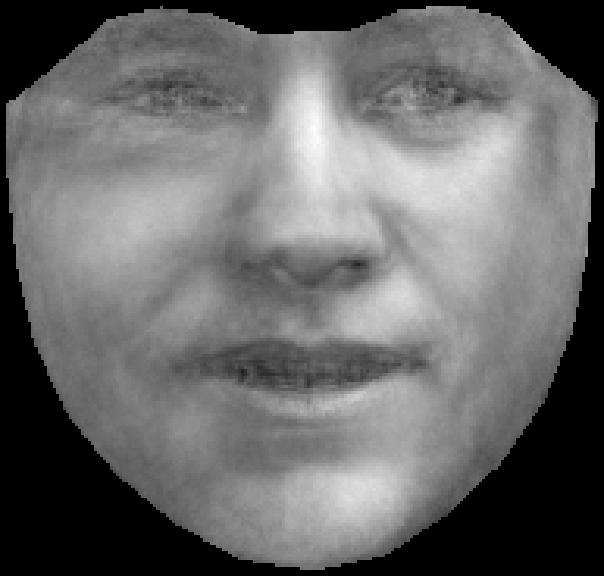}}\end{subfigure} \\
\end{tabular}
\newline
In-the-Wild
\end{center}
\caption{Reconstructed frontal images of unseen subjects under controlled and in-the-wild conditions.}
\label{fig:frontal_rec}
\end{figure}

\begin{figure*}[!ht]
\begin{center}
\renewcommand\tabcolsep{0pt}
\begin{tabular}{cccc}
\begin{subfigure}{\includegraphics[width=0.25\textwidth]{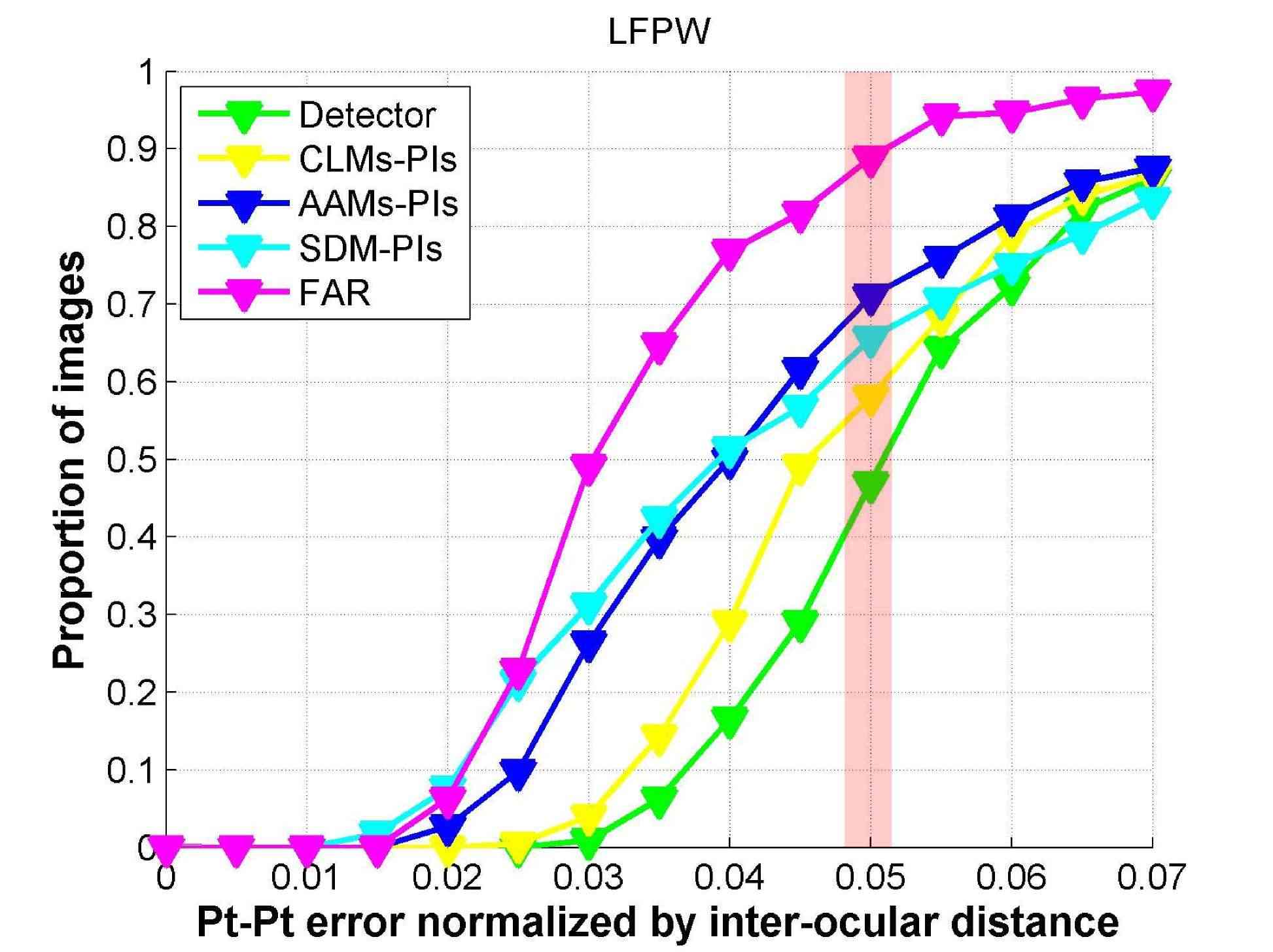}}\end{subfigure} &
\begin{subfigure}{\includegraphics[width=0.25\textwidth]{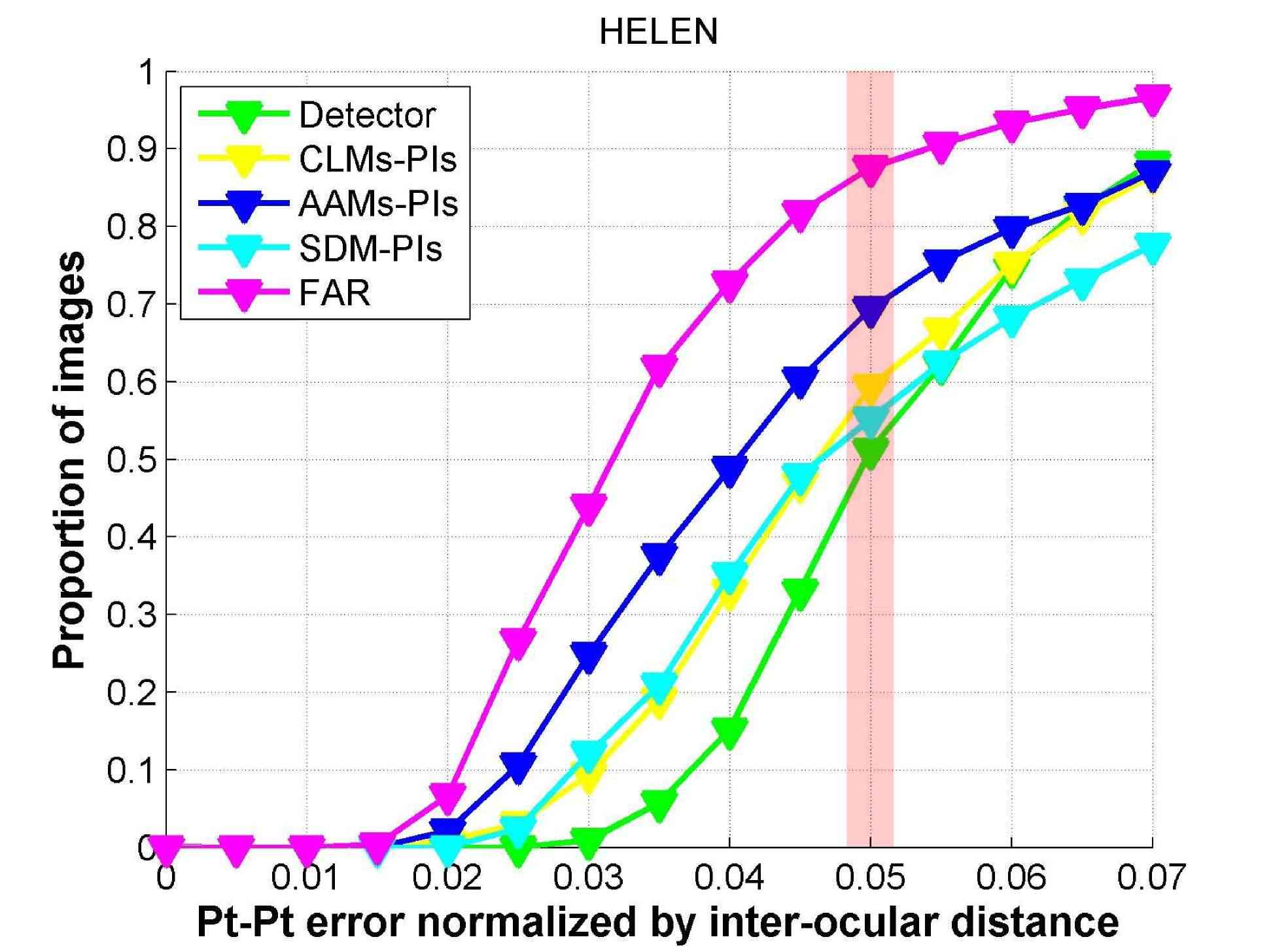}}\end{subfigure} &
\begin{subfigure}{\includegraphics[width=0.25\textwidth]{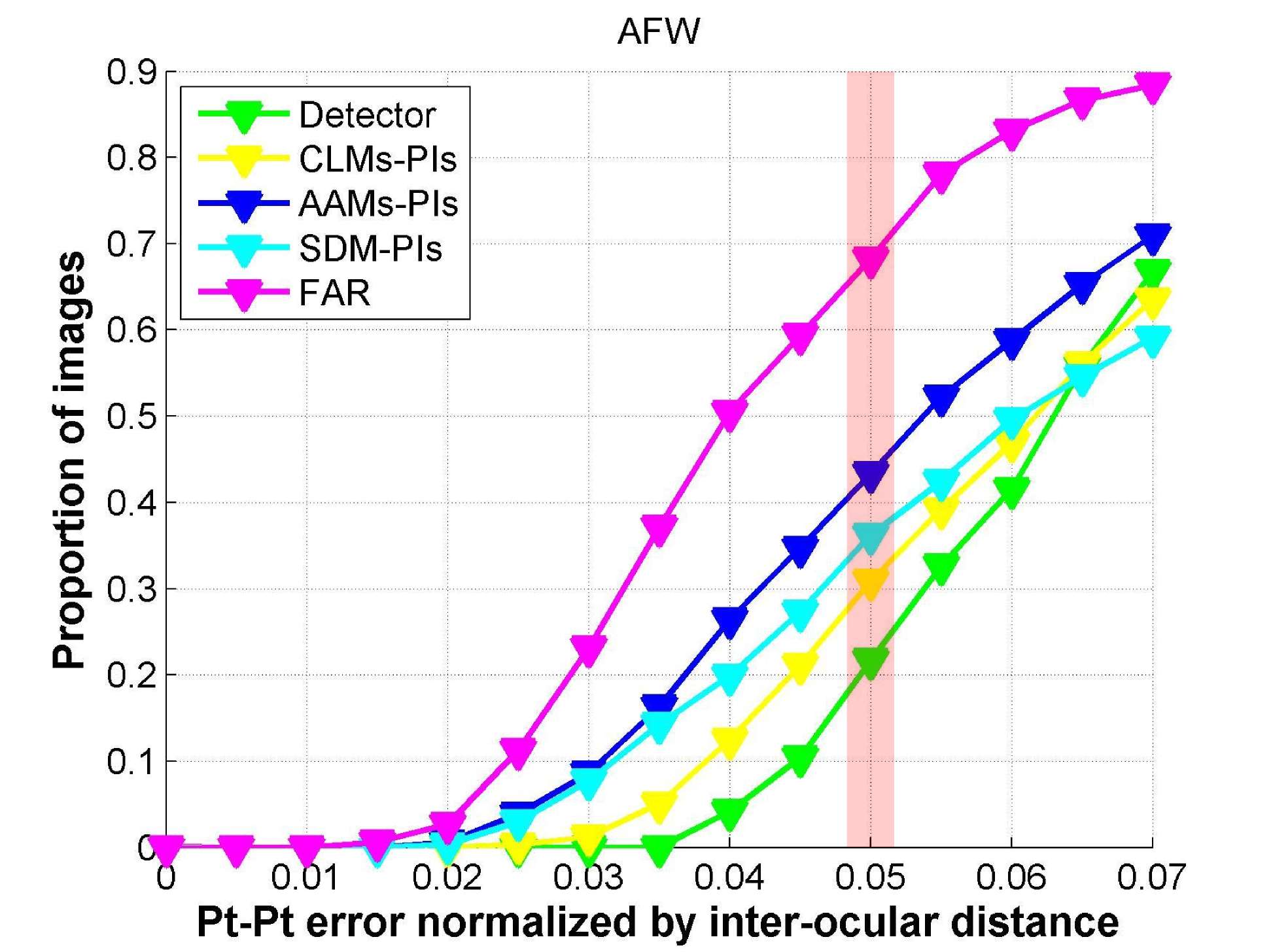}}\end{subfigure} &
\end{tabular}
\end{center}
\caption{Cumulative error distribution curves produced by the CLMs-PIs, the AAMS-PIs, the SDM-PIs, and the FAR for the LFPW, the HELEN, and the AFW.}
\label{fig:gfa_results}
\end{figure*}

\begin{figure*}[!ht]
\begin{center}
\renewcommand\tabcolsep{0pt}
\begin{tabular}{cccc}
\begin{subfigure}{\includegraphics[width=0.24\textwidth]{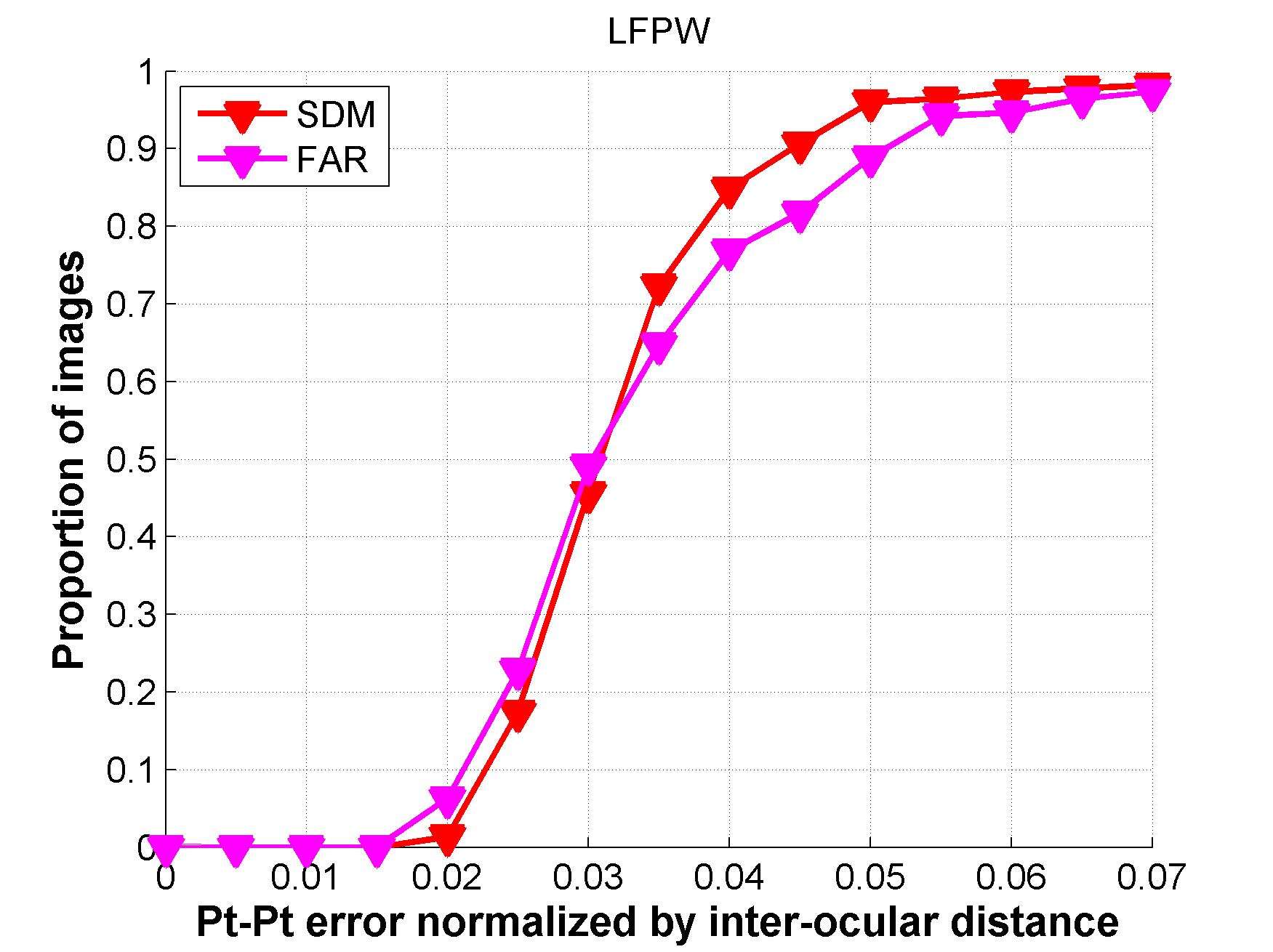}}\end{subfigure} &
\begin{subfigure}{\includegraphics[width=0.24\textwidth]{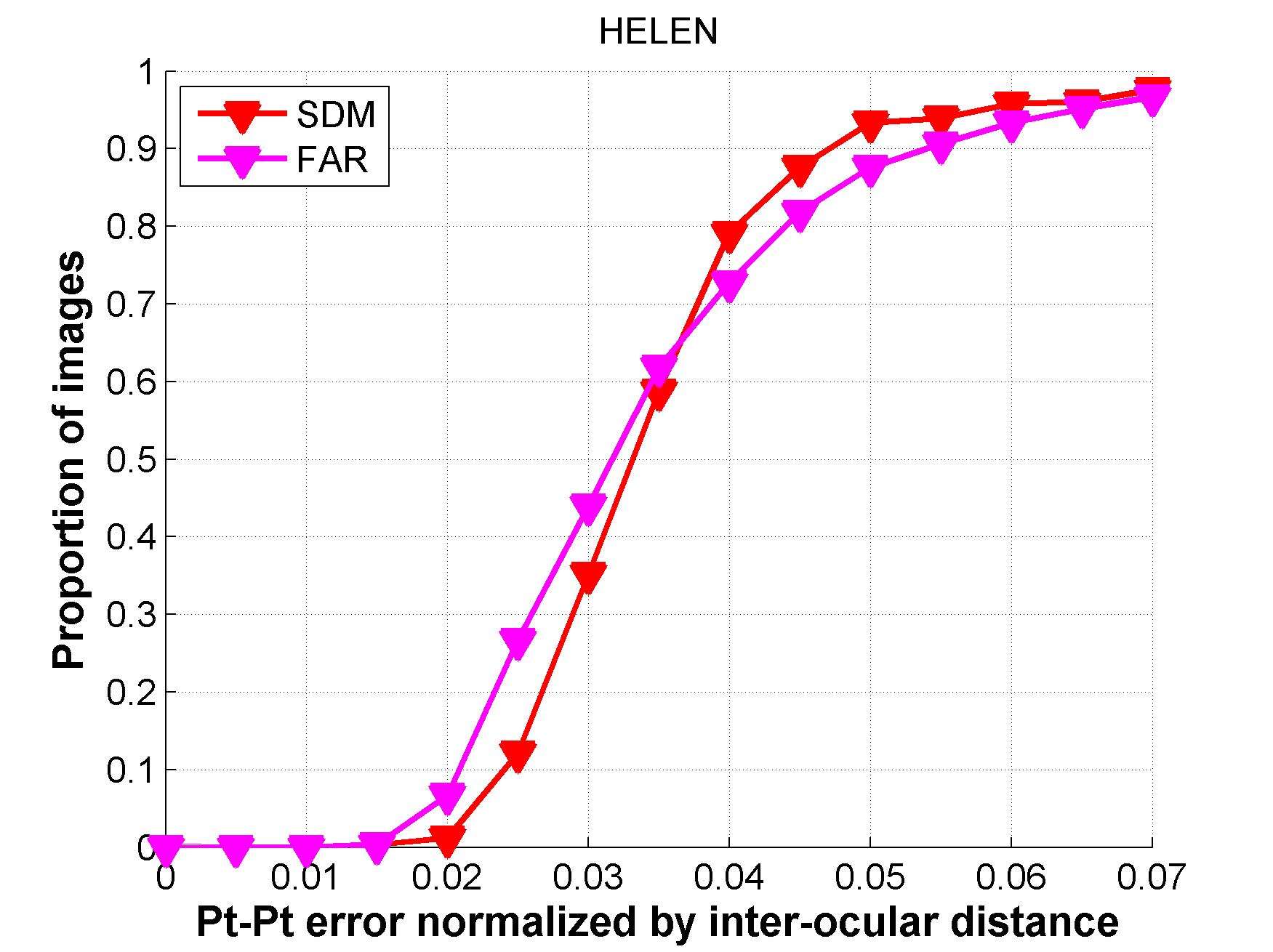}}\end{subfigure} &
\begin{subfigure}{\includegraphics[width=0.24\textwidth]{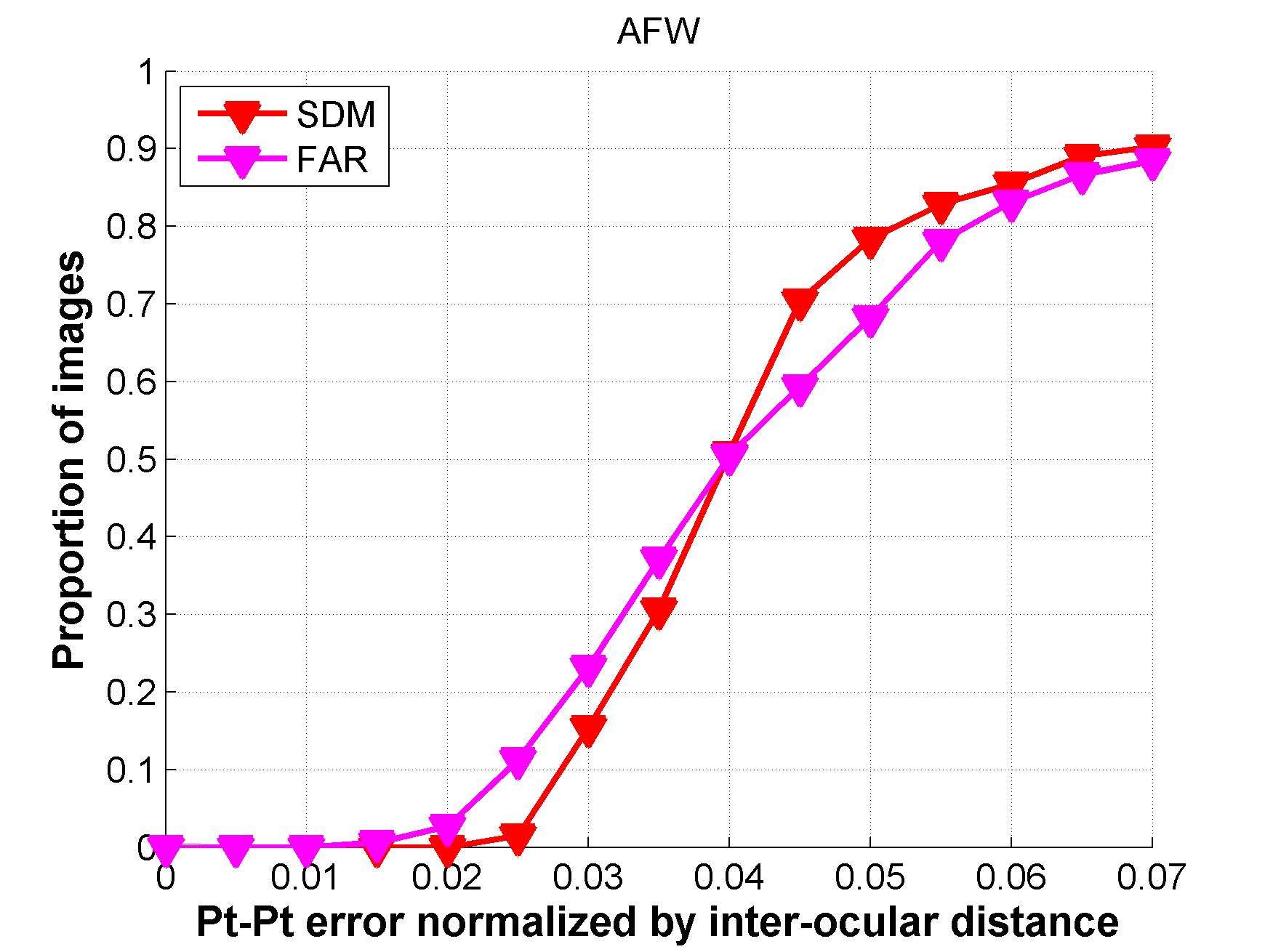}}\end{subfigure} &
\end{tabular}
\end{center}
\caption{Comparison of the cumulative error distribution curves obtained by the FAR and the SDM on the LFPW, the HELEN, and the AFW databases.}
\label{fig:gfa_results_sdm_far}
\end{figure*}

Given an input facial image, the initialization was produced by applying the detector \cite{ramanan}. The image, the initialization and $\b{U}$ were given as input into Algorithm~\ref{alg:main_alg}. By using the produced $\Delta \b{p}$ the outer loop was rerun again for one iteration without updating $\Delta \b{p}$ the execution of sub-problem (\ref{eq:sub_dp_solution}). 

In Fig. \ref{fig:frontal_rec} (rows: $1$-$2$) the reconstructed frontal faces from the non-frontal images (`ba', `bc', `bd', `be', `bf', `bg', and `bh') of `00268' subject from FERET database are illustrated. Fig. \ref{fig:frontal_rec} (rows: $3$-$4$) depicts the frontal reconstructed views from the images taken from MultiPIE with (a) `Surprise' at $-30^\circ$, (b) `Scream' at $-15^\circ$, (c) `Squint' at $0^\circ$, (d) `Neutral' at $+15^\circ$, and (e) `Smile' at $+30^\circ$. By visually inspecting Fig. \ref{fig:frontal_rec}, it is clear that the FAR  is robust to pose, expression, and lighting conditions variations. This attributed to the fact that the matrix $\ell_1$-norm was adopted for non-Gaussian noise characterization.  Frontal reconstructed views from in-the-wild images are depicted in Fig. \ref{fig:frontal_rec} (rows: $5$-$6$). 

To quantitatively assess the quality of the frontalized images the following experiment was conducted. To this end, `Neutral' images of  $20$ different subjects of MultiPIE under poses $-30^0:30^0$ ($5$ for each subject, $100$ in total) were selected. The images of each subject were frontalized by employing the FAR. The Root Mean Square Error (RMSE) between each frontalized image and the real frontal image of the subject is used as evaluation metric. The performance of the FAR with respect to RMSE is compared with that obtained by the frontalization system of the DeepFace \cite{deepface}. The average RMSEs of the FAR and DeepFace are $0.0817$ and $0.1025$, respectively. It is worth noting that, even DeepFace employs a 3D model to handle out-of-planar rotations, the FAR performs better without using any kind of 3D information.

\subsection{Face landmark localization} \label{subsec:face_alignment}

The performance of the FAR in the generic face alignment problem is assessed by conducting experiments on in-the-wild databases namely. the LFPW, the HELEN and the AFW. To this end, the performance of the FAR is compared against that obtained by (a) the AAMs, the CLMs, and the SDM using exactly the same training data as well as the same features and (b)  the state-of-the-art method and features. 
The annotations provided in \cite{sagonas2013semi,sagonas300} have been employed for evaluation purposes. The average point-to-point Euclidean distance of $49$ interior landmark points (excluding the points correspond to face boundary) normalized by the Euclidean distance of the outer corner of eyes is used as the evaluation measure. In addition, the cumulative error distribution curve (CED) for each method is computed by using the fraction of test images for which the average error was smaller than a threshold. 

\textbf{Same train set and features}: In order to compare fairly the competing methods, the same training data, initialization, and features were employed. The $500$ frontal images used to build the $\b{U}$, were used as the training set while the pixel intensities (PIs) were selected as the texture representation. The results produced by the detector \cite{ramanan} were used to initialize the methods. For this experiment the implementations provided by the platform  MENPO \cite{menpo14} were used for all methods.

\begin{figure}
\begin{center}
\begin{tabular}{ccccc}
\begin{subfigure}{\includegraphics[width=0.09\textwidth]{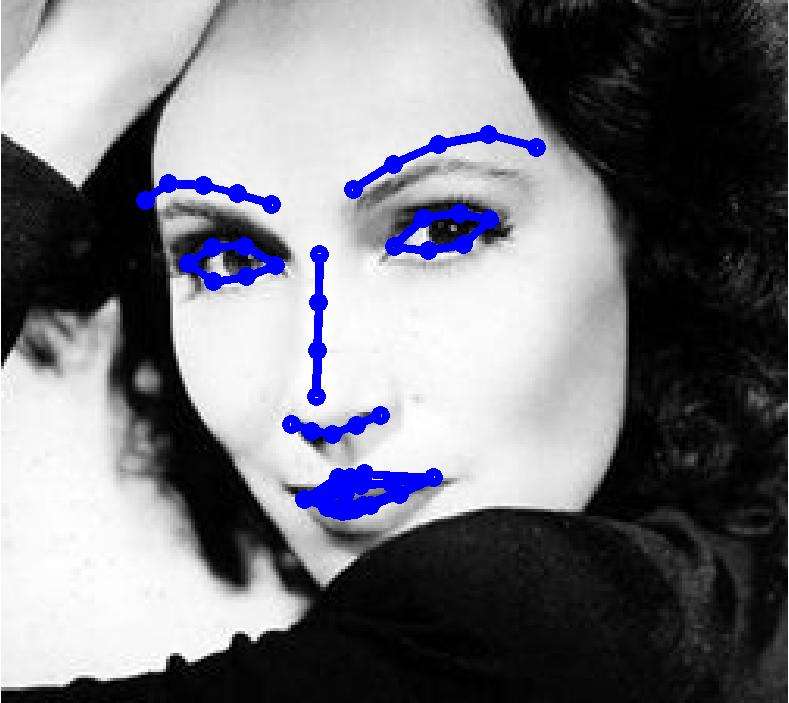}}\end{subfigure} &
\begin{subfigure}{\includegraphics[width=0.09\textwidth]{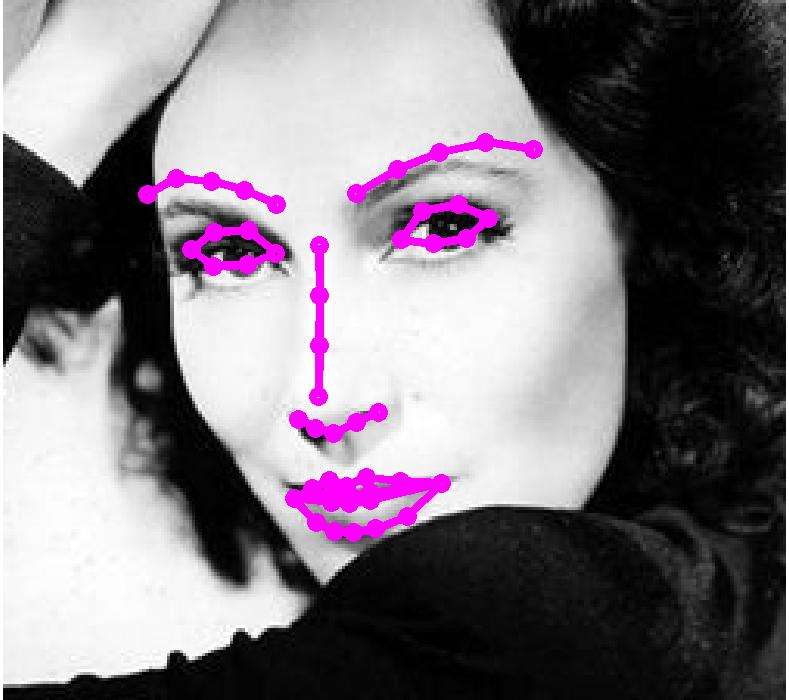}}\end{subfigure} &
\begin{subfigure}{\includegraphics[width=0.09\textwidth]{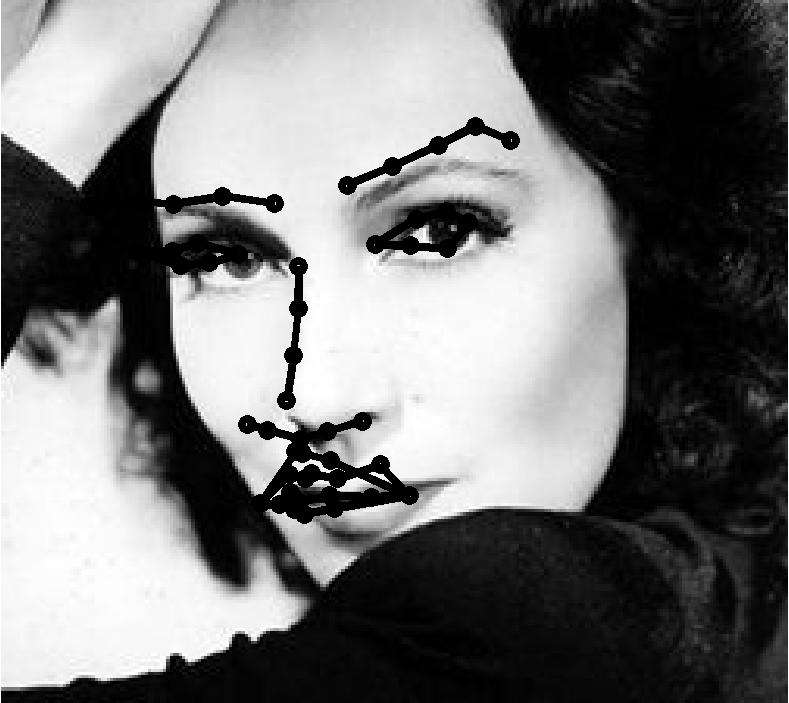}}\end{subfigure} & 
\begin{subfigure}{\includegraphics[width=0.09\textwidth]{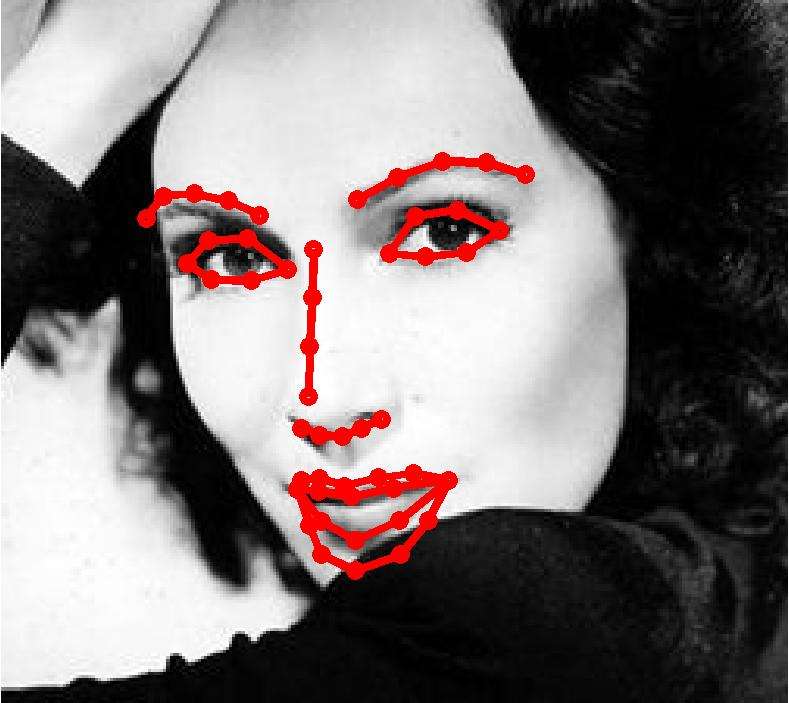}}\end{subfigure} &
\begin{subfigure}{\includegraphics[width=0.09\textwidth]{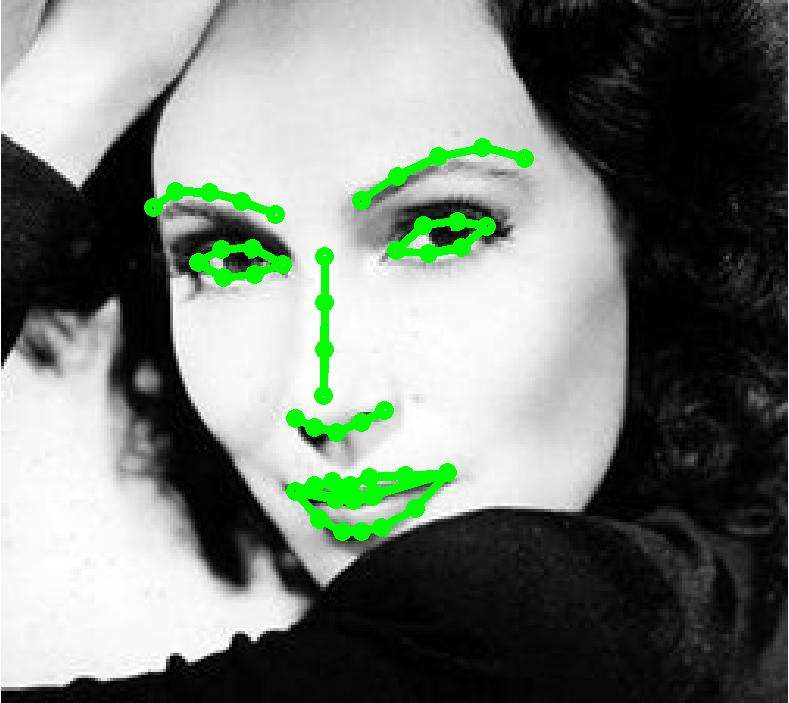}}\end{subfigure} \\
\vspace{-0.75cm} \\

\begin{subfigure}{\includegraphics[width=0.09\textwidth]{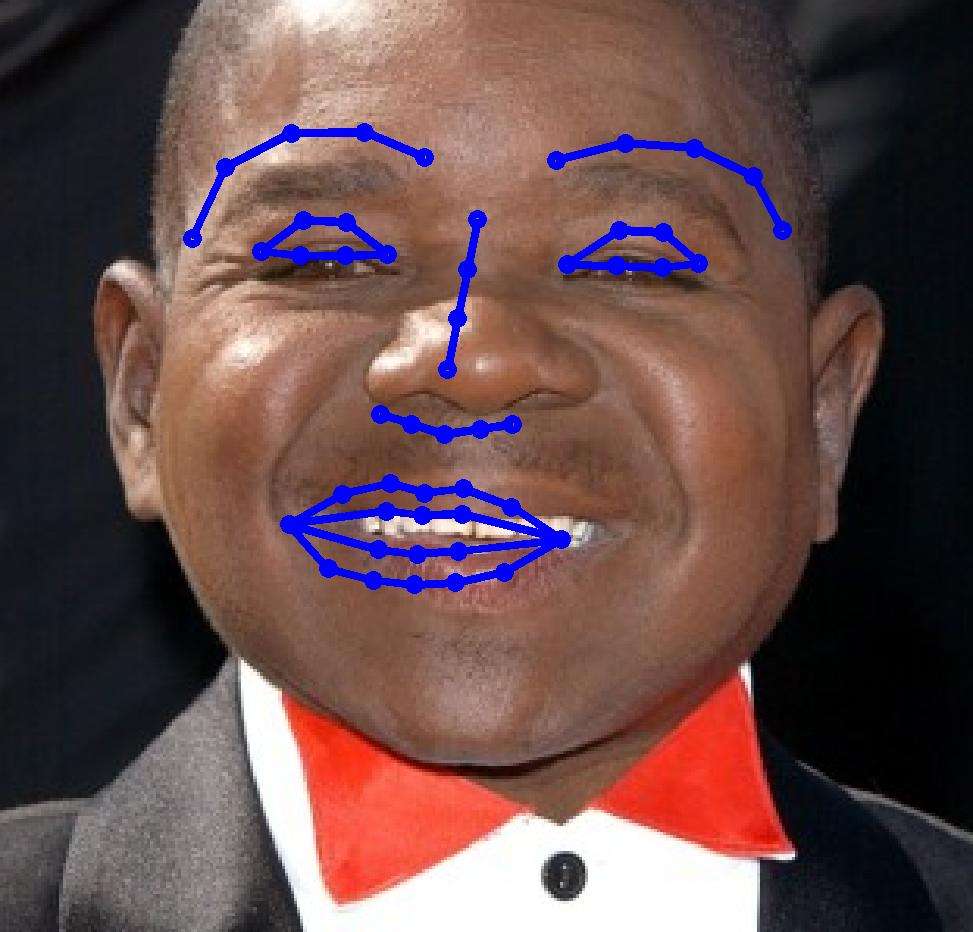}}\end{subfigure} &
\begin{subfigure}{\includegraphics[width=0.09\textwidth]{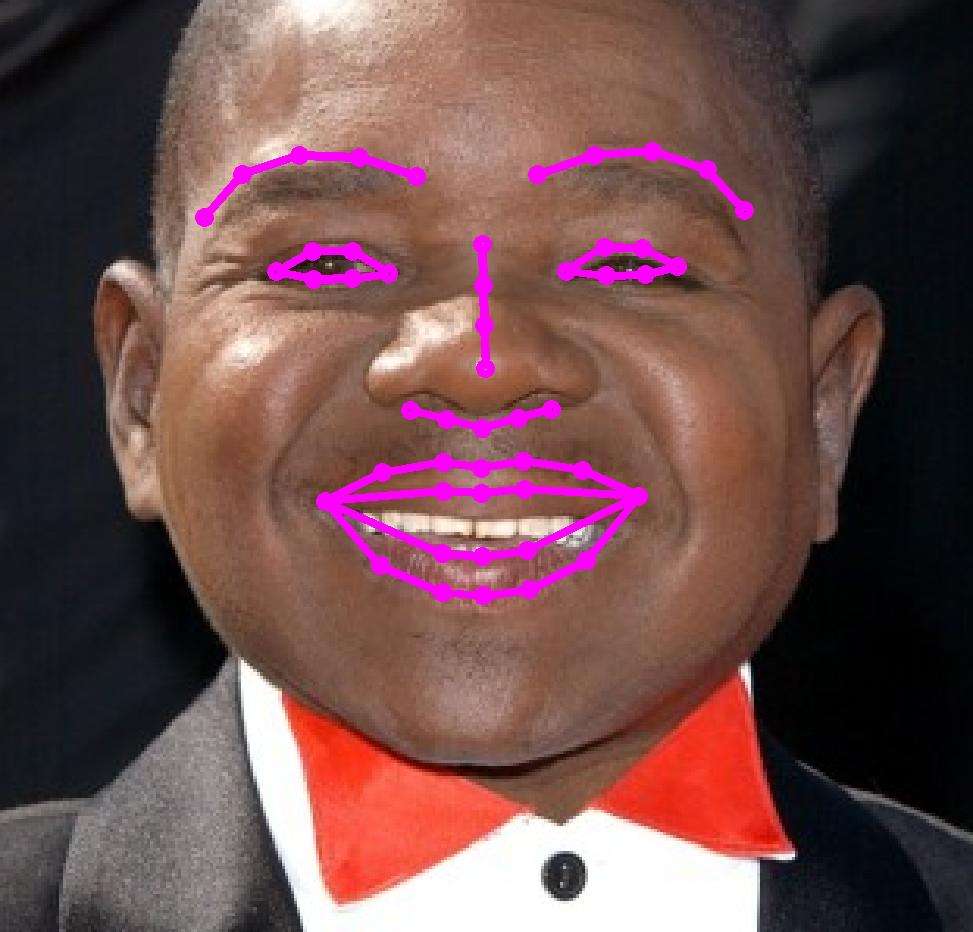}}\end{subfigure} &
\begin{subfigure}{\includegraphics[width=0.09\textwidth]{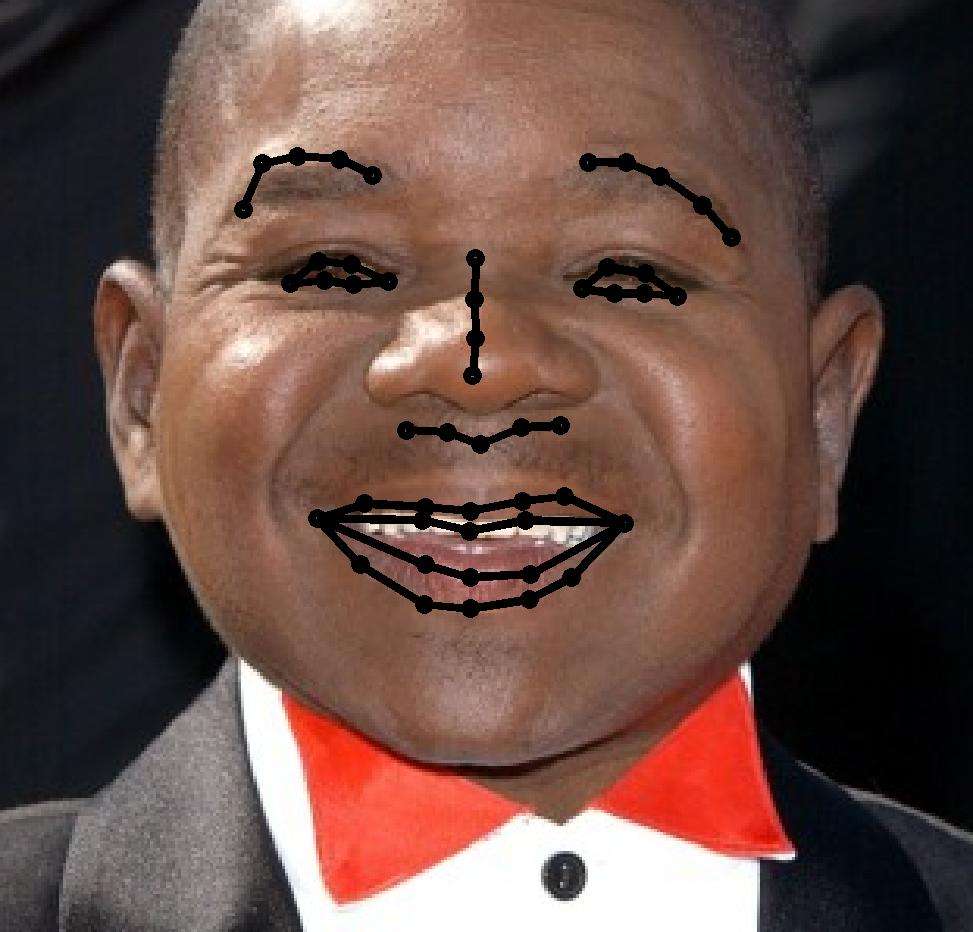}}\end{subfigure} & 
\begin{subfigure}{\includegraphics[width=0.09\textwidth]{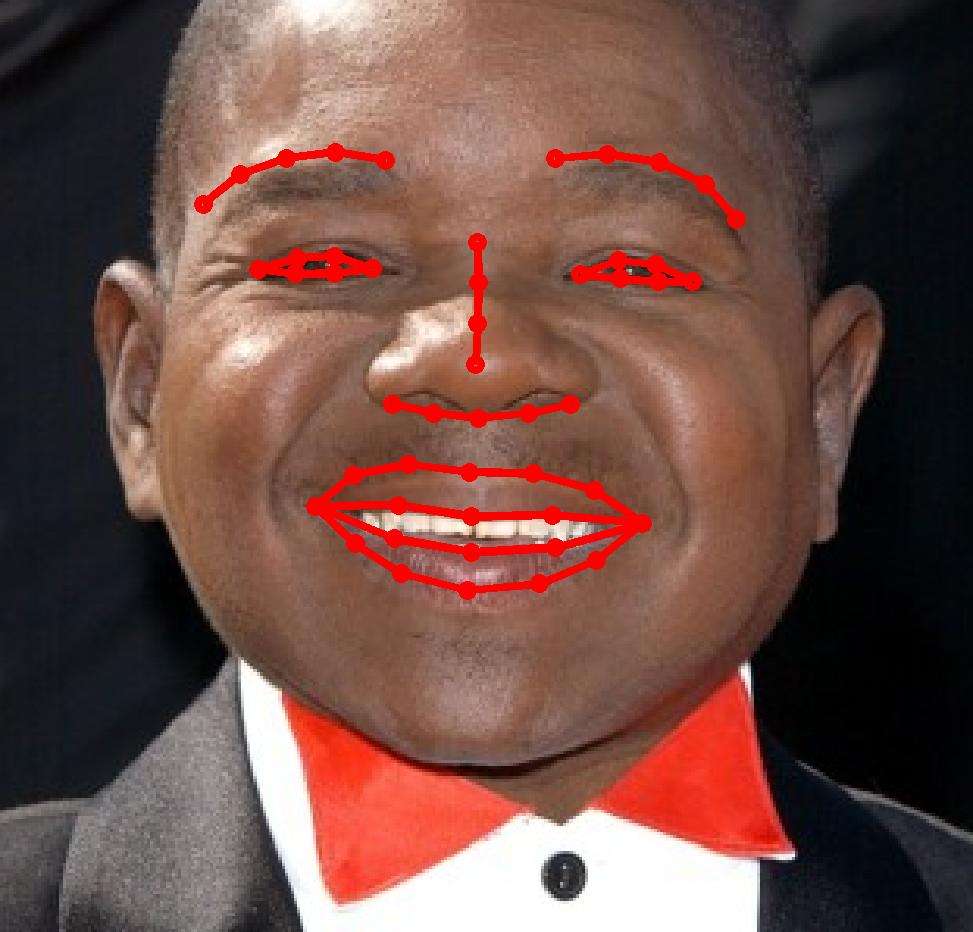}}\end{subfigure} &
\begin{subfigure}{\includegraphics[width=0.09\textwidth]{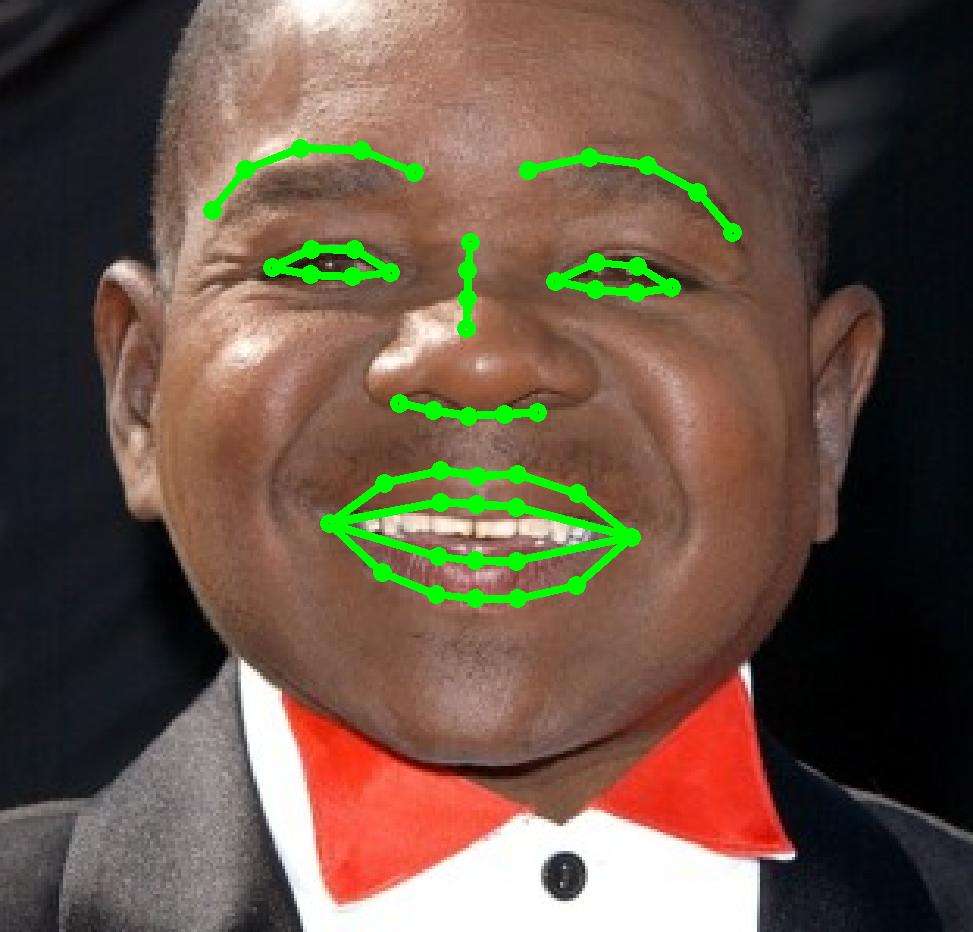}}\end{subfigure} \\
\vspace{-0.75cm} \\

\begin{subfigure}{\includegraphics[width=0.09\textwidth]{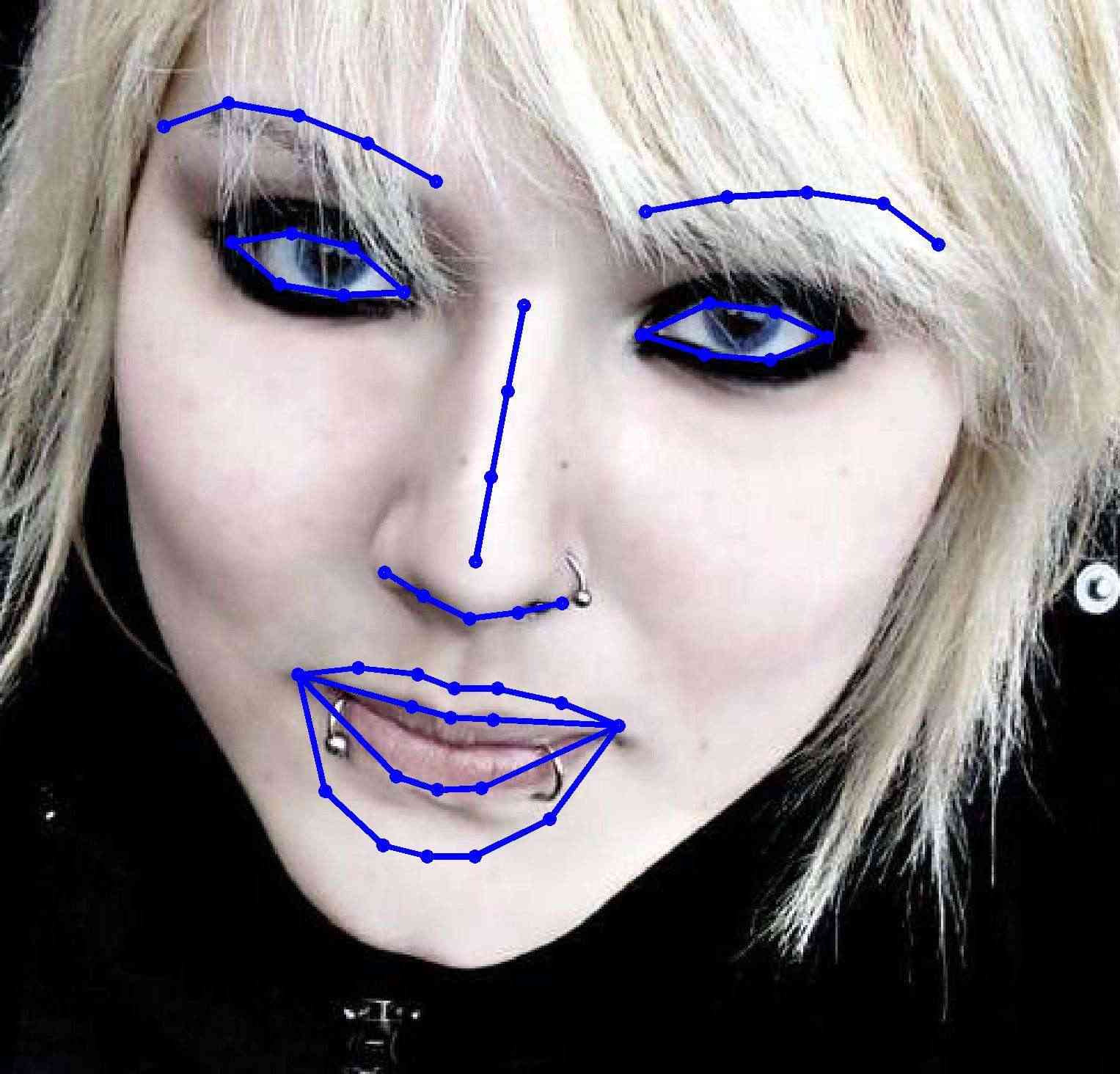}}\end{subfigure} &
\begin{subfigure}{\includegraphics[width=0.09\textwidth]{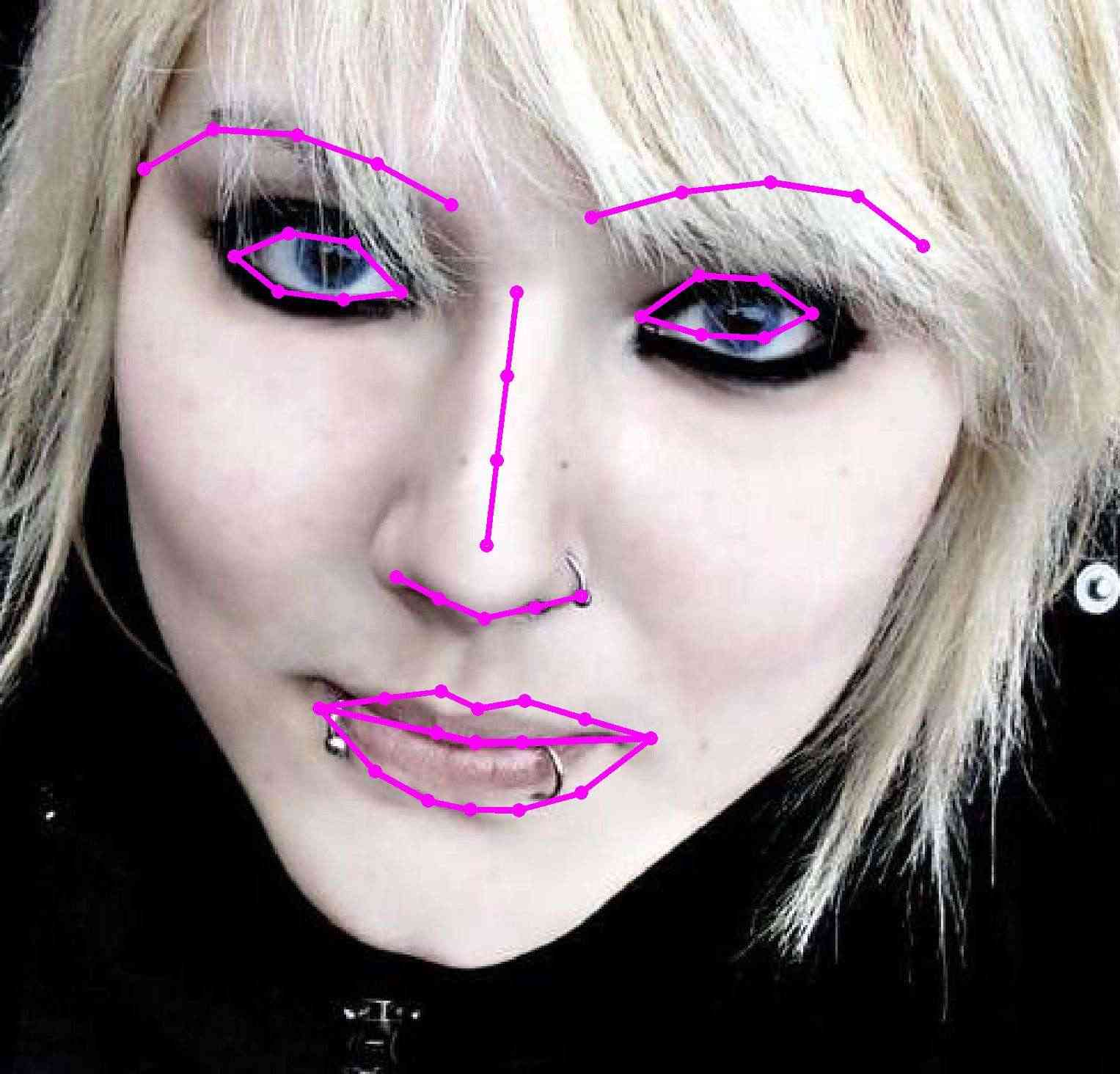}}\end{subfigure} &
\begin{subfigure}{\includegraphics[width=0.09\textwidth]{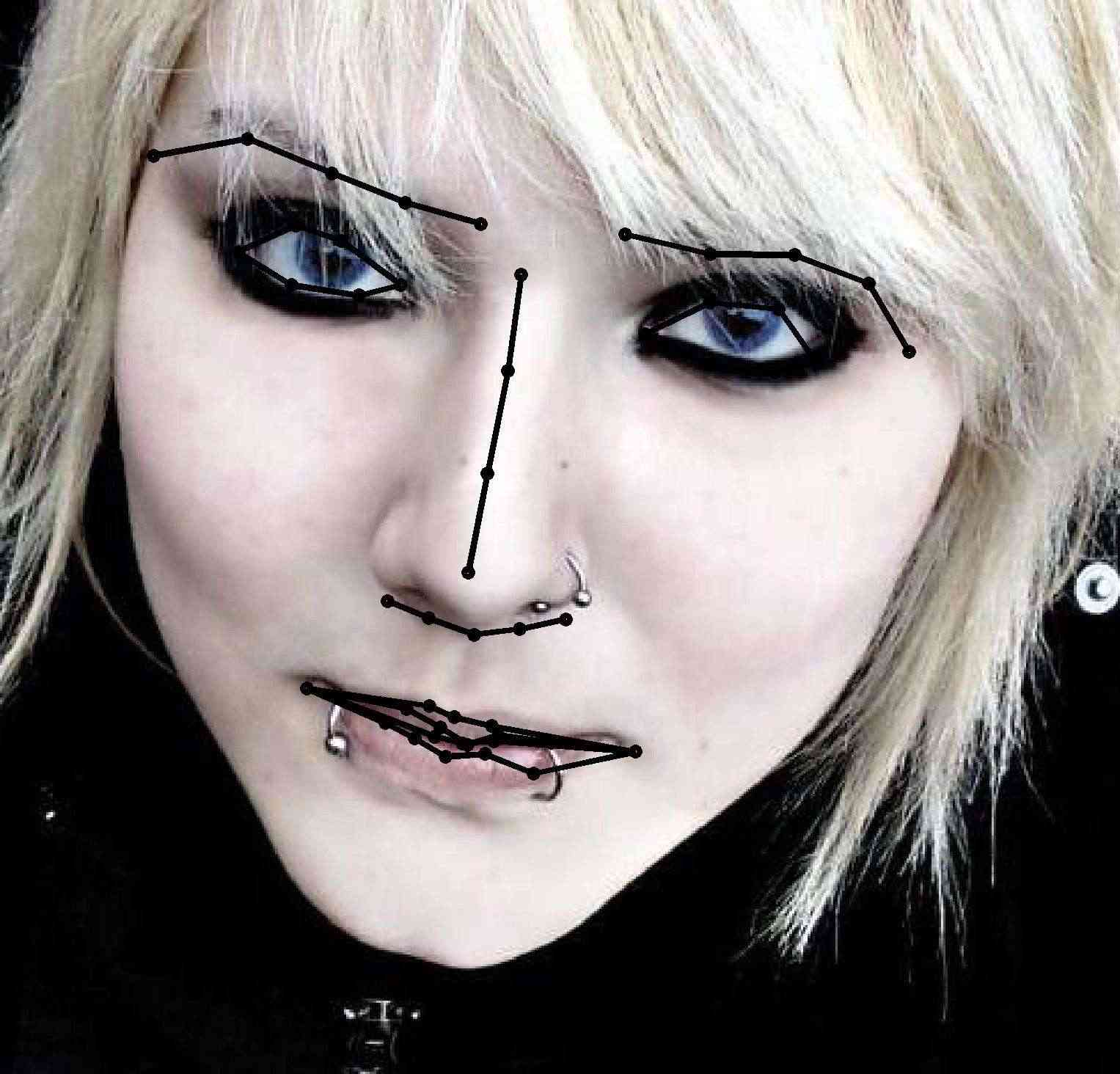}}\end{subfigure} & 
\begin{subfigure}{\includegraphics[width=0.09\textwidth]{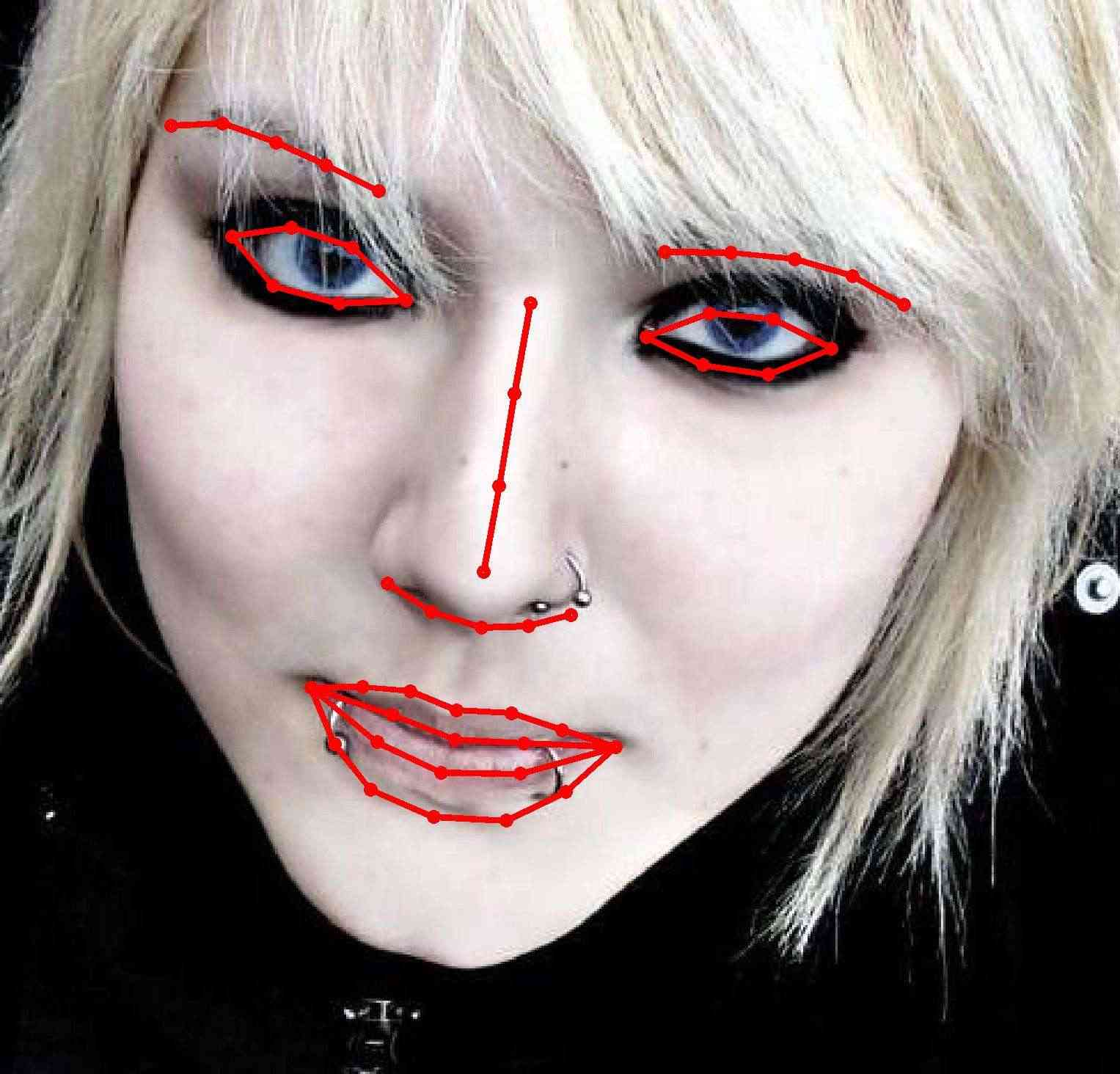}}\end{subfigure} &
\begin{subfigure}{\includegraphics[width=0.09\textwidth]{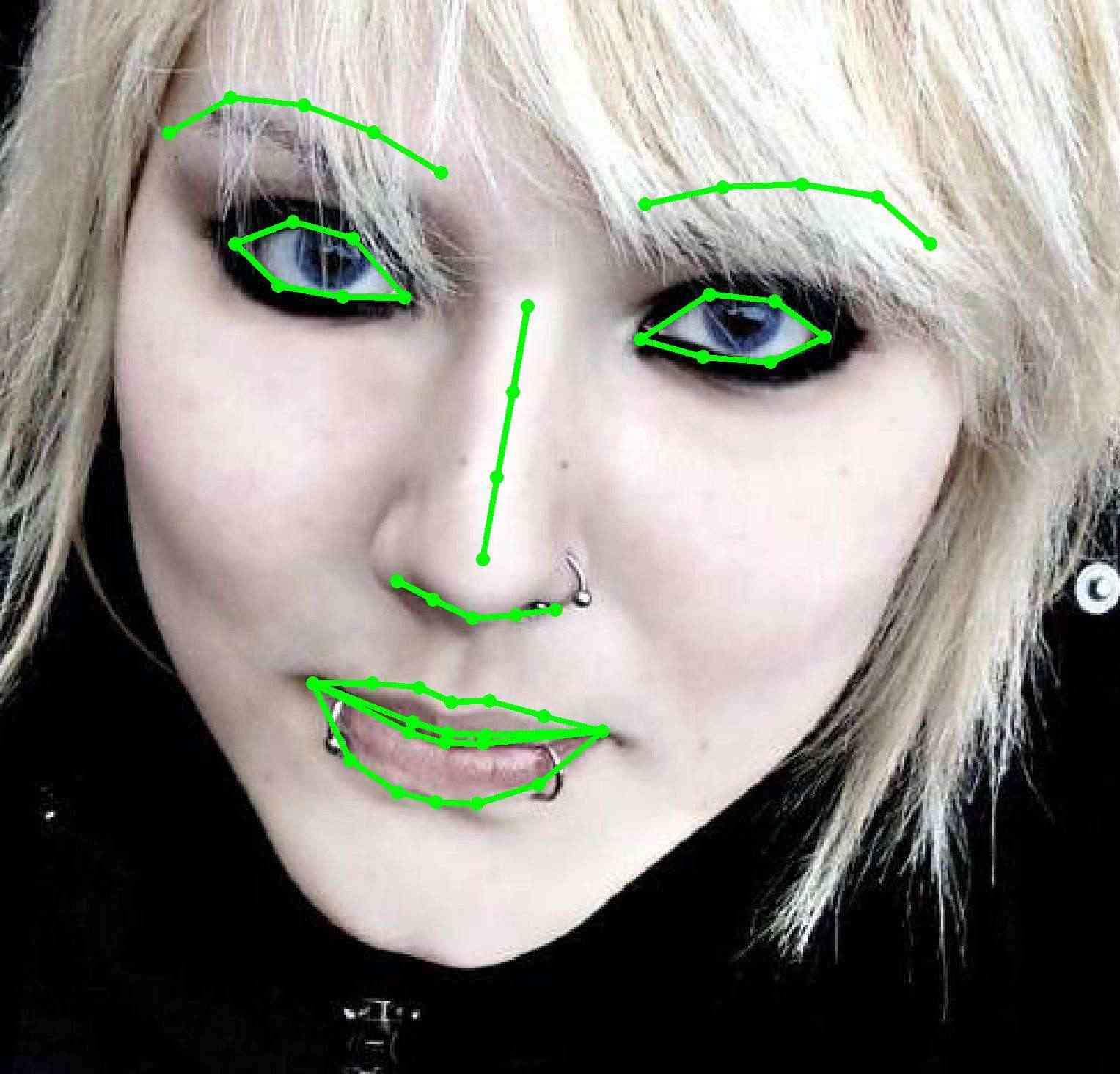}}\end{subfigure} \\
\vspace{-0.75cm} \\

\begin{subfigure}{\includegraphics[width=0.09\textwidth]{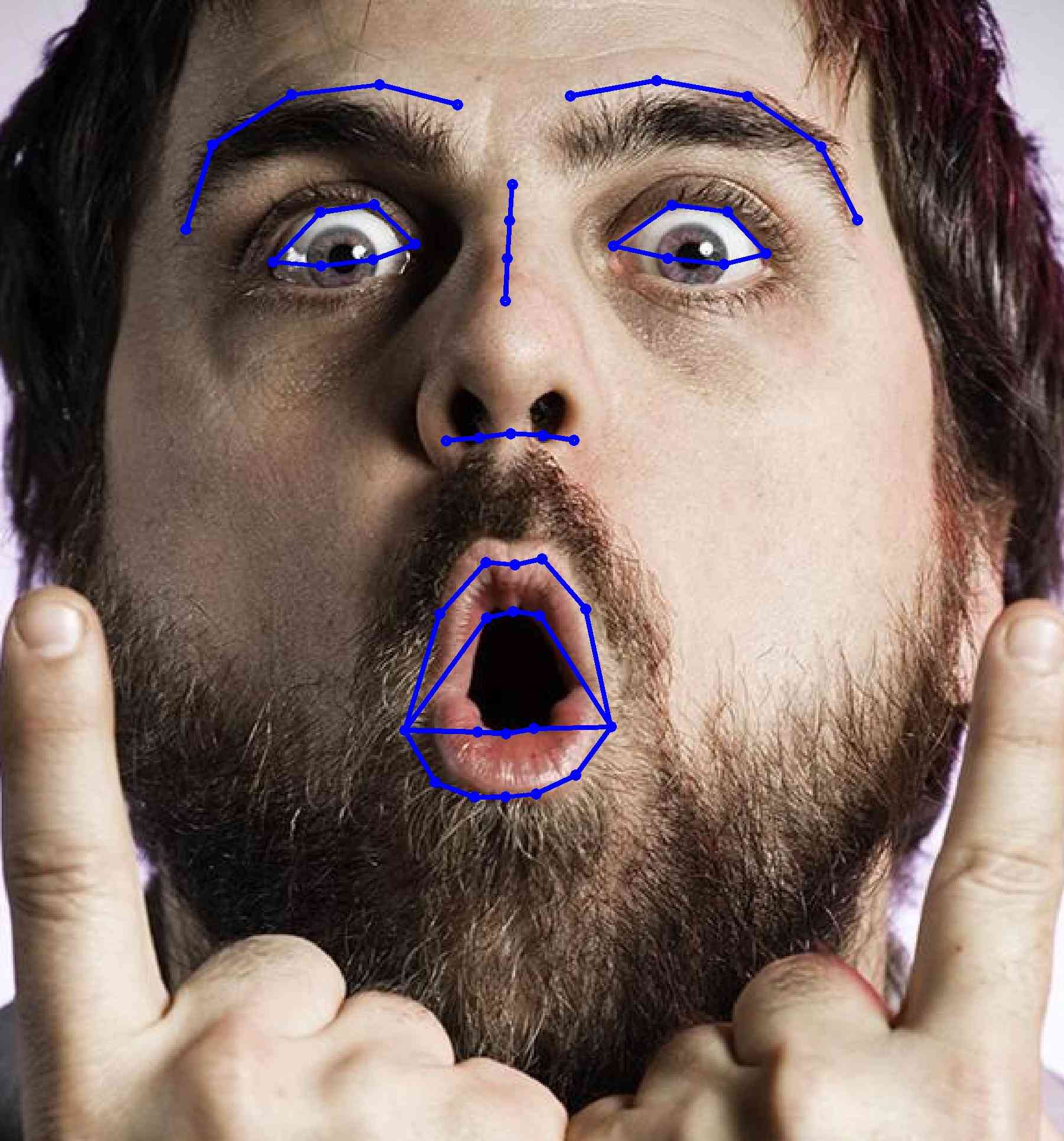}}\end{subfigure} &
\begin{subfigure}{\includegraphics[width=0.09\textwidth]{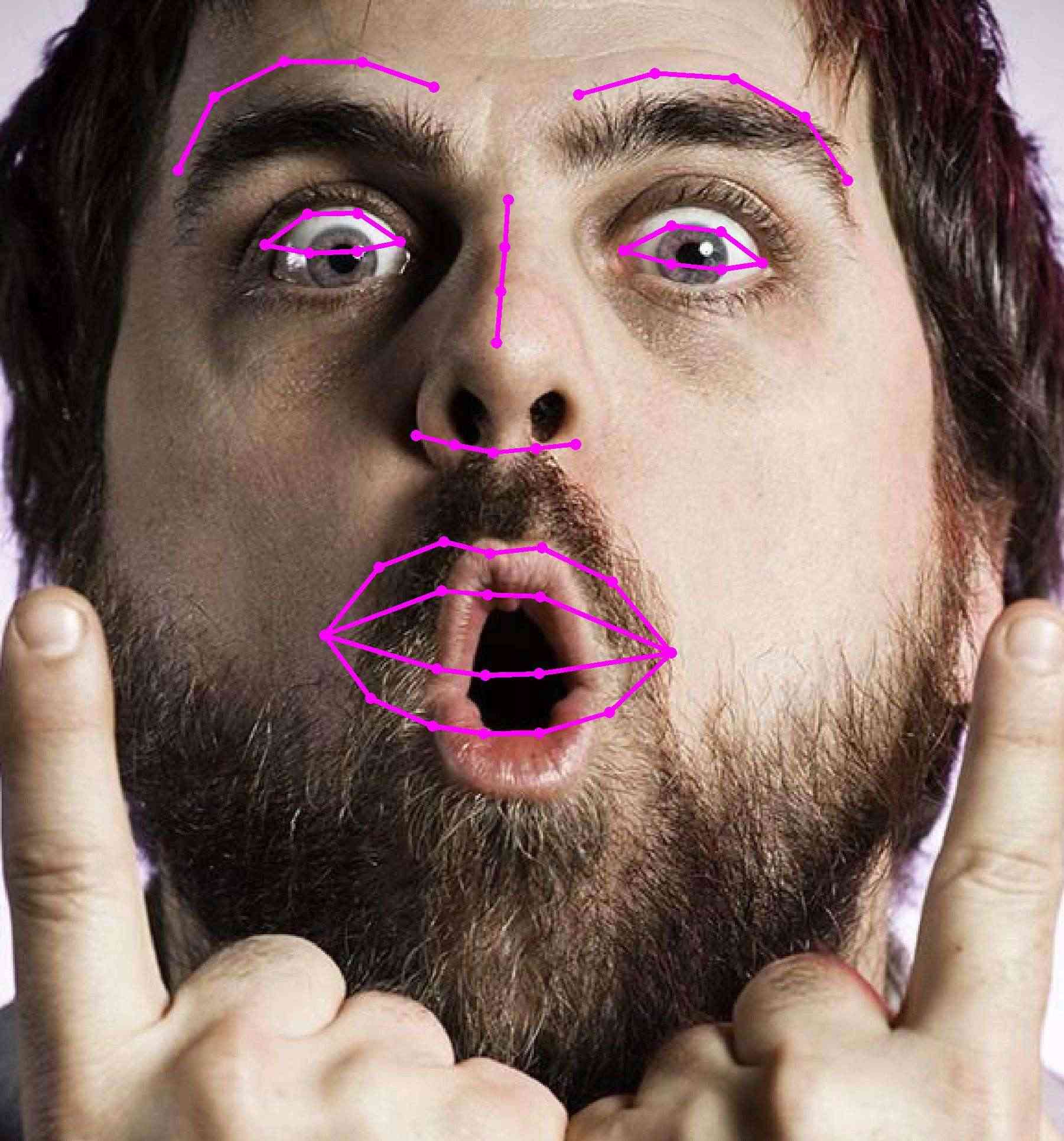}}\end{subfigure} &
\begin{subfigure}{\includegraphics[width=0.09\textwidth]{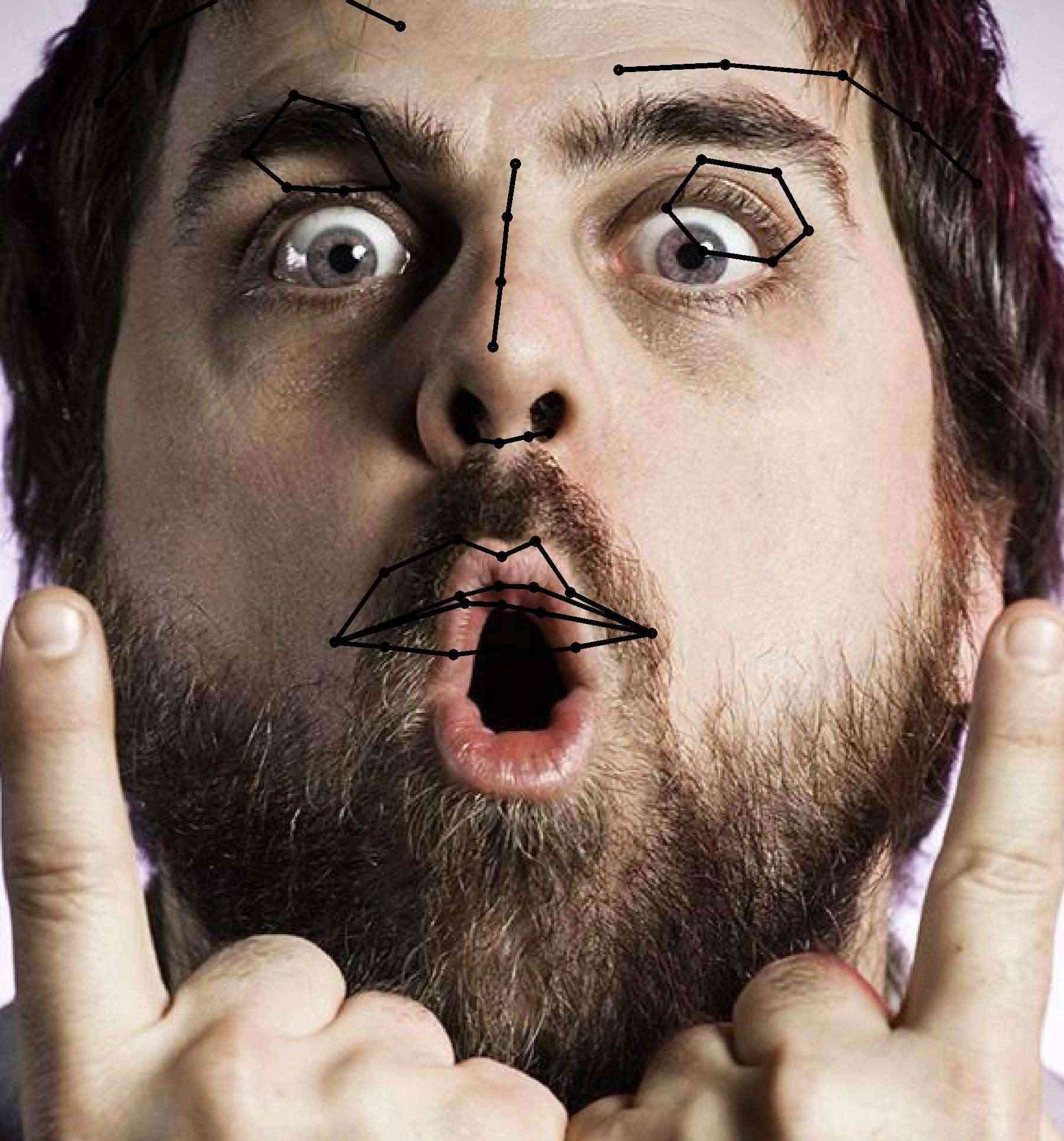}}\end{subfigure} & 
\begin{subfigure}{\includegraphics[width=0.09\textwidth]{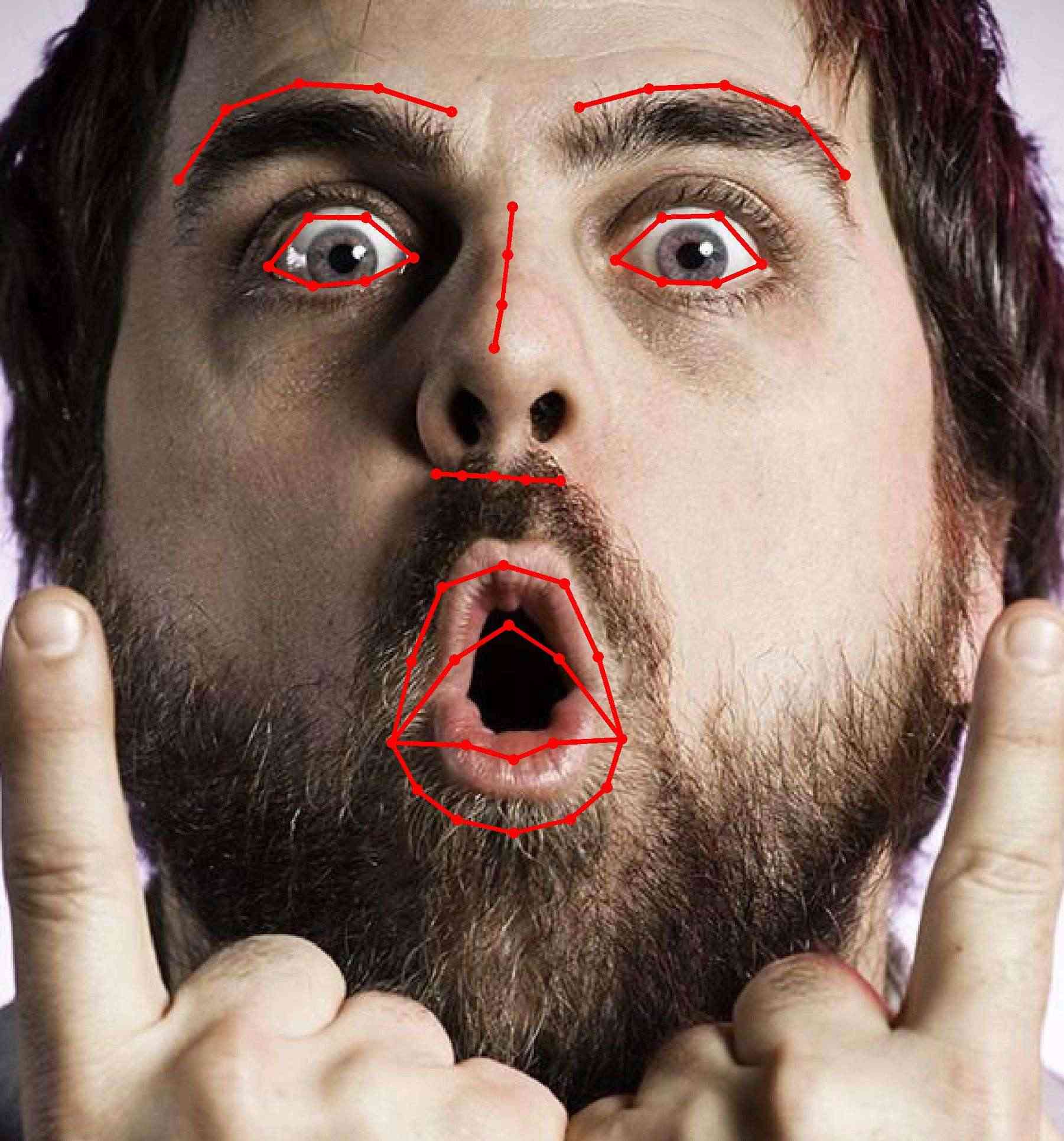}}\end{subfigure} &
\begin{subfigure}{\includegraphics[width=0.09\textwidth]{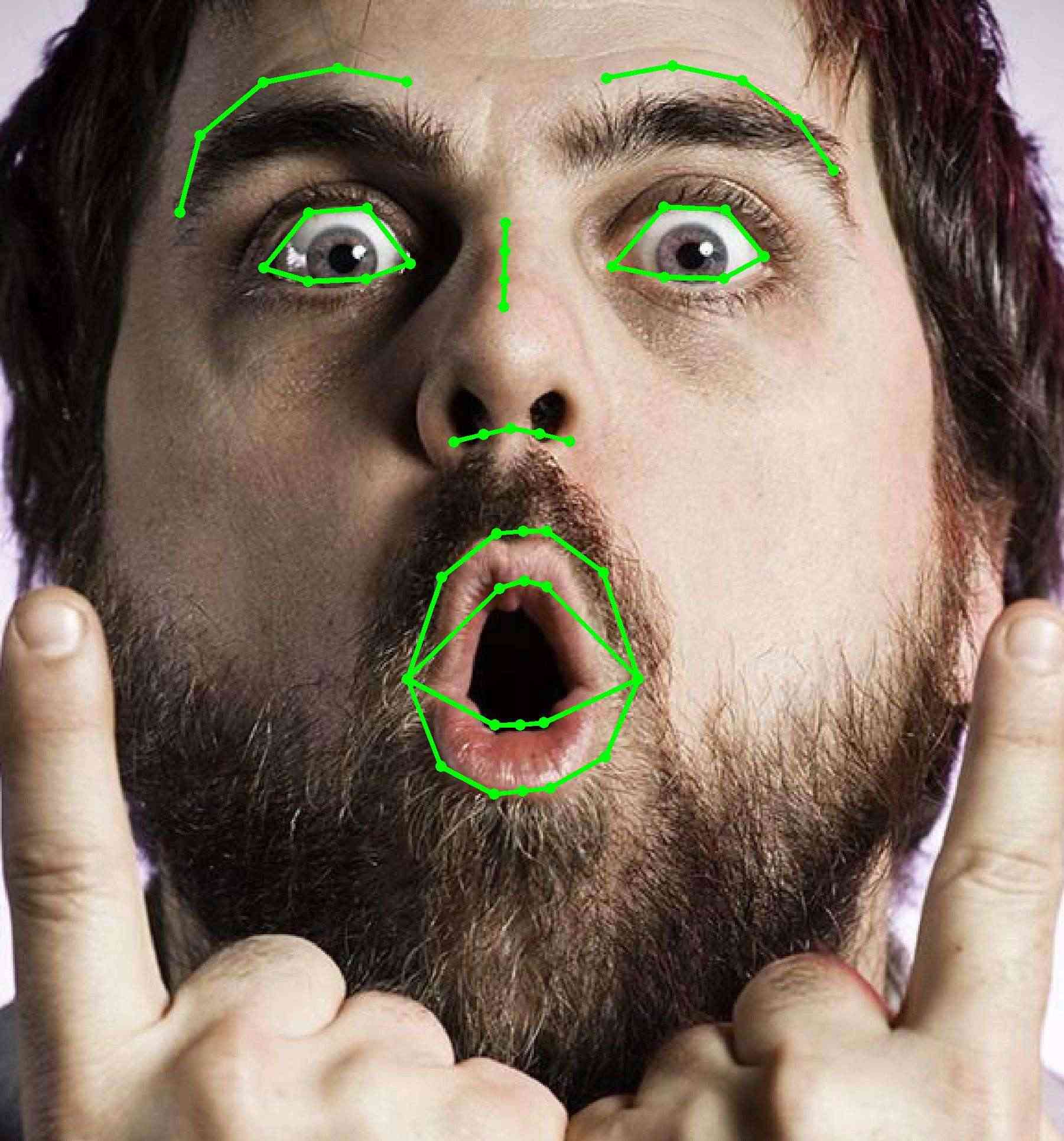}}\end{subfigure} \\
\vspace{-0.76cm} \\

\begin{subfigure}{\includegraphics[width=0.09\textwidth]{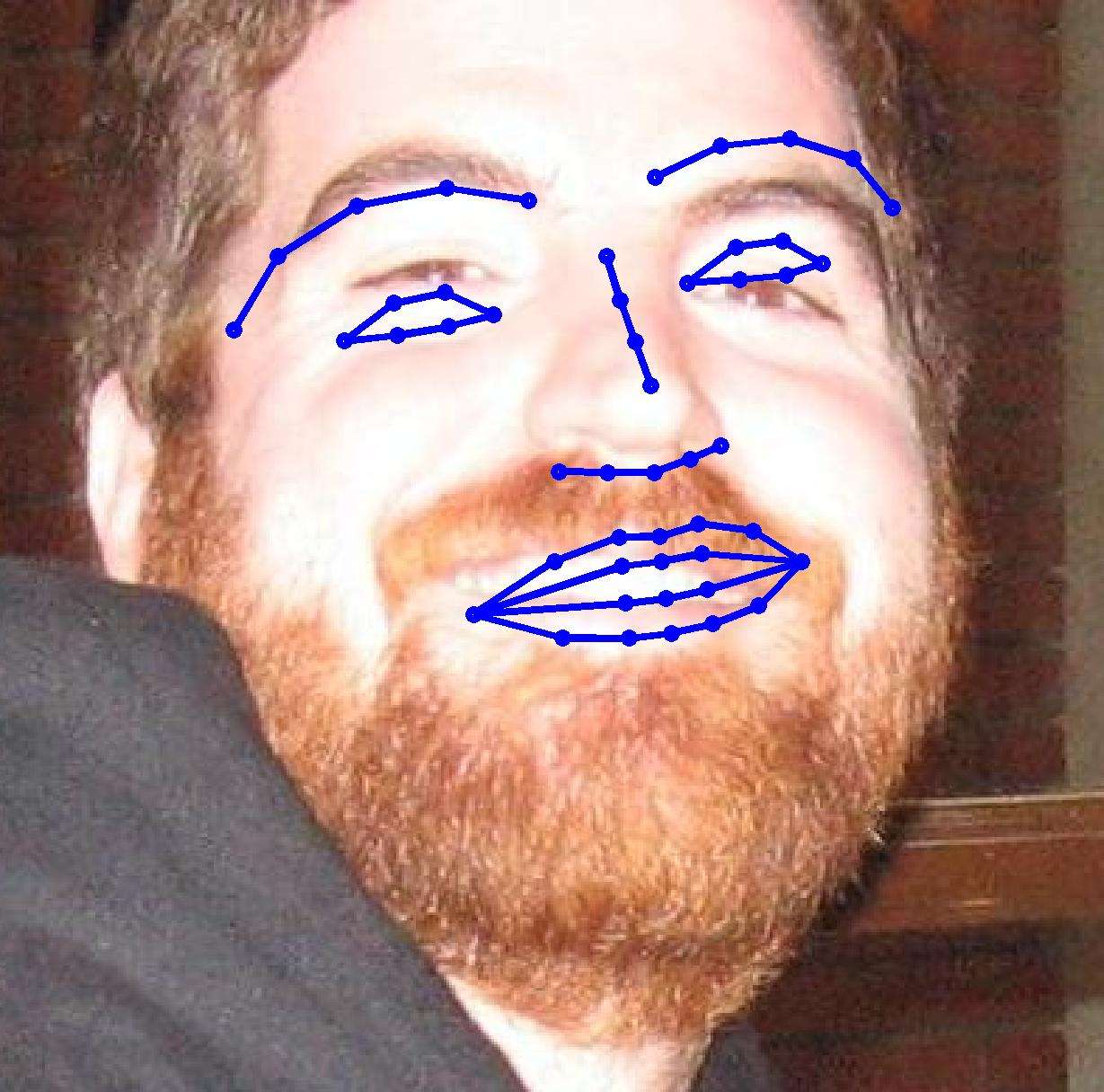}}\end{subfigure} &
\begin{subfigure}{\includegraphics[width=0.09\textwidth]{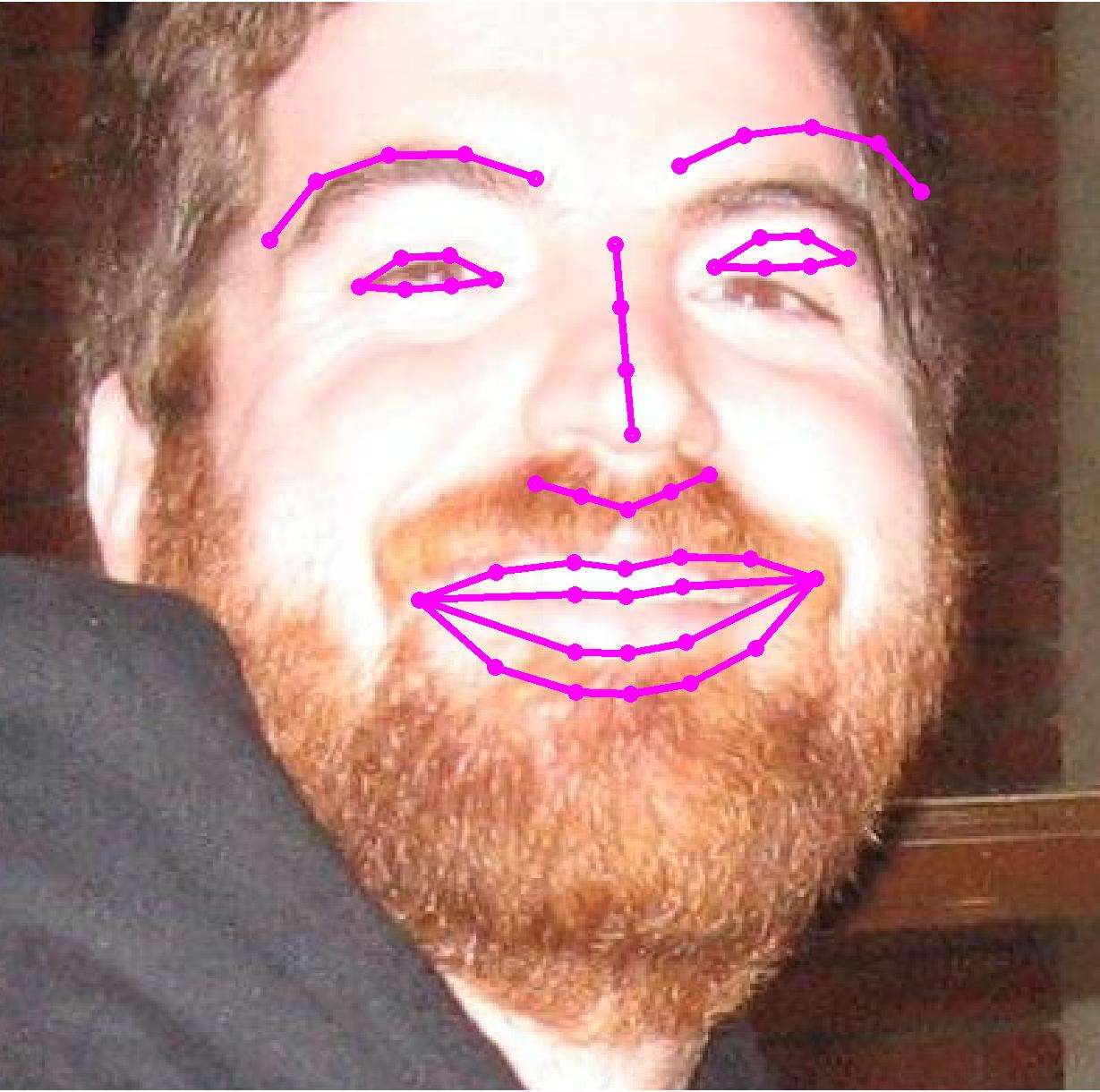}}\end{subfigure} &
\begin{subfigure}{\includegraphics[width=0.09\textwidth]{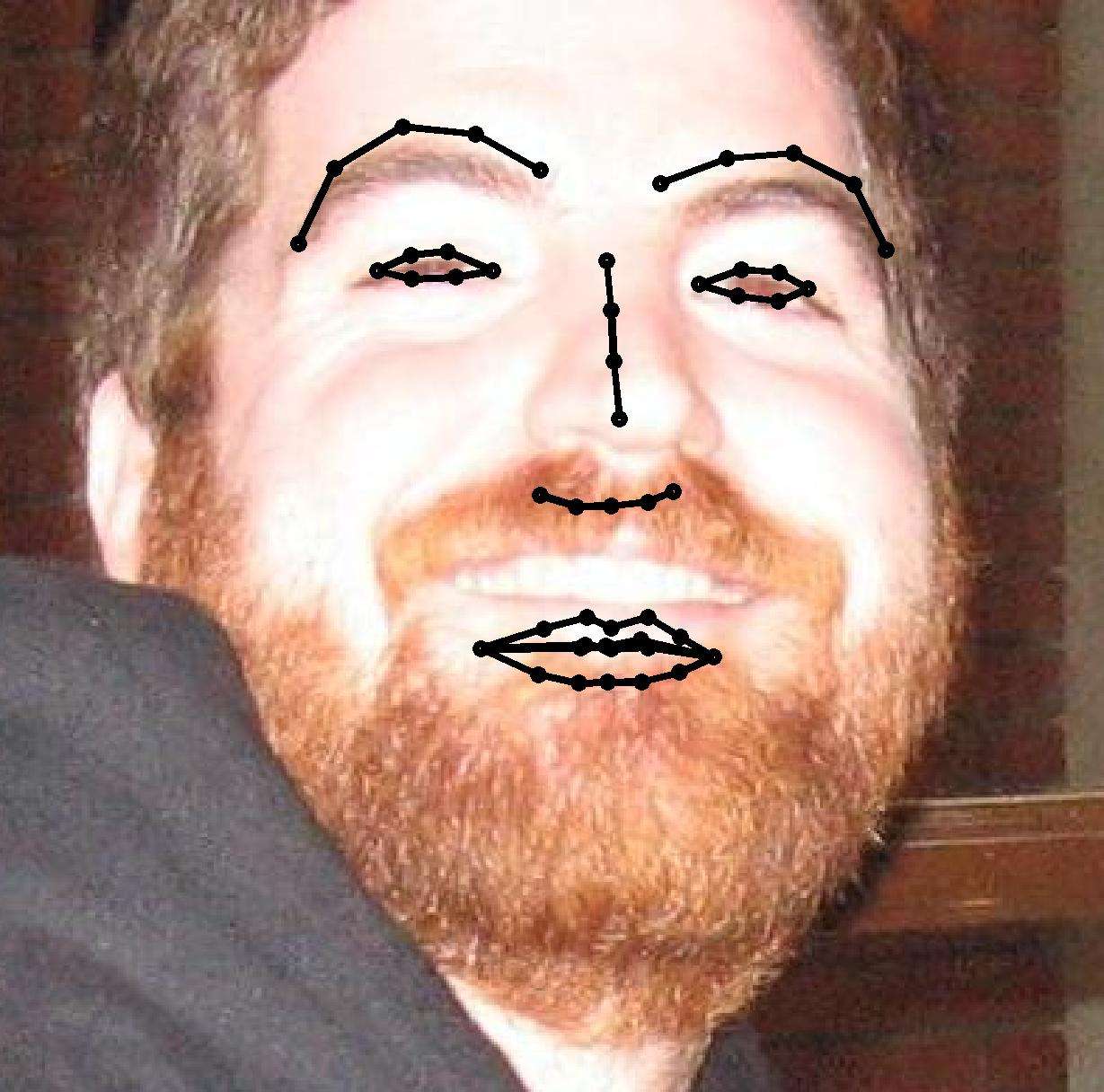}}\end{subfigure} & 
\begin{subfigure}{\includegraphics[width=0.09\textwidth]{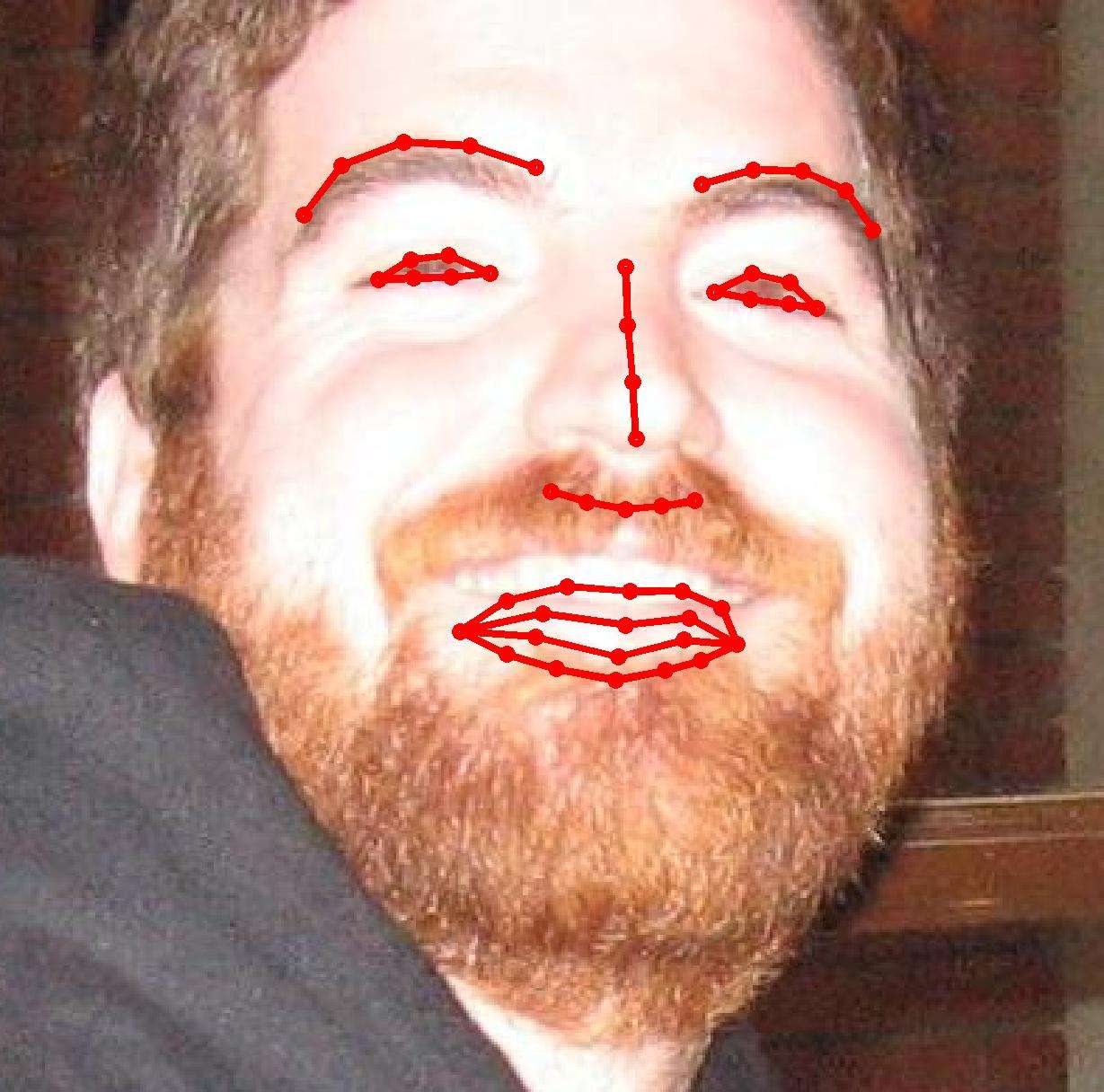}}\end{subfigure} &
\begin{subfigure}{\includegraphics[width=0.09\textwidth]{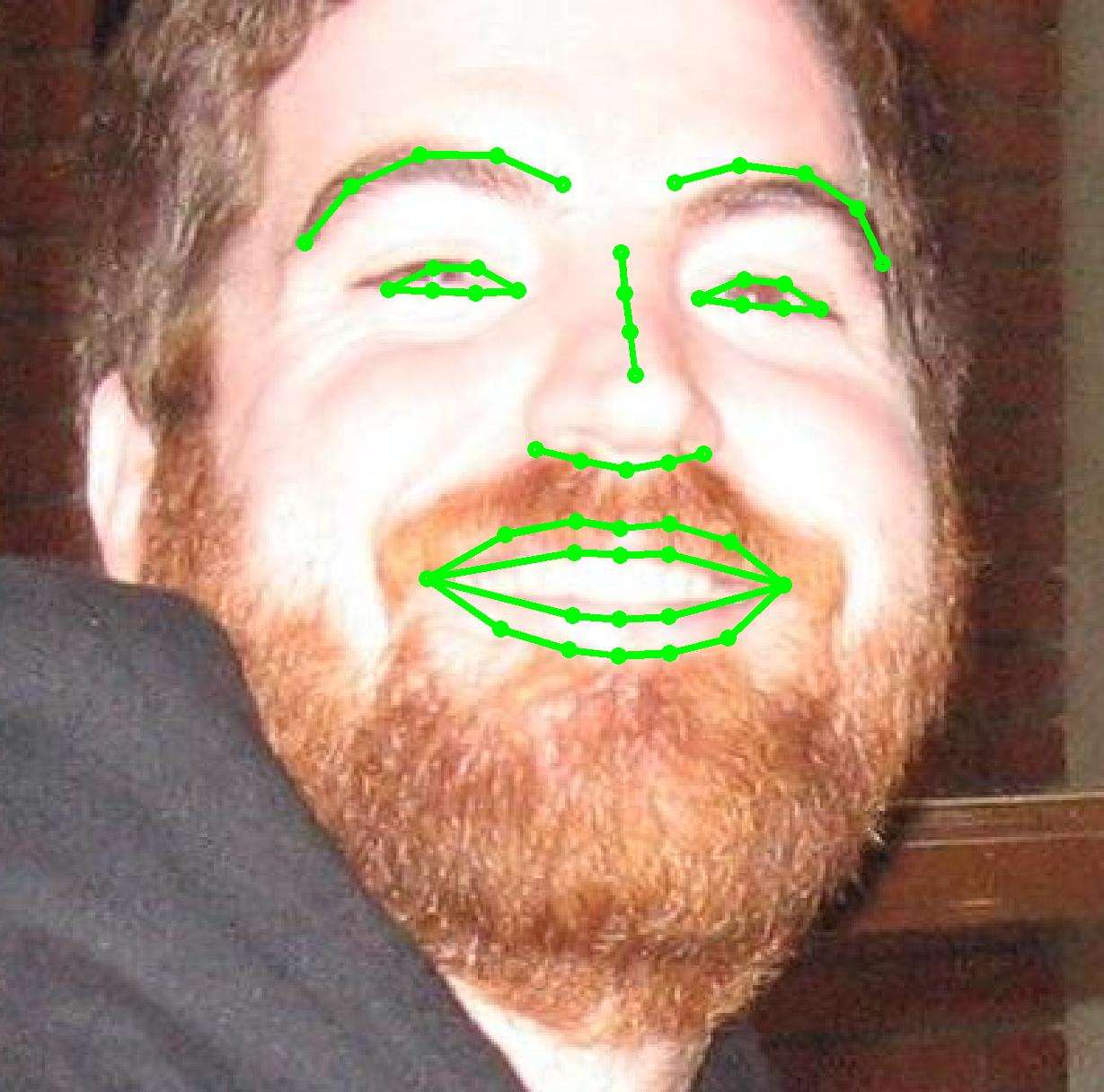}}\end{subfigure} \\
\vspace{-0.76cm} \\

\begin{subfigure}{\includegraphics[width=0.09\textwidth]{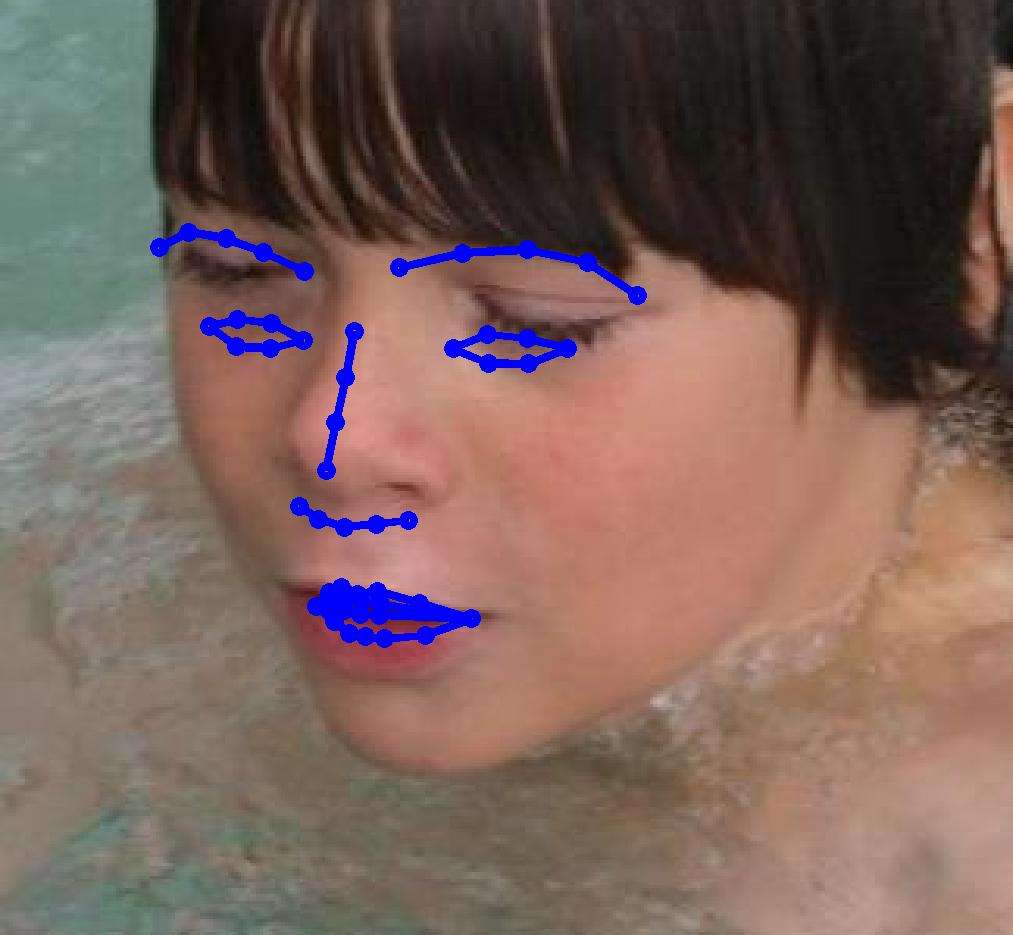}}\end{subfigure} &
\begin{subfigure}{\includegraphics[width=0.09\textwidth]{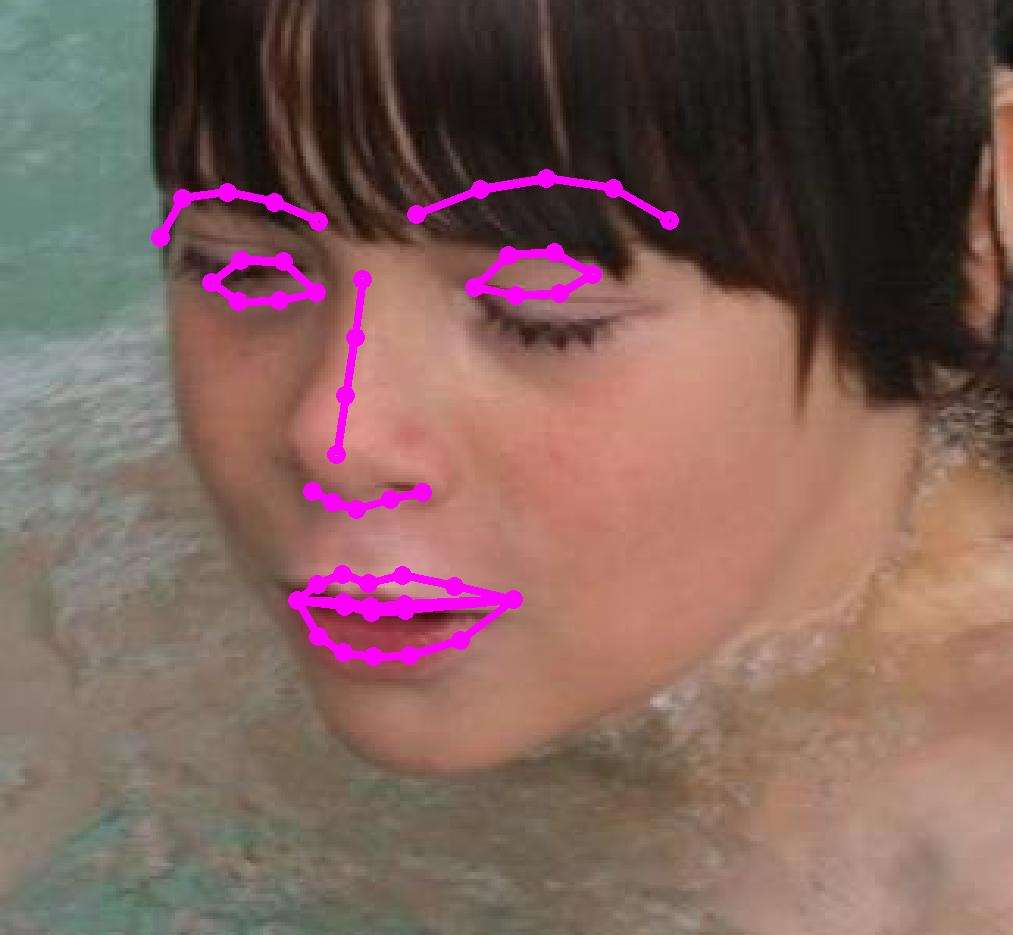}}\end{subfigure} &
\begin{subfigure}{\includegraphics[width=0.09\textwidth]{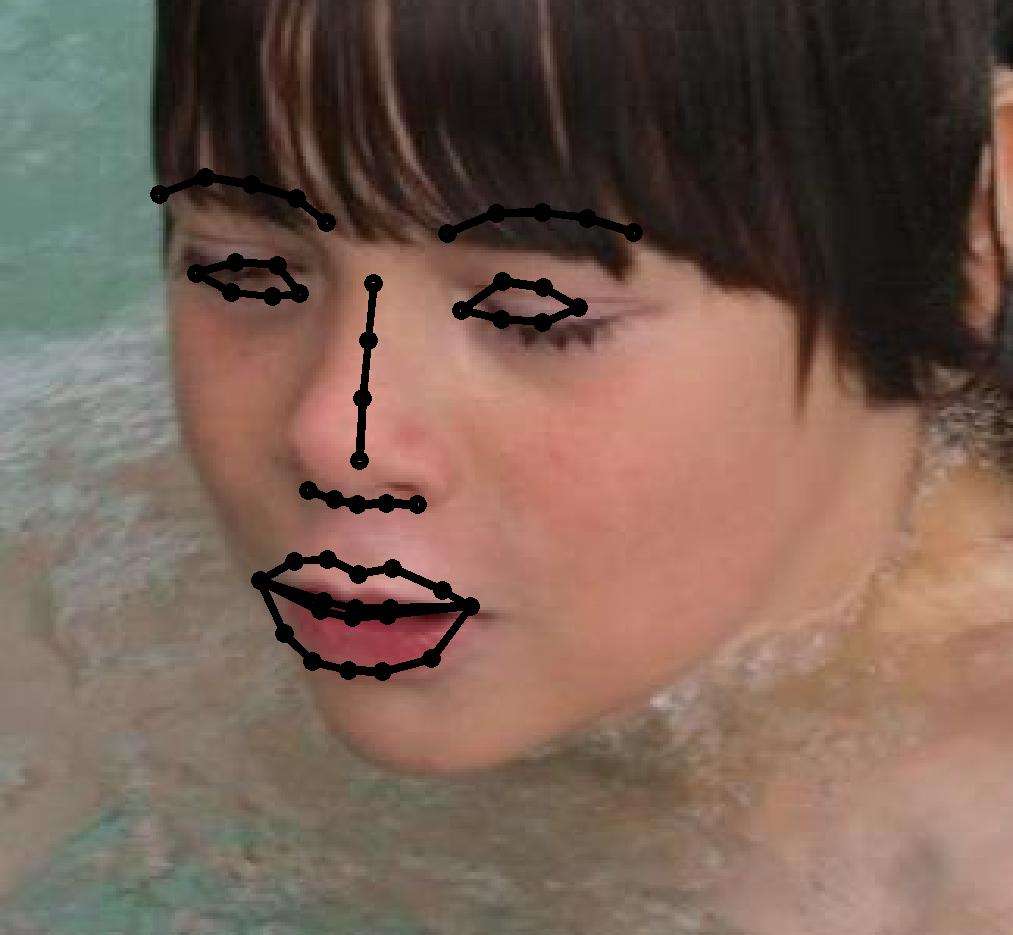}}\end{subfigure} & 
\begin{subfigure}{\includegraphics[width=0.09\textwidth]{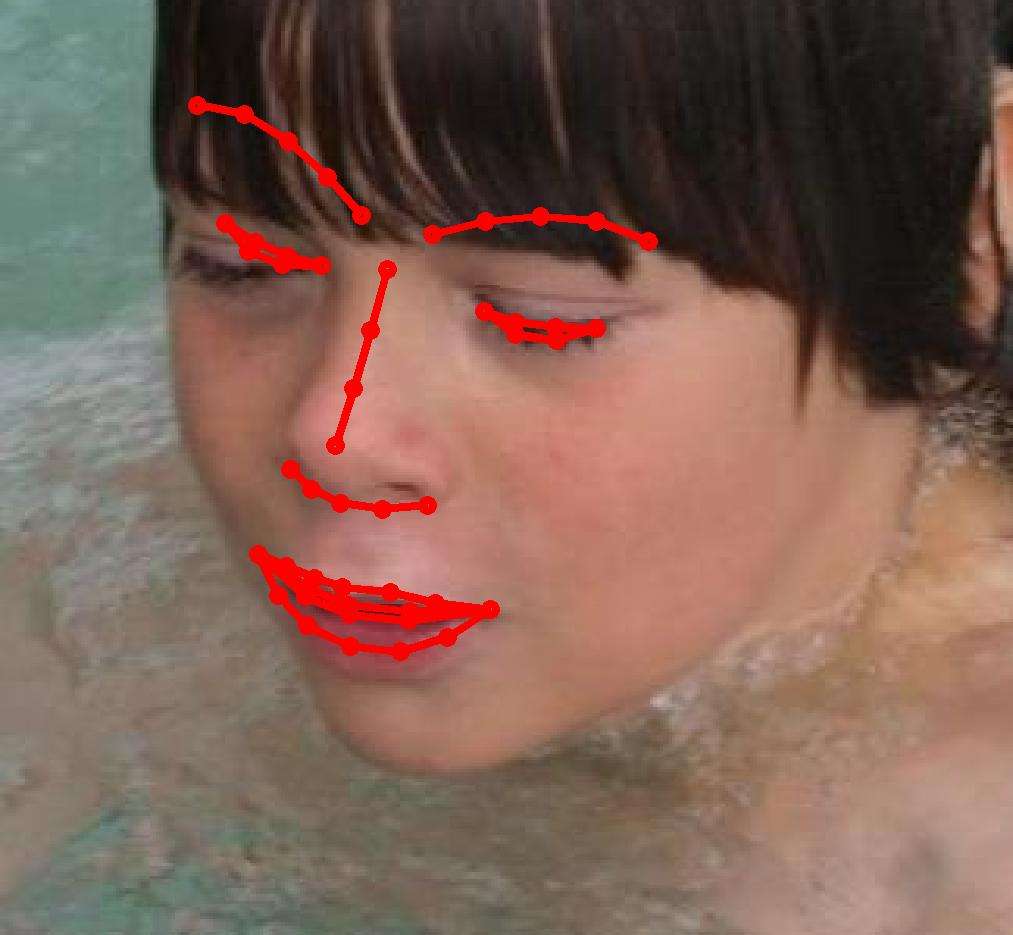}}\end{subfigure} &
\begin{subfigure}{\includegraphics[width=0.09\textwidth]{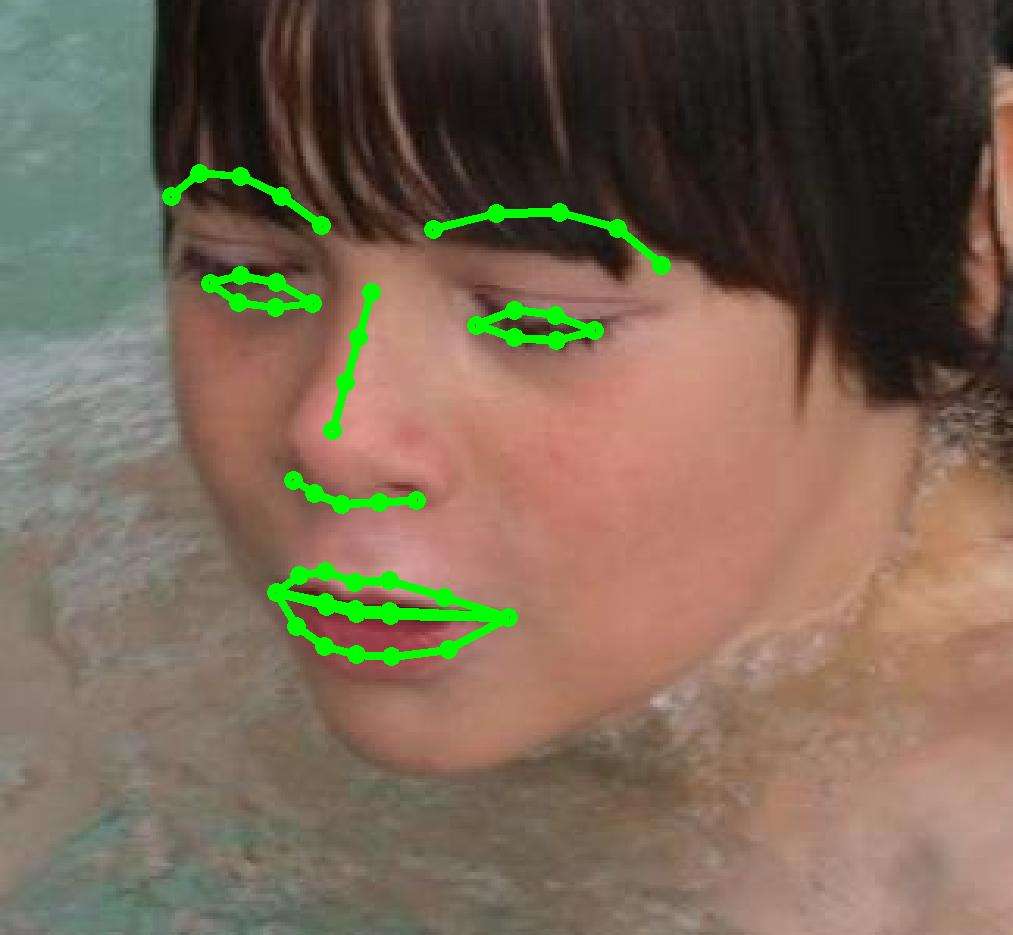}}\end{subfigure} \\

AAMs-PIs & CLMs-PIs & SDM-PIs & SDM-SIFT & FAR-PIs
\end{tabular}
\end{center}
\caption{Sample fitting results from the LFPW (rows: $1$-$2$), the HELEN (rows: $3$-$4$), and the AFW (rows: $5$-$6$) databases. (The Figure is best viewed in color)}
\label{fig:gfa_exmples_imgs}
\end{figure}

The CEDs produced by all methods for the LFPW (test set), the HELEN (test set), and the AFW databases are depicted in Fig. \ref{fig:gfa_results}. Clearly, the FAR outperforms the AAMs-PIs, the CLMs-PIs, and the SDM-PIs.  More specifically, for normalized error equal to $0.05$\footnote{This value was found by visually inspecting the results.} the FAR yield an $20.1\%$, $21.5\%$ and $24.6\%$  improvement compared to that obtained by the AAMs-PIs across the test databases, respectively. A few fitting examples from the test databases are depicted in Fig. \ref{fig:gfa_exmples_imgs}.

\renewcommand\tabcolsep{0pt}

\textbf{State-of-the-art method and features}:

In this experiment, the FAR is compared against the state-of-the-art method SDM \cite{sdm}. In particular, the implementation as well as the pre-trained model provided by the authors were used. Both the FAR and the SDM were initialized by using the same detector \cite{ramanan}, and the SDM was initialized following the instructions of the authors\footnote{\url{http://www.humansensing.cs.cmu.edu/intraface}}. The CEDs from this experiment are shown in Fig.\ref{fig:gfa_results_sdm_far} where  the FAR achieves comparable performance with that obtained by SDM using only \textbf{$\b{5}\b{0}\b{0}$ frontal images}. It is worth mentioning, the SDM was trained on thousand images captured under several variations including different poses, illuminations and expression. Furthermore, the SDM method takes full advantage of SIFT - a powerful hand-crafted feature - while the FAR employs only pixel intensities.

\subsection{Pose-invariant face recognition and verification}\label{subsec:face_recognition}

The performance of the FAR on pose invariant face recognition and verification is assessed by conducting experiments on the MultiPIE, FERET, and LFW databases. Apart from the experimental results reported in this section more can be found on the supplementary materials.

\subsubsection{Pose invariant face recognition}

The frontal views of all images were reconstructed following the methodology described in Section \ref{subsec:frontal_rec}. The reconstructed images were cropped to remove the surrounding black pixels. The Image Gradient Orientations (IGOS) features  \cite{tzimiro_igos} were used for image representation. The dimensionality of IGOs was reduced by applying PCA. The classifier in \cite{crc} was used.   

The performance of the FAR is compared against of that obtained by the following 2D based methods LGBP \cite{Feret_LGBP_2d_iccv_2005} and PIMRF \cite{Feret_chellapa_2d_tip_2013}, 3D based methods 3DPN \cite{Feret_akshay_3d_iccv_2011}, EGFC \cite{Feret_3d_eccv_2012}, and PAF \cite{Feret_yi_3d_cvpr_2013} as well as the Deep learning based methods SPAE \cite{mpie_spae_cvpr_2014} and DIPFS \cite{mpie_dipfs_iccv_2013}. It should be noticed that all methods were evaluated under the fully automatic scenario; where both the bounding box of the face region and the facial landmark were located automatically.

\textbf{Results on FERET}: One frontal image, denoted as `ba', from each of the $200$ subjects was used to form the gallery set, while the images captured at $6$ different poses i.e., from $-40^\circ$ to $40^\circ$ were selected as the probe images. 

\renewcommand\tabcolsep{2pt}
\begin{table}[!htb]
\begin{footnotesize}
  \begin{center}
  \caption{Recognition rates ($\%$) achieved by the compared methods on the FERET database.}
    \begin{tabular}{|l||c|c|c|c|c|c||c||}
    \hline
    \multicolumn{1}{|l||}{\multirow{2}[4]{*}{Method}} & \multicolumn{1}{c|}{bh} & \multicolumn{1}{c|}{bg} & \multicolumn{1}{c|}{bf} & \multicolumn{1}{c|}{be} & \multicolumn{1}{c|}{bd} & \multicolumn{1}{c||}{bc} & \multicolumn{1}{c||}{\multirow{2}[4]{*}{Avg}}\\
    \multicolumn{1}{|l||}{} & $-40^\circ$   & \multicolumn{1}{c|}{$-25^\circ$} & \multicolumn{1}{c|}{$-15^\circ$} & \multicolumn{1}{c|}{$+15^\circ$} & \multicolumn{1}{c|}{$+25^\circ$} & \multicolumn{1}{c||}{$+40^\circ$} & \multicolumn{1}{c||}{}   \\
     \hline
    LGBP \cite{Feret_LGBP_2d_iccv_2005}           & 90.5\% & 98.0\% & 98.5\% & 97.5\% & 97.0\% & 91.9\% & 95.6\%  \\ \hline
    3DPN \cite{Feret_akshay_3d_iccv_2011}    & 90.5\% & 98.0\% & 98.5\% & 97.5\% & 97.0\% & 91.9\% & 95.6\%  \\ \hline
    PIMRF \cite{Feret_chellapa_2d_tip_2013}  & 91.0\% & 97.3\% & 98.0\% & 98.5\% & 96.5\% & 91.5\% & 95.5\%  \\ \hline
    PAF \cite{Feret_yi_3d_cvpr_2013}         & 98.0\% & 98.5\% & 99.25\% & 99.25\% & 98.5\% & 98.0\% & 98.56\% \\ \hline
    FAR                                 & 96.5\% & 99.0\% & 100.0\% & 100.0\% & 100\%& 96\% & 98.58\% \\ \hline
    \end{tabular}%
  \label{tbl:fr_feret}%
  \end{center}
\end{footnotesize}
\end{table}%

In Table \ref{tbl:fr_feret} the recognition rates achieved by the competing methods in the different poses are reported. Clearly, the FAR (recognition accuracy $98.58\%$) outperforms both the 2D and 3D state-of-the-art methods PIMRF and PAF, respectively. It is worth mention that the PIMRF employs $200$ images from the FERET database (different from the test set) in order to train the frontal synthesizer. Consequently, the different lighting conditions of the database are taken into account. This is not the case for the FAR where only frontal images from a generic in-the-wild database (i.e., the LFPW and HELEN) have been used. Even the FAR does not use any kind of 3D information it outperforms the PAF where an elaborated 3D model (trained from $4.624$ facial scans) has been used to find the 3D pose and extract pose adaptive features. The reported results of the EGFC \cite{Feret_3d_eccv_2012} were not included in Table \ref{tbl:fr_feret} as they were obtained using a semi-automatic protocol (i.e., 5 manually annotated landmark points used).

\textbf{Results on \textit{MultiPIE}}: The images of the $137$ subjects (Subject ID 201: 346) with `Neutral’ expression and poses $-30^\circ : +30^\circ$  captured under $4$ different sessions were selected. The gallery was created by the frontal images of the earliest session for each subject, while the rest images including frontal and non-frontal views were used as probes. It should be mentioned that images of first $200$ subjects which include all poses ($4207$ in total) were used for training purposes. In particular, the above mentioned images were used in the 3DPN    to train view-based models, in the SPAE, DIPFS to train the deep networks, and in the EGFC to train the pose estimator and matching model. The recognition accuracy achieved by the just mentioned methods is reported in Table \ref{tbl:fr_multipie}. Again, the FAR outperforms four out of five methods that is compared to, verifying the high quality of the frontalized images. The FAR also performs comparable with the DIPFS by simply using just $500$ frontal images outside the MultiPIE.

\begin{table}[h]
 \begin{footnotesize}
  \begin{center}
  \caption{Recognition rates ($\%$) achieved by the compared methods on the MultiPIE database.}
    \begin{tabular}{|l||c|c|c|c|c||c||}
    \hline
    \multicolumn{1}{|l||}{\multirow{2}[4]{*}{Method}} & \multicolumn{1}{c|}{$130\_06$} & \multicolumn{1}{c|}{$140\_06$} & \multicolumn{1}{c|}{$051\_07$} & \multicolumn{1}{c|}{$050\_08$} & \multicolumn{1}{c||}{$041\_08$}  & \multicolumn{1}{c||}{\multirow{2}[4]{*}{Avg}}  \\
     \multicolumn{1}{|l||}{} & $-30^\circ$   & \multicolumn{1}{c|}{$-15^\circ$} & \multicolumn{1}{c|}{$0^\circ$} & \multicolumn{1}{c|}{$15^\circ$} & \multicolumn{1}{c||}{$30^\circ$}  & \multicolumn{1}{c||}{}  \\
     \hline
    PIMRF \cite{Feret_chellapa_2d_tip_2013} & 89.7\% & 91.7\% & 92.5\% & 91.0\% & 89.0\% & 90.78\% \\ \hline
    3DPN \cite{Feret_akshay_3d_iccv_2011}   & 91.0\% & 95.7\% & 96.9\% & 95.7\% & 89.5\% & 93.76\% \\ \hline
    SPAE \cite{mpie_spae_cvpr_2014}         & 92.6\% & 96.3\% & -      & 95.7\% & 94.3\% & 94.72\% \\ \hline
    EGFC \cite{Feret_3d_eccv_2012}          & 95.0\% & 99.3\% & -      & 99.0\% & 92.9\% & 96.55\% \\ \hline
    DIPFS \cite{mpie_dipfs_iccv_2013}       & 98.5\% & 100\%  & -      & 99.3\% & 98.5\% & 99.07\%  \\ \hline    
    FAR                                     & 94.3\% & 98.7\% & 99.4\% & 97.3\% & 95.6\% & 97.06\%  \\ \hline
    \end{tabular}%
  \label{tbl:fr_multipie}%
  \end{center}
 \end{footnotesize}
\end{table}%

\subsubsection{Face verification}

The performance of the FAR in the face verification under in-the-wild conditions is assessed by conducting experiment in the LFW database, using the `image-restricted, no outside data results' protocol. The reported results are obtained using $10$-fold cross validation.

In this experiment the basis $\b{U}$ and the detector \cite{ramanan} were not used since they based on images outside the database. To create the initializations and a new $\b{U}_{LFW}$, the method for automatic construction of deformable models presented in \cite{autoConDM} was employed. The goal of this method is to build a deformable model using only a set of images with the corresponding bounding boxes. To define the bounding boxes without using a pre-trained detector, the deep funneled images of the LFW \cite{DeepLFW} were employed. Therefore, since these images are aligned the exact bounding box is known. Subsequently, a deformable model was built automatically from the training images of each fold. The created model was fitted to all images and those (from training images) with fitted shapes similar to mean shape were selected to build the bases $\b{U}_{LFW}$. In each fold the images were frontalized using the $\b{U}_{LFW}$ and they were  cropped next. In the sequel, the gradient orientations $\phi_1$, $\phi_2$ of each image pair were extracted and the cosine of difference between them  $\Delta \phi=\phi_1-\phi_2$ normalized to the range $[0-2\pi]$,  was used as the feature of the pair. These features are classified by a Support Vector Machine (SVM) with an RBF kernel. The performance of the FAR is compared against that obtained by the MRF-MLBP \cite{arashloo2013efficient}, Fisher Vector Faces \cite{simonyan2013fisher} and the EigenPEP \cite{li2014eigen} methods\footnote{These methods are the $3$ top performing in the LFW according to \url{http://vis-www.cs.umass.edu/lfw/results}.}. The mean classification accuracy and the corresponding standard deviation on LFW are reported in Table \ref{tbl:lfw_results}. By inspecting Table \ref{tbl:lfw_results} the FAR outperforms the MRF-MLBP and the Fisher Vector Faces and performance comparably with the recently published method EigenPEP.

\begin{table}[ht]
\begin{footnotesize}
\begin{center}
\begin{tabular}{|c|c|}
\hline
MRF-MLBP \cite{arashloo2013efficient} & $0.7908 \pm 0.0014$\\\hline
Fisher vector faces \cite{simonyan2013fisher} & $0.8747 \pm 0.0149$  \\\hline
EigenPEP \cite{li2014eigen} & $0.8897 \pm 0.0132$  \\\hline
FAR & $0.8881 \pm 0.0078$\\ \hline
\end{tabular}
\end{center}
\caption{Mean classification error and standard deviation on the LFW database.}
\label{tbl:lfw_results}
\end{footnotesize}
\end{table}

\section{Conclusions}

In this paper we developed the first, to the best our knowledge, method that jointly performs landmark localization and face frontalization using only a simple statistical model of few hundred frontal images. The proposed method outperforms state-of-the-art methods for face landmark localization that were trained on thousands of images in many poses and achieves comparable results in pose invariant face recognition and verification without using 3D elaborate models or Deep Learning-based features extraction.

\bibliographystyle{ieee}
\bibliography{Face_frontalization_for_Alignment_and_Recognition}

\end{document}